\documentclass{article}

\usepackage[preprint]{neurips_2025}

\usepackage[utf8]{inputenc} % allow utf-8 input
\usepackage[T1]{fontenc}    % use 8-bit T1 fonts
\usepackage{url}            % simple URL typesetting
\usepackage{booktabs}       % professional-quality tables
\usepackage{amsmath, amssymb, amsopn, amsthm} 
\usepackage{amsfonts}       % blackboard math symbols
\usepackage{nicefrac}       % compact symbols for 1/2, etc.
\usepackage{microtype}      % microtypography
\usepackage{xcolor}         % colors
\usepackage{subcaption}
\usepackage[shortlabels]{enumitem}
\usepackage{stix}
\usepackage[most]{tcolorbox}
\usepackage{svg}
\svgpath{{./}} 

\usepackage[noline,plain,algosection,linesnumbered,titlenumbered]{algorithm2e}
\makeatletter
\renewcommand{\@algocf@capt@plain}{above}% formerly {bottom}
\makeatother

\usepackage[colorlinks,backref=page]{hyperref}

\renewcommand*{\backref}[1]{}
\renewcommand*{\backrefalt}[4]{%
    \ifcase #1%
          \or (Cited on page~#2.)%
          \else (Cited on pages~#2.)%
    \fi%
    }
\usepackage[capitalize,noabbrev,nameinlink]{cleveref}
\crefformat{equation}{(#2#1#3)}

\numberwithin{equation}{section}
\newtheoremstyle{exampstyle}
{4pt} % Space above
{4pt} % Space below
{\itshape} % Body font
{} % Indent amount
{\bfseries} % Theorem head font
{.} % Punctuation after theorem head
{.5em} % Space after theorem head
{} % Theorem head spec (can be left empty, meaning `normal')
\theoremstyle{exampstyle}
\newtheorem{theorem}{Theorem}[section]

\newtheorem{nb}[theorem]{Note}

\theoremstyle{definition}
\newtheorem{definition}[theorem]{Definition}
\newtheorem{assumption}[theorem]{Assumption}

\theoremstyle{remark}

\DeclareMathOperator{\TopK}{TopK}
\DeclareMathOperator{\JumpReLU}{JumpReLU}

\newcommand{\RR}{\mathbb{R}}

\newcommand{\cL}{\mathcal{L}}

\definecolor{IBMultramarine}{HTML}{648fff}
\definecolor{IBMindigo}{HTML}{785ef0}
\definecolor{IBMmagenta}{HTML}{dc267f}
\definecolor{IBMorange}{HTML}{fe6100}
\definecolor{IBMgold}{HTML}{ffb000}

\definecolor{GGreen}{RGB}{0,128,0}
\hypersetup{
    linkcolor= {IBMmagenta},%{GGreen},%{red!50!black},
    citecolor={blue}, %{blue!50!black},
    urlcolor={IBMorange}
}

\newtcolorbox[auto counter, number within=section
    ]{tcbAlgorithm}[2][]{%
    enhanced,
    colback=IBMorange!10,
    colframe=IBMorange,
    title={\textbf{Algorithm \thetcbcounter.} \ifstrempty{#2}{\ignorespaces}{~#2.}}, 
    #1
    }

    \SetKwComment{Comment}{//}{}

\title{SplInterp: Improving our Understanding and Training of Sparse Autoencoders
}

\author{%
  Jeremy Budd\thanks{Corresponding author:   \texttt{j.m.budd@bham.ac.uk}} 
    \\
  School of Mathematics\\
  University of Birmingham\\
  \And
  Javier Ideami \\
  Ideami Studios  \\
  \And  Benjamin Macdowall Rynne \\
  Department of Mathematics and Statistics \\ University of Limerick \\
  \And
  Keith Duggar\\
  XRAI Inc. \\
  \And
  Randall Balestriero \\
  Department of Computer Science \\
  Brown University \\
}

\begin{document}

\maketitle

\begin{abstract}
  % The abstract paragraph should be indented \nicefrac{1}{2}~inch (3~picas) on
  % both the left- and right-hand margins. Use 10~point type, with a vertical
  % spacing (leading) of 11~points.  The word \textbf{Abstract} must be centered,
  % bold, and in point size 12. Two line spaces precede the abstract. The abstract
  % must be limited to one paragraph.
  Sparse autoencoders (SAEs) have received considerable recent attention as tools for mechanistic interpretability, showing success at extracting interpretable features even from very large LLMs. However, this research has been largely empirical, and there have been recent doubts about the true utility of SAEs. In this work, we seek to enhance the theoretical understanding of SAEs, using the spline theory of deep learning. By situating SAEs in this framework: we discover that SAEs generalise ``$k$-means autoencoders'' to be piecewise affine, but sacrifice accuracy for interpretability vs. the optimal ``$k$-means-esque plus local principal component analysis (PCA)'' piecewise affine autoencoder. We characterise the underlying geometry of (TopK) SAEs using power diagrams.  And we develop a novel proximal alternating method SGD (PAM-SGD) algorithm for training SAEs, with both solid theoretical foundations and promising empirical results in MNIST and LLM experiments, particularly in sample efficiency and (in the LLM setting) improved sparsity of codes. All code is available at: \url{https://github.com/splInterp2025/splInterp}
\end{abstract}

\section{Introduction}
\begin{figure}[h]
    \centering
\includegraphics[width=0.232\linewidth,,clip,trim ={0 40 0 0}]{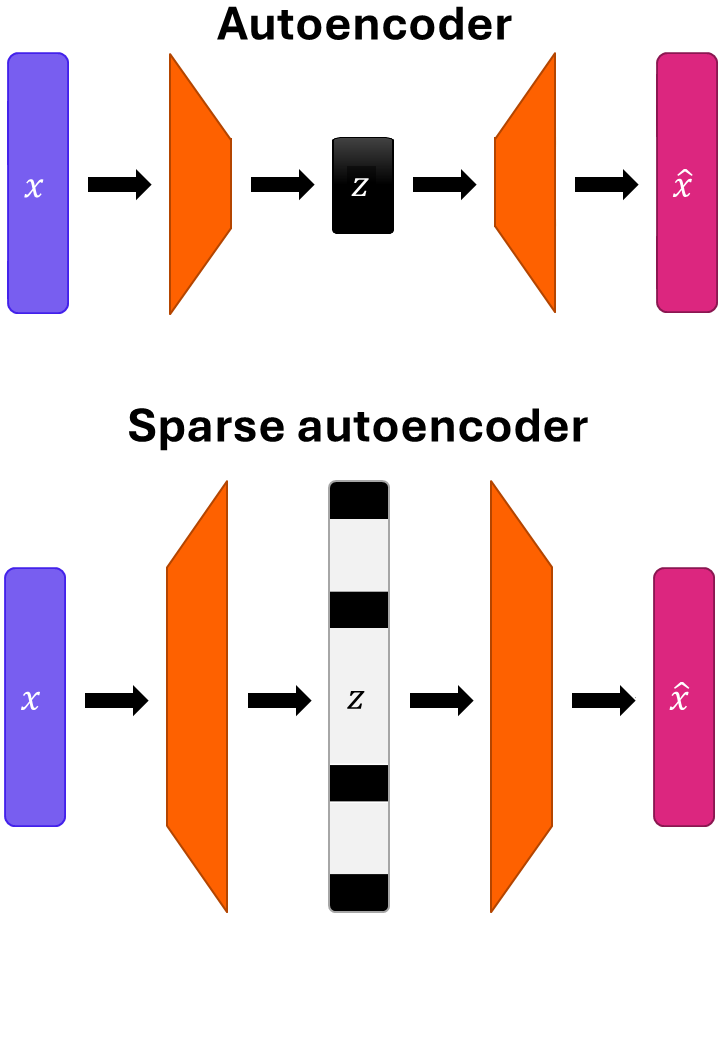}\hfill %\includegraphics[width=0.195\linewidth]{imagenew.png}\hfill
\includegraphics[width = 0.75\linewidth,clip,trim ={0 0 0 0}]{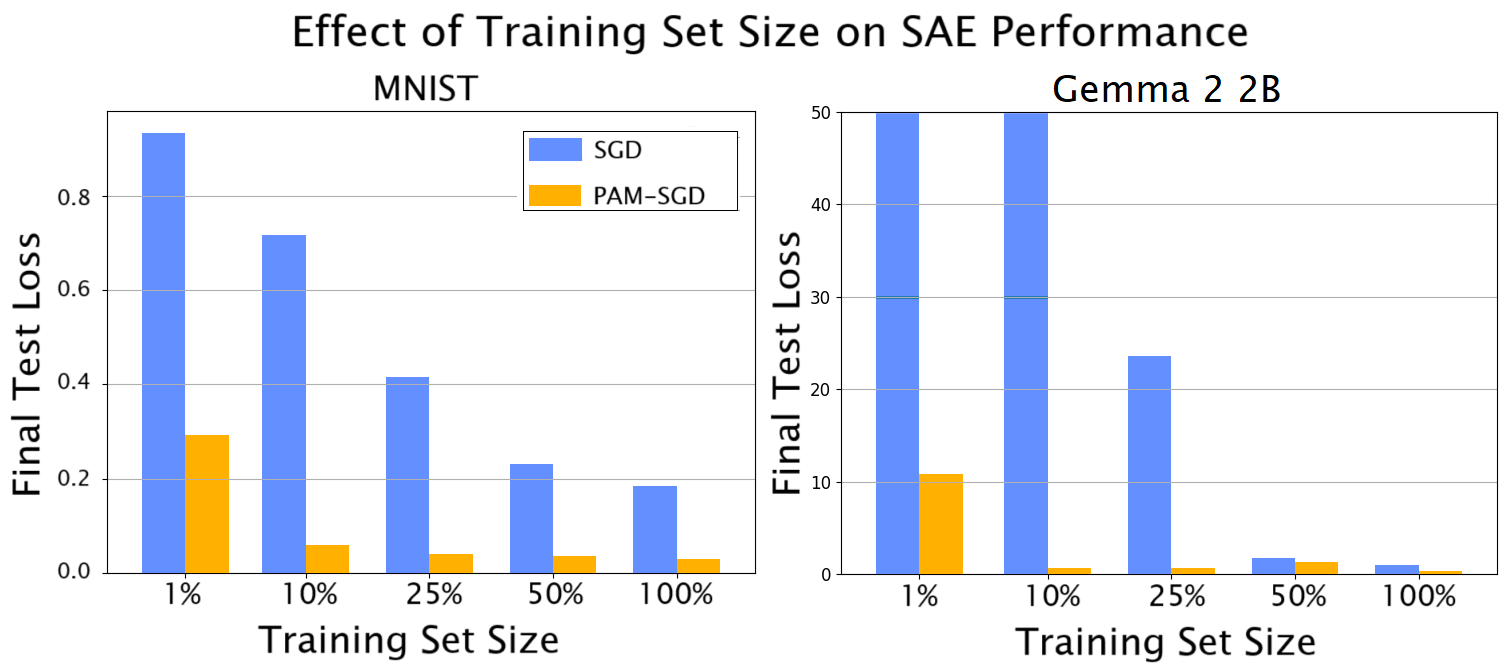}
    \caption{\small \textbf{Key ideas in this work.} \textbf{Left:} Sparse autoencoders (SAEs) vs. regular autoencoders; in an SAE, the code $z$ can have very high dimension, but most entries are zero (greyed out). %\textbf{Middle:}  Visualising our link between SAEs and $k$-means extended via local PCA--colours indicate piecewise affine regions. 
    \textbf{Right:} The superior sample efficiency (on MNIST and Gemma-2-2B) of our novel proximal alternating method SGD (PAM-SGD) algorithm.}
    \label{fig:intro}
\end{figure}
%\subsection{Motivation and contributions of this work} 
One of the fundamental challenges in modern AI is the \textit{interpretability} of machine learning systems: how can we look inside these increasingly more complicated (and capable) ``black boxes''? The better we understand what makes these systems tick, the better we can diagnose problems with them, direct their behaviour, and ultimately build more effective, reliable, and fair systems. Indeed, for some tasks interpretability may be a legal requirement, for example Article 86 of the European Union's AI Act describes a ``right to explanation'' for persons affected by the decisions of certain types of AI system. 

A mechanistic interpretability technique that has seen considerable recent attention is to use \textit{sparse autoencoders} (SAEs), see e.g. \cite{bricken2023monosemanticity,huben2024sparse,gao2025scaling}. A major obstacle in interpreting a neural network is that neurons seem to be \textit{polysemantic}, responding to mixtures of seemingly unrelated features \cite{olah2017feature}. One hypothesis is that this is caused by \textit{superposition}, a phenomenon in which a neural network represents features via (linear) combinations of neurons, to pack in more features. In \cite{elhage2022toy}, superposition was shown to arise in a toy model in response to sparsity of the underlying features. The key idea of using SAEs was that these might be able to disentangle this superposition and extract \textit{monosemantic} features. 

Initial work found some success, e.g. \cite{templeton2024scaling} used SAEs to extract millions of features from Claude 3 Sonnet, including highly interpretable ones such as a ``Golden Gate Bridge'' feature, which could be used to make Claude incessantly talk about the bridge. However, there have been some recent doubts about the utility of SAEs for mechanistic interpretability. In a recent post by the Google DeepMind mechanistic interpretability team
\cite{Smith_Rajamanoharan_Conmy_McDougall_Kramar_Lieberum_Shah_Nanda_2025}, works finding key issues with SAEs were highlighted, e.g. \cite{leask2025sparseautoencoderscanonicalunits}, and SAEs were found to underperform linear probes at a downstream task. The team argued that whilst SAEs are not useless, they will not be a game-changer for interpretability, and speculated that the field is over-invested in them. Furthermore, in what we will dub a ``dead salmon'' experiment (in honour of \cite{BENNETT2009salmon}), \cite{heap2025sparseautoencodersinterpretrandomly} found that SAEs can extract features from randomly weighted Transformers that have very similar auto-interpretability scores to features extracted from a trained network--suggesting that SAE ``interpretations'' may not reflect what is actually going on in a model. 

This empirical uncertainty motivated us to look at SAEs through a more theoretical lens, inspired by the \textit{spline theory of deep learning} \cite{balestriero18b}. Using this perspective, we:
\begin{enumerate}[I.]
    \item \textit{Unify}, situating SAEs within the spline theory framework, and showing how SAEs form a bridge between the classical ML techniques of $k$-means and principal component analysis (PCA) and contemporary deep learning. (\cref{sec:splinegeometry}, proofs in \cref{app:splineproofs})
    \item \textit{Interpret}, characterising and visualising the spline geometry of (TopK) SAEs in terms of weighted Voronoi diagrams. (\cref{sec:splinegeometry,app:Voronoi,app:splineproofs})
    \item \textit{Innovate}, developing a novel proximal alternating method SGD (PAM-SGD) algorithm for training SAEs, with both solid theoretical foundations and promising empirical results, which is inspired by the spline geometric way of thinking. In particular, we find in both MNIST and LLM experiments that PAM-SGD outperforms SGD in low-training-data settings, addressing an important concern with SAEs. (\cref{sec:optdecoding,app:optdecoding,app:figures})
\end{enumerate}

\section{The spline geometry of sparse autoencoders (SAEs)} 
\label{sec:splinegeometry}
\subsection{A primer on SAEs}
% \begin{tcolorbox}[colback=IBMultramarine!20,colframe=IBMultramarine]  

% \begin{definition}[SAEs] \label{def:SAEs}
An SAE composes an \textit{encoding}, which maps an input $x \in \RR^n$ to a \textit{code} $z \in \RR^d$ (enginereed to be sparse), with a \textit{decoding} which maps $z$ to an output $\hat x \in \RR^n$ (engineered so that $\hat x \approx x$). Unlike a traditional autoencoder, in an SAE one may choose the hidden dimension $d\gg n$, but the sparsity of $z$ will be engineered to be much less than $n$, see \cref{fig:intro}(left).
The SAE encoding is given by
\begin{equation*}
    z := \rho(W_\text{enc}x + b_\text{enc}),
\end{equation*}
where  $W_\text{enc} \in \RR^{d \times n}$, $b_\text{dec} \in \RR^d$, and $\rho$ is a given \textit{activation function}. Notable choices for $\rho$ include ReLU \cite{bricken2023monosemanticity}, JumpReLU \cite{rajamanoharan2024jumpingaheadimprovingreconstruction} where  $\tau\in \RR$ is a parameter and
\[
\rho(v)_i := \begin{cases}
    v_i, &\text{if } v_i> \tau, \\ 0,& \text{otherwise},
\end{cases}
\]
 and TopK \cite{makhzani2014ksparseautoencoders,gao2025scaling} where $K \in \mathbb{N}$ is a parameter and
\[
\rho(v)_i := \begin{cases}
    v_i, &\text{if $v_i$ is among the $K$ largest entries of $v$,} \\ 0, &\text{otherwise}.
\end{cases}
\]
The decoding is then given by
\[
\hat x := W_\text{dec}z + b_\text{dec},
\]
where $W_\text{dec} \in \RR^{n \times d}$ and $b_\text{dec} \in \RR^n$. The columns of $W_\text{dec}$ can be understood as \textit{dictionary atoms} (see \cite{olshausen1996emergence}) which are sparsely recombined (with bias) to recover $\hat x$.

The full SAE is therefore given by 
\begin{align*}
    S_{\rho}(x)&:= W_\text{dec} \rho(W_\text{enc}x + b_\text{enc})+b_\text{dec}. 
\end{align*}
Finally, following \cite{rajamanoharan2024jumpingaheadimprovingreconstruction}, given training data $\{x^r\}_{r=1}^N \in \RR^n$, we will consider loss functions for training an SAE of the form:
\begin{align*}
    \cL &= \sum_{r=1}^N \|S_{\rho}(x^r) - x^r\|_2^2 + \lambda \cL_\text{sparsity}\left(\{\rho(W_\text{enc}x^r + b_\text{enc})\}_{r=1}^N\right) + \cL_\text{aux},%\qquad \text{[OR  KL LOSS??]},
    % \\
    % \cL_{\TopK} &= \sum_{r=1}^N \|S_{\TopK}(x^r) - x^r\|_2^2.
\end{align*}
where $\cL_\text{aux}$ might include regularisation, e.g. weight decay. Some activations, e.g. TopK, always produce a sparse $z$, so one may set $\cL_\text{sparsity} =0$. Others, e.g. ReLU, do not inherently make $z$ sparse, in which case common choices of $\cL_\text{sparsity}$ include the $\ell_1$ norm \cite{bricken2023monosemanticity}, the $\ell_0$ norm \cite{rajamanoharan2024jumpingaheadimprovingreconstruction}, and the Kullback--Leibler divergence to a sparse distribution \cite{ng2011sparse}.
% \end{definition}
% \end{tcolorbox}
\subsection{SAEs are piecewise affine splines}
We first note a simple fact about our SAEs, also observed in \cite{hindupur2025projectingassumptionsdualitysparse} (in different notation). In all three cases of ReLU, JumpReLU, and TopK, for some $S\subseteq\{1,...,d\}$ we have that 
\[
\rho(v) = P_Sv, 
\]
where $P_S \in \RR^{d \times d}$ is the projection that zeroes the entries of $v$ which are not in $S$. In the case of JumpReLU (of which ReLU is a special case) $S$ is the set of indices $i$ such that $v_i > \tau$, and in the case of TopK $S$ is the set of indices containing the largest $K$ entries. Therefore, let us define 
\begin{align*}
    \Omega_S^{\JumpReLU} &:= \{ x\in \RR^n : \forall i \in S, (W_\text{enc}x + b_\text{enc})_i > \tau \text{ and } \forall j \notin S, (W_\text{enc}x + b_\text{enc})_j < \tau\}, \\ \Omega_S^{\TopK}&:= \{ x\in \RR^n : \forall i \in S, j \notin S,\: (W_\text{enc}x + b_\text{enc})_i > (W_\text{enc}x + b_\text{enc})_j\},
\end{align*}
where in the former $S$ can be any subset of $\{1,...,d\}$ and in the latter $S$ must be a subset of size $K$. Then for $ \rho = \JumpReLU$ or $\rho =\TopK$ the SAE becomes:
\[
S_{\rho}(x) = \begin{cases}
   W_\text{dec} P_S(W_\text{enc}x + b_\text{enc})+b_\text{dec}, & x \in \Omega^\rho_S,
\end{cases}
\]
which is a piecewise affine spline. Note that the $\Omega_S^{\JumpReLU}$ and $ \Omega_S^{\TopK}$ do not entirely partition the space, e.g. what if $W_\text{enc}x + b_\text{enc}$ has an entry equal to $\tau$ or has ties for the top $K$? Such $x$s (a set of measure zero) form the \textit{boundaries} of these pieces, and $S_\rho$ is discontinuous at these boundaries (except in the ReLU case, i.e. $\tau = 0$). Both $\Omega_S^{\rho}$ can be written in the form $\{ x\in \RR^n : Hx > c\}$,
for appropriate matrices $H$ and vectors $c$, see \cref{thm:openpoly}. They are thus open and convex sets, and by \cref{thm:openpoly} are the interiors of convex polyhedra except in degenerate cases of $W_\text{enc}$. 

Going beyond this simple characterisation, we introduce a new geometric characterisation of the TopK pieces as the cells of a \textit{power diagram}. (For some visualisations of these notions, see \cref{app:Voronoi}.)
\begin{tcolorbox}[colback=IBMultramarine!20,colframe=IBMultramarine]  
\begin{definition}[Power and Voronoi diagrams]
    A \textit{power diagram} (a.k.a. a  \textit{Laguerre–Voronoi diagram}) is a partition of $\RR^n$ into $k$ cells, defined by taking centroids $\{\mu_i\}_{i=1}^k \in \RR^n$ and weights $\{\alpha_i\}_{i=1}^k \in \RR$ and defining the $i^\text{th}$ cell to be
    \[
    C_i := \{x \in \RR^n : \|x - \mu_i\|_2^2 - \alpha_i <  \|x - \mu_j\|_2^2 - \alpha_j \text{ for all } j\neq i \}. 
    \]
    A \textit{Voronoi diagram} is given by the special case when the $\alpha_i$ are constant in $i$. 

    We further define a \textit{$K^\text{th}$-order power diagram} with centroids $\{\mu_i\}_{i=1}^k \in \RR^n$, weights $\{\alpha_i\}_{i=1}^k \in \RR$, and $\binom{k}{K}$ cells, where for $S\subseteq\{1,...,k\}$ with $|S|=K$, let the $S^\text{th}$ cell be
    \[
    C_S := \{x \in \RR^n : \|x - \mu_i\|_2^2 - \alpha_i <  \|x - \mu_j\|_2^2 - \alpha_j \text{ for all $i \in S$ and $j \in S^c$} \}.
    \] 
    An identical power diagram (of any order) is given if a constant is added to all the weights. 
\end{definition}
\end{tcolorbox}
%We now show that $K^\text{th}$-order power diagrams are just a special type of power diagram. 
\begin{tcolorbox}[colback=gray!10,colframe=gray]
\begin{nb}\label{nb:hexagon}
    All $K^\text{th}$-order power diagrams are power diagrams, see \cref{thm:Kthpowerdiagram}, but the converse does not hold, i.e. not all power diagrams with $\binom{k}{K}$ centroids and weights can be written as a $K^\text{th}$-order power diagram with $k$ centroids and weights. As a counterexample, let $k = 4$, $K=2$, and the power diagram centroids be the six vertices of the regular hexagon. There are no four vectors in $\RR^2$ whose pairwise means are the vertices of the regular hexagon.  
\end{nb}
\end{tcolorbox}
\begin{tcolorbox}[colback=IBMgold!20,colframe=IBMgold] 
\begin{theorem}
\label{thm:powerdiagram}
    The cells $\{\Omega^{\TopK}_S\}$ form a $K^\text{th}$-order power diagram with $\binom{d}{K}$ cells. Conversely, for any $K^\text{th}$-order power diagram with $\binom{d}{K}$ cells with centroids $\{\mu_i\}_{i=1}^d$ and weights $\{\alpha_i\}_{i=1}^d$, there exist $W_{enc} \in \RR^{d \times n}$ and $b_{enc} \in \RR^d$ such that the resulting TopK SAE is affine on the cells of that $K^\text{th}$-order power diagram. The translations between each setting are given by:
    \begin{subequations}
                \label{eq:omegatopower}
    \begin{align}
          e^T_i W_{enc}= \mu_i^T && \text{and} && (b_{enc})_i = \frac12\alpha_i - \frac12 \|\mu_i \|_2^2, \\
    \mu_i= W^T_{enc}e_i && \text{and} && \alpha_i = 2(b_{enc})_i +  \|W^T_{enc}e_i\|_2^2,
    \end{align}
        \end{subequations}
        for all $i$, where $e_i$ is the elementary basis vector with $1$ in coordinate $i$ and $0$ in every other coordinate.  It follows from \Cref{thm:Kthpowerdiagram} 
        that the cells $\{\Omega^{\TopK}_S\}$ form a power diagram with $\binom{d}{K}$ cells, given by centroids $\{\nu_S\}$ and weights $\{\beta_S\}$ defined by 
    \begin{align}\label{eq:TopKpower2}
        \nu_S := \frac1K \sum_{i \in S} W^T_{enc} e_i && \text{and} && \beta_{S} := \left\|  \frac1K \sum_{i \in S} W^T_{enc} e_i   \right\|_2^2 +  \frac1K \sum_{i \in S} 2(b_{enc})_i.
    \end{align}
    The converse is false: a given power diagram with $\binom{d}{K}$ cells can describe the cells upon which a TopK SAE is piecewise affine if and only if it can be written as a $K^\text{th}$-order power diagram. 
\end{theorem}
\end{tcolorbox}
 What \cref{thm:powerdiagram} tells us is exactly the spline geometries that TopK SAEs can have, namely that these are exactly the $K^\text{th}$-order power diagrams. Indeed, we can explicitly derive the encoding parameters that give rise to a particular spline geometry. This opens the door to engineering TopK SAEs with desirable geometric features, by translating those features into constraints on the parameters. As an example of a geometric feature one might desire to encourage, \cite{humayun2024deep} related the generalisability and robustness produced by neural network grokking (see \cite{power2022grokkinggeneralizationoverfittingsmall}) to the local complexity of the spline geometry. 

\subsection{SAEs, $k$-means, and principal component analysis (PCA)}\label{sec:SAEandPCA}
By the above, all of the above SAEs are piecewise affine functions on regions $\{\Omega_S\}$, with rank $|S|$ on $\Omega_S$. 
% have the form 
% \begin{align*}
% S_{\rho}(x) &= \begin{cases}
%     \sum_{i \in S} W_\text{dec}^i ( e_i^TW_\text{enc}x + (b_\text{enc})_i) + b_\text{dec}, & x \in \Omega_{S},
%     \end{cases}
%\end{align*}
% where $%w_i = 
% W_\text{dec}^i \in \RR^n$ is the $i^\text{th}$ column of $W_\text{dec}$.
We can compare this to the $k$-means clustering. 
\begin{tcolorbox}[colback=IBMultramarine!20,colframe=IBMultramarine] 
\begin{definition}[$k$-means clustering]
    Given data $\{x^r\}_{r=1}^N \in \RR^n$, the \textit{$k$-means clustering} \cite{Steinhaus1957} seeks $k$ regions $\{R_i\}_{i=1}^k \subseteq \RR^n $ and centroids $\{\nu_i\}_{i=1}^k \in \RR^n$ minimising:
    \[
   \sum_{i=1}^k \sum_{x^r \in R_i} \|x^r - \nu_i\|^2_2.
    \]
    This is achieved when $\nu_i$ are the in-region means and $R_i$ are the following Voronoi cells:
    \[
    \nu_i = \bar{x}_i:= \frac{1}{|\{r:x^r \in R_i \}|} \sum_{x^r \in R_i} x^r, \qquad
    R_i = \{ x \in \RR^n : \|x - \nu_i\|_2^2 \leq \|x - \nu_j\|_2^2 \text{ for all $j \neq i$}\}. 
    \]
\end{definition}
\end{tcolorbox}
\begin{tcolorbox}[colback=gray!10,colframe=gray] 
\begin{nb} Suppose we have $k$ regions $\{\Omega_{S_i}\}_{i=1}^k$.
Consider the piecewise constant encoding $f_{enc}(x) := e_i$ for $x \in \Omega_{S_i}$ and the linear decoding $\hat x = W_\text{dec}z$ where $ W_\text{dec}\in \RR^{n \times k}$ has $i^\text{th}$ column $\nu_{S_i}$. Then if we replace $S_{\rho}$ by $F(x):=W_\text{dec}f_\text{enc}(x)$ in $\cL$ and take $\cL_\text{sparsity} = \cL_\text{aux}=0$,  
\[
\cL = \sum_{r=1}^N \|x^r - F(x^r) \|_2^2 =\sum_{i=1}^k \sum_{x^r \in \Omega_{S_i}} \|x^r - \nu_{S_i}\|_2^2,
\]
which is exactly the $k$-means objective. %for $k=\binom{d}{K}$. 
An SAE is therefore a generalisation of this ``$k$-means autoencoder'', where the encoding is allowed to be piecewise affine, the number of regions is allowed to be greater than the hidden dimension ($k=2^d$ in the (Jump)ReLU case and $k=\binom{d}{K}$ in the TopK case) and the SAE is overall piecewise affine with piecewise ranks $|S_i|$. 
\end{nb}
\end{tcolorbox}
But if SAEs are ``$k$-means autoencoders'' generalised to allow piecewise affine behaviour, how do they compare to the most general piecewise affine autoencoder?
 \begin{tcolorbox}[colback=IBMgold!20,colframe=IBMgold] 
\begin{theorem}\label{thm:generalPA}

On any partition $\{R_i\}_{i=1}^k$ define the general piecewise affine autoencoder with piecewise ranks $K_i$ (where $U_i,V_i\in \RR^{n \times K_i }$, $U_i^TU_i=I$, and $c_i \in \RR^n$),
\[
G(x) := \begin{cases} U_iV_i^Tx + c_i , & x \in R_i. \end{cases}
\]
%where unlike in $S_{\rho}$, the regions $R_i$ are independent of these parameters. 
Let $\cL_\text{aux} = 0$ and $\cL_\text{sparsity} = \|\cdot\|_0$, which counts the non-zero entries.
Then we have the loss 
\[
\cL = \sum_{i=1}^k  \sum_{x^r \in R_i} \|x^r - G(x^r) \|_2^2 + \lambda \sum_{i=1}^k  N_i K_i,
\]
where $N_i:= |\{r:x^r \in R_i\}|$. 
This has optimal parameters:
$U_i = V_i$ where the columns of $U_i$ are the top $K_i$ (normalised) eigenvectors $\{\xi^i_\ell\}_{\ell=1}^{K_i}$ of the covariance matrix
\[
X_i := \frac{1}{N_i}\sum_{x^r \in R_i} (x^r - \bar{x}_i) (x^r - \bar{x}_i)^T,
\]
$c_i = (I-U_iV_i^T) \bar{x}_i$, 
and optimal regions $R_i$ minimising
\begin{equation*} %\label{eq:PCAL}
\cL = \sum_{i=1}^k \left(\sum_{x^r \in R_i} \|(x^r -\bar{x}_i) \|_2^2 \right) + N_i K_i\left(\lambda - \frac{1}{K_i}\sum_{\ell=1}^{K_i} \lambda_{\ell}(X_i)\right),
\end{equation*}
where $ \lambda_{\ell}(X_i)$ are the eigenvalues of $X_i$ in descending order. 
Therefore,
\[
G(x) = \begin{cases}
    %U_iU_i^Tx + (I-U_iU_i^T)\bar{x}_i  = 
    \bar{x}_i + \sum_{\ell = 1}^{K_i} (\xi^i_\ell)^T(x-\bar{x}_i)\xi^i_\ell, & x \in R_i,
    \end{cases}
\]
 Thus, the general piecewise affine autoencoder combines a $k$-means-esque clustering %(with $k$ the number of regions) 
with a variable-rank local PCA correction. The optimal $K_i$ will occur when $\lambda_{K_i}(X_i) > \lambda \geq \lambda_{K_i + 1}(X_i) $, and hence $G$ is sensitive to the local intrinsic dimension of the data. 
\end{theorem}
\end{tcolorbox}
% \begin{tcolorbox}[colback=gray!10,colframe=gray] 
% \begin{nb}
% %In the $S_{\TopK}$ analogy, the $K_i = K$ for all $i$, and so the sparsity promoting term $\lambda N_i K_i$ plays no role. 
% If the $K_i$ are allowed to vary freely, then the optimal $K_i$ will occur when $\lambda_{K_i}(X_i) > \lambda \geq \lambda_{K_i + 1}(X_i) $, and hence $G$ is sensitive to the local intrinsic dimension of the data. 
% \end{nb}
% \end{tcolorbox}
\begin{tcolorbox}[colback=gray!10,colframe=gray] 
\begin{nb}\label{nb:GvsSAE}
The key takeaway from \cref{thm:generalPA} is that whilst SAEs generalise $k$-means to allow piecewise affine behaviour, they are less accurate than the most general piecewise affine autoencoder, which is $k$-means-like with a local PCA that tracks the local intrinsic dimension. 

However, $G$ achieves this greater accuracy by using a different encoding/decoding in each region. The code entry
    % It is important to contrast what $G$ does with what SAEs do. $G$ encodes $x \in R_i$ by subtracting the in-region mean and then computing the coordinates in the leading eigenvectors of $X_i$, i.e. for $j=1$ to $K_i$, 
    $ z_j := (\xi^i_j)^T(x-\bar{x}_i)$
    % .
    % $G$ decodes $z$ via those same eigenvectors and bias the in-region mean. %, i.e. 
    % % \[
    % % \hat x = \bar{x}_i + \sum_{j = 1}^{K_i} z_j\xi^i_j.
    % % \]
    % That is, $G$ uses $K_i$ encoding/decoding vectors which are unique to each region. This allows for more accuracy, but at the cost of the entries in the code $z$ 
    for an $x \in R_i$ is semantically unrelated to the $z_j$ for an $x$ in another region. By contrast, SAEs sacrifice some accuracy to encode all inputs as \emph{monosemantic} sparse codes $z$. This leads to SAEs having $d$ decoding vectors (the columns of $W_\text{dec}$) of which a subset are deployed per region, so these vectors are shared between regions. 
\end{nb}
\end{tcolorbox}

\subsection{Visualising the SAE bridge between $k$-means and PCA}
\begin{figure}[htbp]
    \centering
        \includegraphics[width=0.9\textwidth]{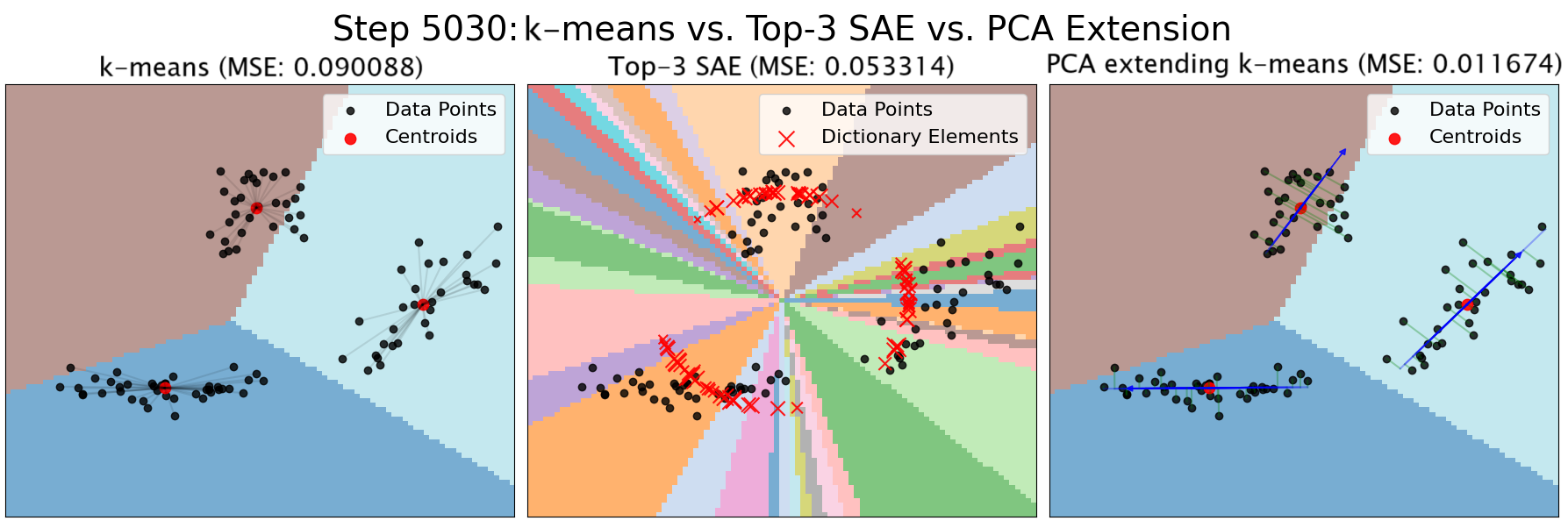}
    \caption{\textbf{Visualising the SAE bridge between $k$-means clustering and PCA.} {The cyan/brown boundary intersecting the point cloud (left and right figure) is a visualiser bug.}  %(Top) Top-1 SAE %(b,c) Soft top-1 SAE 
    %(Bottom) %(d) 
    %Top-3 SAE.
    }
    \label{fig:sae_spectrum}
\end{figure}

% This experiment explores how k-sparse autoencoders represent a theoretical bridge between clustering (k-means) and linear dimensionality reduction (PCA).
We just saw mathematically how SAEs generalise $k$-means, but sacrifice accuracy for interpretability compared to the optimal piecewise affine autoencoder, which is $k$-means-esque extended via local PCA. 
Our first experiment is a quick empirical exploration of this bridge. We chose TopK SAEs, and trained them on 100 points in $\RR^2$ drawn from $k=3$ clusters. For details on the experimental set-up, see \cref{app:SAEbridgesetup} and for another figure see \cref{app:figures}. We compared the SAEs to (i) a $k$-means ``autoencoding'', which maps each data point to its centroid, and (ii) a local 1-PCA extension of that $k$-means autoencoding. %, which allows a correction in the direction of the principal component for each cluster. 
This is the optimal $G$ from \cref{thm:generalPA} with $\{R_i\}_{i=1}^3$ fixed to be the $k$-means cells and $K_i \equiv1$.  
We found (see \cref{fig:sae_spectrum,fig:sae_spectrum2}) that the SAE consistently had a lower mean squared error (MSE) than $k$-means but a higher MSE than the PCA extension, in accordance with the theory.

%\subsection{Superposition??}

\section{A proximal alternating method (PAM-SGD) for training SAEs}
\label{sec:optdecoding}
\subsection{Convergence theory and the PAM-SGD algorithm}
It turns out that for any SAE that has the sort of loss function we have been considering, if we fix the encoding, then learning the decoding reduces to linear least squares regression: the decoding seeks an affine map that sends codes $z^r:=\rho(W_\text{enc}x^r + b_\text{enc})$ to $x^r$. 
We can therefore solve for the decoding in closed form, which suggests the idea of training an SAE by alternating between updating our encoding holding the decoding fixed, e.g. by SGD, and then updating our decoding holding the encoding fixed, via the closed form optimum. This dovetails well with the spline theory perspective, as we saw in the previous section that the spline geometry depends entirely on the encoding parameters. Therefore, this alternation can be viewed as (i) updating the SAE spline geometry to better use a given decoding, and then (ii) finding the most accurate decoding given that geometry. 

In particular, we will consider the following \textit{proximal alternating method} with quadratic costs to move, as this has solid theoretical foundations \cite{attouch2010proximal}:
\begin{subequations}
    \label{eq:PAM}
\begin{align}
   \label{eq:PAMenc} (W_\text{enc}^{t+1},b_\text{enc}^{t+1}) &= \operatorname*{argmin}_{W_\text{enc},b_\text{enc}}  &\sum_{r=1}^N \| W_\text{dec}^t \rho(W_\text{enc} x^r + b_\text{enc}) + b_\text{dec}^t - x^r\|_2^2  + F(W_\text{enc},b_\text{enc})& \\&&+ \mu_\text{enc}^t \|W_\text{enc} - W_\text{enc}^t\|_F^2 + \nu_\text{enc}^t\|b_\text{enc} - b_\text{enc}^t\|_2^2& \notag\\
     \label{eq:PAMdec} (W_\text{dec}^{t+1},b_\text{dec}^{t+1}) &= \operatorname*{argmin}_{W_\text{dec},b_\text{dec}}  &\sum_{r=1}^N \| W_\text{dec} \rho(W^{t+1}_\text{enc} x^r + b^{t+1}_\text{enc}) + b_\text{dec} - x^r\|_2^2 + G(W_\text{dec},b_\text{dec})&   \\&& + \mu_\text{dec}^t \|W_\text{dec} - W_\text{dec}^t\|_F^2 + \nu^t_\text{dec} \|b_\text{dec} - b_\text{dec}^t\|_2^2& \notag
\end{align}
\end{subequations}

We next use \cite{attouch2010proximal} to analyse the convergence of \cref{eq:PAM}, and find that under some assumptions (which will require minor adjustments to our SAE settings, see \cref{ass:conv,nb:phomodifications}) the sequence defined by \eqref{eq:PAM} converges to a critical point of the following loss:
\[
\cL(W_\text{enc},b_\text{enc},W_\text{dec},b_\text{dec}):= \sum_{r=1}^N \| W_\text{dec} \rho(W_\text{enc}x^r + b_\text{enc}) + b_\text{dec} - x^r\|_2^2  + F(W_\text{enc},b_\text{enc}) + G(W_\text{dec},b_\text{dec}).
\]
We summarise the convergence result as follows, for details see \cref{app:convproof}. 

\begin{tcolorbox}[colback=IBMgold!20,colframe=IBMgold,enhanced]

 \begin{theorem}\label{thm:PAMconvlite}
If \cref{ass:conv} holds, then from any initialisation, the sequence of SAE parameters defined by \cref{eq:PAM} monotonically decrease the loss $\cL$ and every convergent subsequence converges to a critical point of $\cL$. Furthermore, if the sequence is bounded (as would be ensured by e.g. weight decay) then the sequence converges to a critical point of $\cL$, with a rate that can be characterised (if the Kurdyka--Łojasiewicz exponent of $\cL$ is known).  
 \end{theorem}
 \end{tcolorbox}
\begin{tcolorbox}[colback=gray!10,colframe=gray,enhanced]
\begin{nb}
\cref{thm:PAMconvlite} (i.e., \cref{thm:PAMconv}) does \textbf{not} prove that our PAM-SGD method (see \cref{alg:PAMSAE} below) converges, as the theorem assumes that \cref{eq:PAMenc} is solved exactly, whilst in \cref{alg:PAMSAE} it will be only approximated via SGD. However, it does give some indication that the PAM-SGD method will approach an approximation to a critical point of $\cL$.
\end{nb}
\end{tcolorbox}

% \begin{tcolorbox}[colback=gray!10,colframe=gray,enhanced]
% \begin{nb}
%     Suppose that $F$, $G$, and $\rho$ satisfy $F(\alpha W_\text{enc},\alpha b_\text{enc}) = \alpha F(W_\text{enc},b_\text{enc})$, $G(\alpha W_\text{dec},\alpha b_\text{dec}) = \alpha G(W_\text{dec},b_\text{dec})$, and $\rho(\alpha z) = \alpha \rho(z)$ for all $\alpha > 0$, as is the case in both $\cL_{\ReLU}$ and $\cL_{\TopK}$. Then, if $(W^t_\text{enc},b^t_\text{enc},W^t_\text{dec},b_\text{dec}^t)$ is a valid trajectory, it follows that $(\alpha^{-1} W^t_\text{enc},\alpha^{-1} b^t_\text{enc},\alpha W^t_\text{dec},b_\text{dec}^t)$ would be a valid trajectory, giving the same SAE, under the change of parameters $(F,G,\mu_\text{enc}^t,\nu_\text{enc}^t,\mu_\text{dec}^t,\nu_\text{dec}^t) \mapsto (\alpha F, \alpha G,\alpha^2\mu_\text{enc}^t,\alpha^2\nu_\text{enc}^t,\frac{1}{\alpha^2}\mu_\text{dec}^t,\nu_\text{dec}^t)$. Hence, the scaling of the initialisation is important. 
% \end{nb}
% \end{tcolorbox}

% Let 
% \[
% G := \alpha\|W_\text{dec}\|^2_F + \beta\|b_\text{dec}\|_2^2 
% \] 
% be a weight decay. Then 

The optimal decoding for \cref{eq:PAMdec} with weight decay $G := \alpha\|W_\text{dec}\|^2_F + \beta\|b_\text{dec}\|_2^2$ can still be found in closed form, see \cref{thm:optdecoding2}. 
This gives the following novel method for training an SAE by solving \cref{eq:PAMdec} exactly, which we call a proximal alternating method SGD (PAM-SGD) algorithm.
\begin{tcolorbox}[colback=IBMorange!10,colframe=IBMorange,enhanced]
\begin{algorithm}[H]
\caption{PAM-SGD method (with optional weight decay) for learning an SAE. % with optimal decoding.
}
%\TitleOfAlgo{title} 
\label{alg:PAMSAE}
\KwIn{Initial SAE parameters $W_\text{enc}^0,b_\text{enc}^0,W_\text{dec}^0,b_\text{dec}^0$, iterations $t_{max}$, quadratic cost parameters $\{\mu_\text{enc}^t,\nu_\text{enc}^t,\mu_\text{dec}^t,\nu_\text{dec}^t\}_{t=0}^{t_{max}-1}$, weight decay parameters $\alpha,\beta$, activation $\rho$, learning rate $\eta$,  batch size $B$, SGD steps $M$. Training data $\mathcal{D}=\{x^r\}_{r=1}^N \in \RR^n $.}
%\KwData{Training data $\mathcal{D}=\{x^r\}_{r=1}^N \in \RR^n $.}
\KwOut{Final SAE parameters $W_\text{enc}^{t_{max}},b_\text{enc}^{t_{max}},W_\text{dec}^{t_{max}},b_\text{dec}^{t_{max}}$.}
$\bar{x} \gets \frac1N \sum_{r=1}^N x^r$\;
%$X \gets \operatorname{cat}(x^r-\bar x)$ \Comment*[r]{Concatenates the $x^r-\bar x$ into an $n \times N$ matrix}
% $X \gets x$\;
% $N \gets n$\;
\For{$t = 0$ \textbf{to} ${t_{max}-1}$}{
\vspace{0.1cm}
\tcc{\hfill--------------------------------------\textbf{ SGD Encoder update }--------------------------------------\hfill}
\vspace{0.1cm}
$(W_\text{enc}^{t+1},b_\text{enc}^{t+1}) \gets \texttt{SGD}(W_\text{enc}^{t}, b_\text{enc}^{t}, \mu_\text{enc}^t,\nu_\text{enc}^t,\eta,B,M)$ \Comment*[r]{Computes \cref{eq:PAMenc} via $M$ steps of SGD with learning rate $\eta$ and batch size $B$}
\vspace{0.1cm}
\tcc{\hfill----------------------------------\textbf{ Optimal Decoder update }----------------------------------\hfill}
\vspace{0.1cm}
 \ForEach{$x^r \in \mathcal{D}$}{
    $z_{t+1}^r \gets \rho(W_\text{enc}^{t+1} x^r + b_\text{enc}^{t+1})$ \; 
 }
 $\bar{z}_{t+1} \gets \frac1N \sum_{r=1}^N z_{t+1}^r$\;
\For{$r=1$ \textbf{to} $N$}{
$\psi_t^r \gets z^r_{t+1} -  \frac{N}{N+\beta+\nu^t_\text{dec}} \bar{z}_{t+1}$; \quad
$\phi_t^r \gets x^r - \frac{\nu^t_\text{dec}}{N+\beta+\nu^t_\text{dec}}b_\text{dec}^t  - \frac{N}{N+\beta+\nu^t_\text{dec}}\bar x$\;
}
$(\psi_t^{N+1},\phi_t^{N+1}) \gets \sqrt{\nu_\text{dec}^t} \frac{1}{N+\beta+ \nu_\text{dec}^t}\left( N\bar{z}_{t+1},N \bar x - (N+\beta) b^t_\text{dec}\right) $ \; \label{ln:psiN1}
$(\psi_t^{N+2},\phi_t^{N+2}) \gets \sqrt{\beta} \frac{1}{N+\beta+ \nu_\text{dec}^t}(N\bar{z}_{t+1},N\bar x +\nu^t_\text{dec} b^t_\text{dec}) $ \; \label{ln:psiN2}
$\Psi_t \gets \texttt{cat}(\psi^r_t)$ ; \quad
$\Phi_t \gets \texttt{cat}(\phi^r_t)$ \Comment*[r]{Concatenates the $\psi^r_t$ and $\phi^r_t$ into a $d \times (N+2)$ matrix and a $n \times (N+2)$ 
matrix, respectively}
$W_\text{dec}^{t+1} \gets \begin{pmatrix}
        \Phi_t & \sqrt{\mu^t_\text{dec}}W^t_\text{dec} & 0 _{n\times d}\end{pmatrix}\begin{pmatrix}
        \Psi_t & \sqrt{\mu^t_\text{dec}} I_d & \sqrt{\alpha} I_d
    \end{pmatrix}^\dagger $ \; %\Comment*[r]{Computed by solving a least squares}
$b^{t+1}_\text{dec} \gets  \frac{\nu^t_\text{dec}}{N+\beta +\nu^t_\text{dec}}b_\text{dec}^t  + \frac{N}{N+\beta+\nu^t_\text{dec}}(\bar x - W^{t+1}_\text{dec} \bar z_{t+1})$ \Comment*[r]{Optimal $W^{t+1}_\text{dec}$ and $b^{t+1}_\text{dec}$, see \cref{thm:optdecoding2}}
}
\end{algorithm}
\vspace{-1em}
\end{tcolorbox}

\subsection{Sample-efficient sparse coding for MNIST and LLMs: PAM-SGD vs. SGD}

We performed two experiments comparing the benefits of %structured sparse coding in simple visual domains, we compared standard stochastic gradient descent (SGD) with 
our PAM-SGD method (\cref{alg:PAMSAE}) vs. standard SGD (using Adam for SGD in both cases) for training SAEs: (i) on simple visual domains, using the MNIST dataset, and (ii) on high-dimensional LLM activations, using Google DeepMind's Gemma-2-2B. %Unlike large language models (LLMs), MNIST offers low-dimensional, structured data with interpretable visual patterns, making it ideal for understanding the dynamics of sparse representation learning.

\begin{figure}[htbp]
    \centering
    \includegraphics[width=0.95\linewidth,clip,trim ={0 0 0 0}]{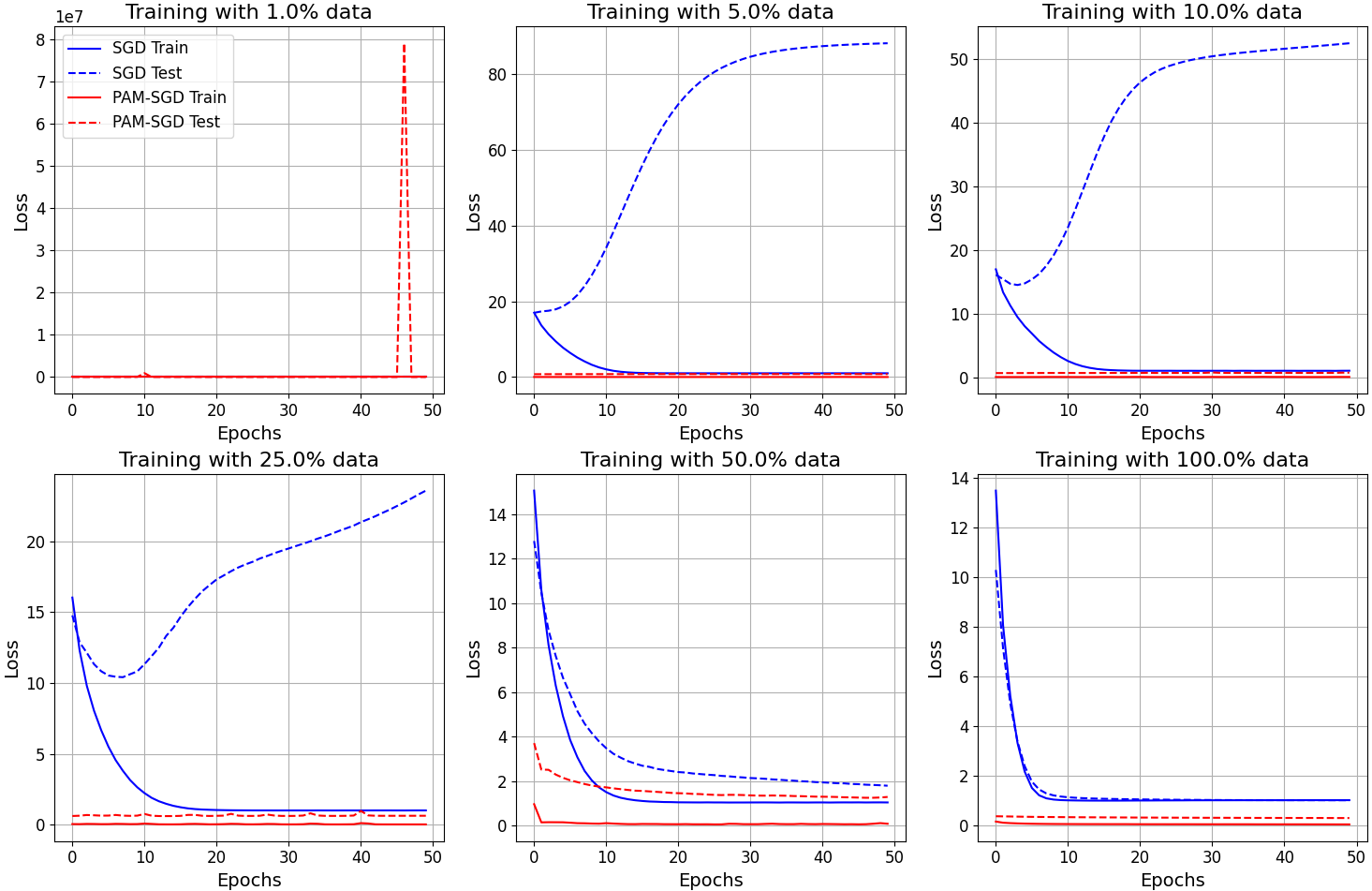}
    \caption{\textbf{Training and test loss curves at different data sizes for MNIST, with ReLU activation.} The chart highlights PAM-SGD's superior sample efficiency and convergence speed.}
    \label{fig:mnist02}
\end{figure}

For MNIST we used $n=28^2$ input dimensions and $d=256$ hidden dimensions. 
Our LLM experiments used activations from the 12th layer of Gemma-2-2B, with $n=2304$ and a highly overcomplete hidden dimension $d =4096$.
%We systematically varied the sparsity parameter $K$ and the number of training samples, evaluating both reconstruction loss and generalization to held-out data. 
Experimental settings are described in \cref{app:setups}, and additional figures and ablation studies can be found in \cref{app:figures}.%PAM-SGD used a decoupled optimization scheme: the encoder was trained via SGD, while the decoder was solved analytically using pseudo-inverse updates.

\begin{figure}[htbp]
    \centering
\includegraphics[width=0.95\linewidth]{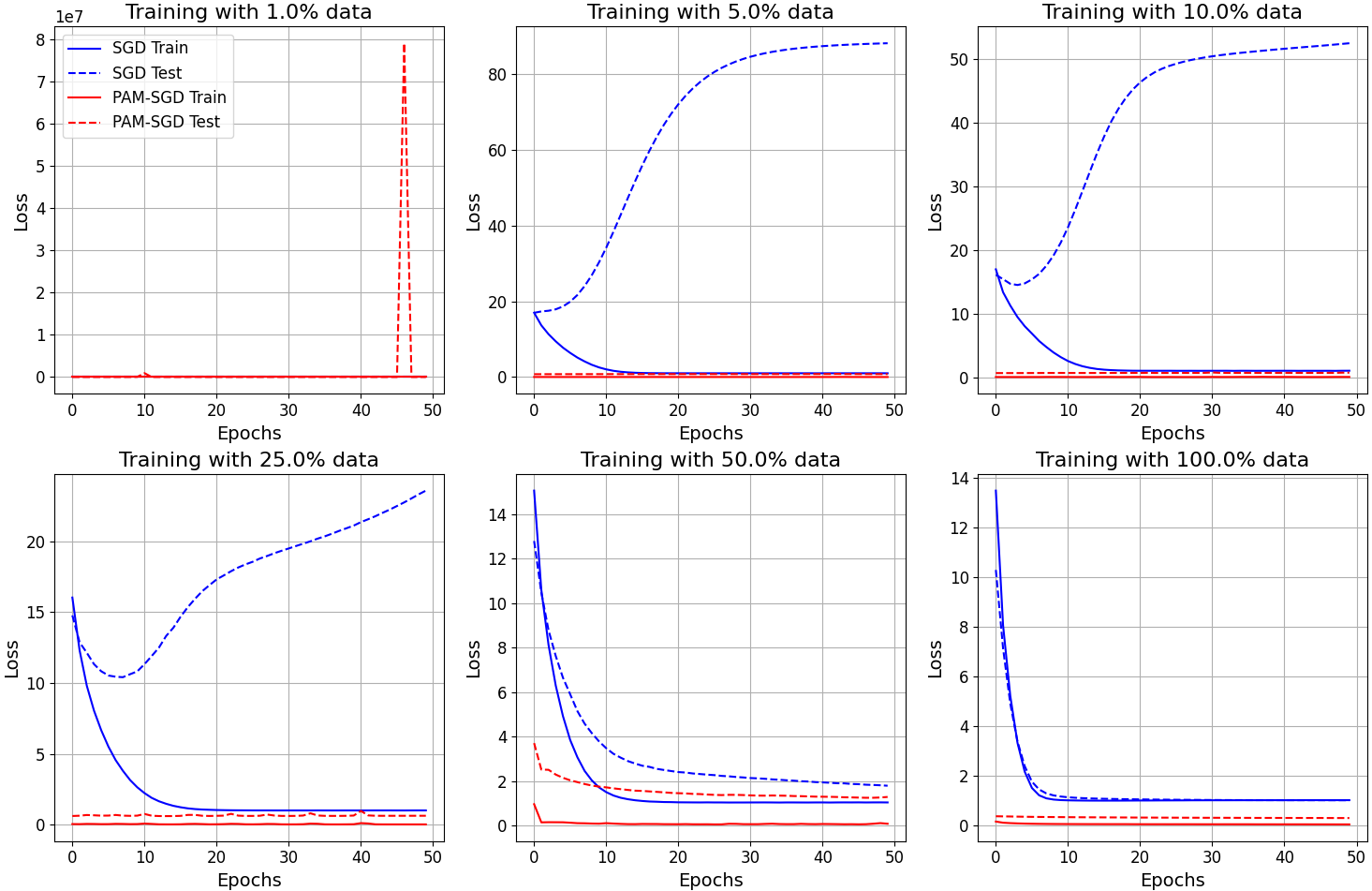}
    \caption{\textbf{Training and test loss curves at different data sizes for Gemma-2-2B, with ReLU activation.} PAM-SGD again has a huge advantage at low data, and remains superior throughout.
    }
    \label{fig:gemma_training_size}
\end{figure}

\paragraph{PAM-SGD generalised better than SGD in low-data regimes. (\cref{fig:mnist02,fig:gemma_training_size,fig:mnistTopK,fig:ReLU_reruns,fig:gemma_K_values,fig:gemma_training_size_Top320,fig:gemma_loss_curves_Top640,fig:gemma_loss_curves_Top320,fig:gemma_training_size_Top640})}
On MNIST, PAM-SGD consistently (using both ReLU and TopK)  substantially outperformed SGD in test loss, especially when trained on just 1\%–25\% of the MNIST training data.
In the LLM experiments, PAM-SGD again outperformed SGD when using ReLU activations, especially for low data. However, PAM-SGD became unstable when TopK was used unless $K$ was large; even for $K=320$ and $K=640$ it underperformed SGD, slightly for low (and high) data and substantially for medium data. 
\paragraph{PAM-SGD was faster, more accurate, and more interpretable on MNIST. (\cref{fig:mnist_reconstruction,fig:mnist_evolution,fig:mnist_filters,fig:mnist_reconstruction_TopK,fig:mnist_evolution_TopK,fig:mnist_filters_TopK})}
Reconstruction comparisons over training epochs show that PAM-SGD reconstructions were cleaner and converged faster than those from SGD. %especially in early training stages. 
This was particularly evident when tracking digit reconstructions over time; PAM-SGD's exhibited sharper edges and more localised structure by early training.
By the end of training, both methods produced visually accurate reconstructions, but PAM-SGD's showed slightly better fidelity and smoothness, and more closely resembled the originals. %—especially under tight sparsity constraints. 
Finally, visualizing encoder and decoder filters in the TopK case reveals that both SGD and PAM-SGD learned edge- and stroke-like patterns, but PAM-SGD's filters were sharper and better structured. %likely due to the explicit decoupling and closed-form decoder updates.
\paragraph{Summary and practical implications.}
PAM-SGD demonstrated clear advantages over SGD on MNIST in terms of generalisation, convergence speed, reconstruction quality, and (using TopK) visual interpretability, particularly in low-data regimes. %Its performance is less sensitive to optimizer hyperparameters and robust across training set sizes. 
Even in a much more challenging real-world LLM setting, %complex LLM representations, 
PAM-SGD with ReLU still substantially outperformed in low data, and improved activation sparsity by about 15\%.  However, issues arose for TopK: small $K$ led to rapidly diverging test loss, and for larger $K$ PAM-SGD still underperformed SGD (though only slightly for low data). %PAM-SGD was furthermore inferior to SGD when 100\% of the training data was used.
In summary, these results suggest that PAM-SGD is a powerful tool for learning overcomplete, sparse representations from visual data and LLM activations in low-data regimes, provided that the sparsity can adapt to the data. This insight is important in downstream applications where data may be scarce.

\section{Conclusions and limitations}

In this work, we have sought to apply a spline theoretical lens to SAEs, to gain insight into how, why, and whether SAEs work. Given the current prominence of SAEs in mechanistic interpretability, and the societal importance of interpreting AI systems, we hope that the development of SAE theory (and our small contribution to it) can help develop more efficient, fairer, and reliable AI systems.  

Building on the piecewise affine spline nature of SAEs, we characterised the spline geometry of TopK SAEs as exactly the $K^\text{th}$-order power diagrams, opening the door to directly incorporating geometric constraints into SAEs. 
We linked SAEs with traditional ML, showing how $k$-means can be viewed as a special kind of SAE, and how SAEs sacrifice  accuracy for interpretability vs. the optimal piecewise affine autoencoder, which we showed to be a $k$-means-esque clustering with a local PCA correction.
% \textit{We have indicated how SAEs bridge between a ``$k$-means autoencoder'' and a ``local PCA autoencoder'', . }
Finally, we developed a new proximal alternating training method (PAM-SGD) for SAEs, with both solid theoretical foundations and promising empirical results, particularly in sample efficiency and activation sparsity for LLMs, two pain points for mechanistic interpretability. PAM-SGD's separate updating of encoding and decoding dovetails well with the spline theory perspective of the encoding shaping the SAE's spline geometry vs. the decoding driving the SAE's autoencoding accuracy.  

This work is the beginning of a longer theoretical exploration, and is thus limited in ways we hope to address in future work. Our characterisation of the spline geometry of SAEs is currently limited to TopK SAEs; future work will seek to extend this, and explore more explicitly the incorporation of geometry into SAE training, perhaps giving insight into how to tailor an SAE architecture for a given task. Our bridge between SAEs and PCA-based autoencoders sets aside the matter (see \cref{nb:GvsSAE}) of shared decoding vectors. %A solution for the optimal parameters akin to \cref{thm:generalPA} can be derived in this setting, but it is unclear how to interpret it; 
Future work will study the optimal autoencoding in that setting, and related results in the superposition hypothesis setting. Finally, PAM-SGD makes approximations which break assumptions of the theory, and had some empirical limitations. Future work will seek to understand more deeply the pros and cons of PAM-SGD, and incorporate the SGD step into the theory.

\clearpage

\begin{ack}
%%% Template acknowledgements

This collaboration did not form in the typical academic way, meeting at a conference or university. 
We instead thank the Machine Learning Street Talk (MLST) team, especially Tim Scarfe, for enabling all the authors to have met through the MLST Discord server. And we thank all the MLST Discord users involved in the discussion on ``Learning in high dimension always amounts to extrapolation'', which set all this in motion.

JB received financial support from start-up funds at the University of Birmingham. BMR received financial support from Taighde \'Eireann – Research Ireland under Grant number [12/RC/2289\_P2]. We declare no conflicts of interest. 
\end{ack}

%\section*{References}

\bibliography{refs}
\bibliographystyle{icml2024}

%%%%%%%%%%%%%%%%%%%%%%%%%%%%%%%%%%%%%%%%%%%%%%%%%%%%%%%%%%%%

\appendix

\section{Visualisations of Voronoi and power diagrams}
\label{app:Voronoi}

\subsection{Spatial Partitioning Methods: From Voronoi to Power Diagrams}
This experiment explores the theoretical connections between different spatial partitioning methods. Starting with standard Voronoi diagrams that divide the plane based on proximity to generator points, we demonstrate how they relate to nearest-neighbor classification ($k=1$), centroidal clustering ($k$-means), and finally to power diagrams which introduce weights to Voronoi cells. These relationships reveal that power diagrams emerge as a generalization of Voronoi diagrams, offering additional flexibility through weighted distance metrics and enabling richer geometric representations of data.
\begin{figure}[htbp]
    \centering
    \includegraphics[width=1\linewidth]{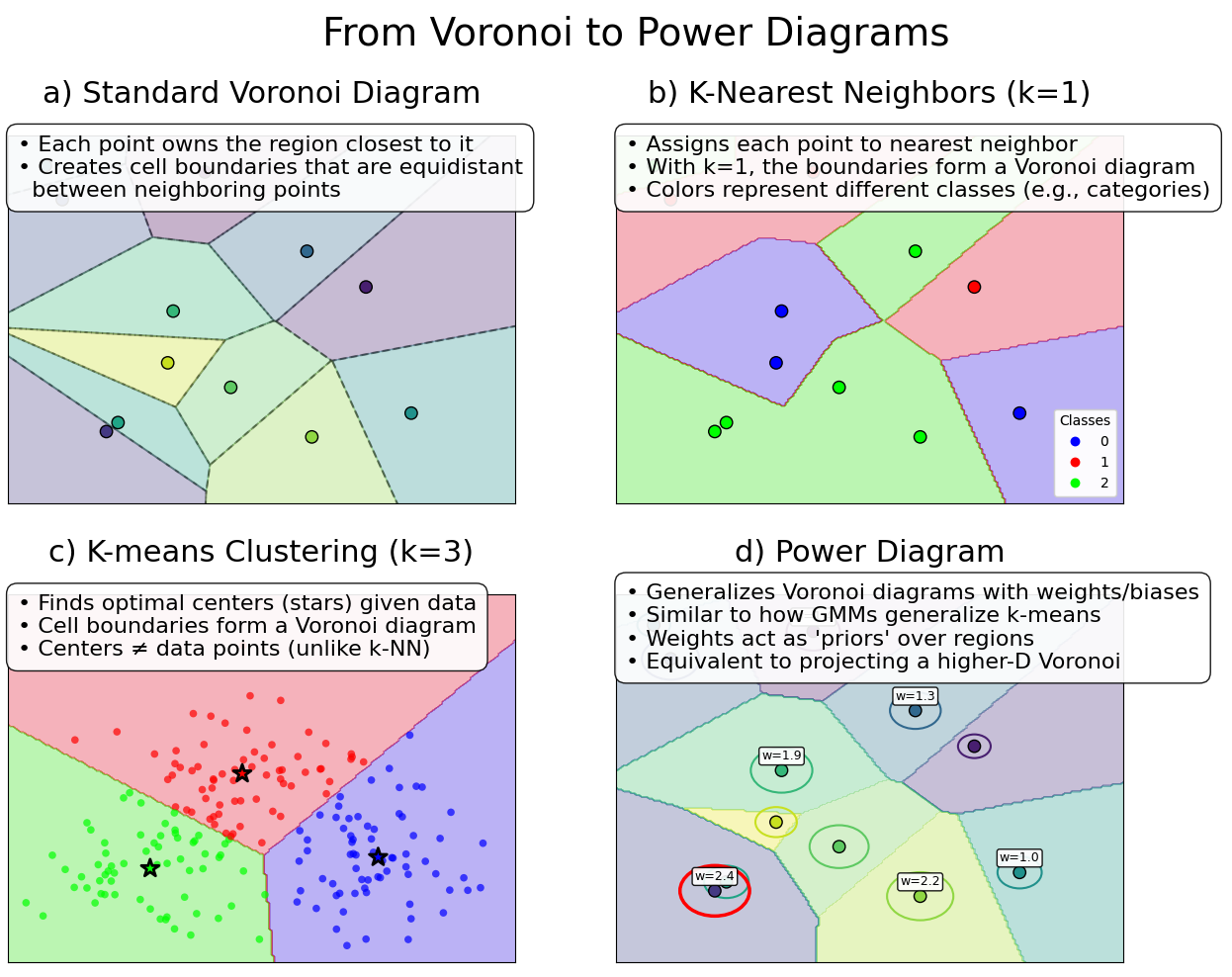}
    \caption{Spatial Partitioning Methods (a) Standard Voronoi diagram where space is partitioned based on the nearest generator point using Euclidean distance; (b) Nearest-neighbor ($k=1$) classification showing how Voronoi cells define decision boundaries for point classification (c) $k$-means clustering with $k=3$ demonstrating how cluster centroids generate Voronoi cells that define cluster boundaries (d) Power diagram (weighted Voronoi) where each generator has an associated weight, creating curved boundaries between regions. Power diagrams generalise Voronoi diagrams and provide additional flexibility for modeling spatial relationships.}
    \label{fig:enter-label}
\end{figure}

\begin{figure}[htbp]
    \centering
    \begin{subfigure}[b]{\linewidth}
        \centering
        \includegraphics[width=0.95\textwidth]{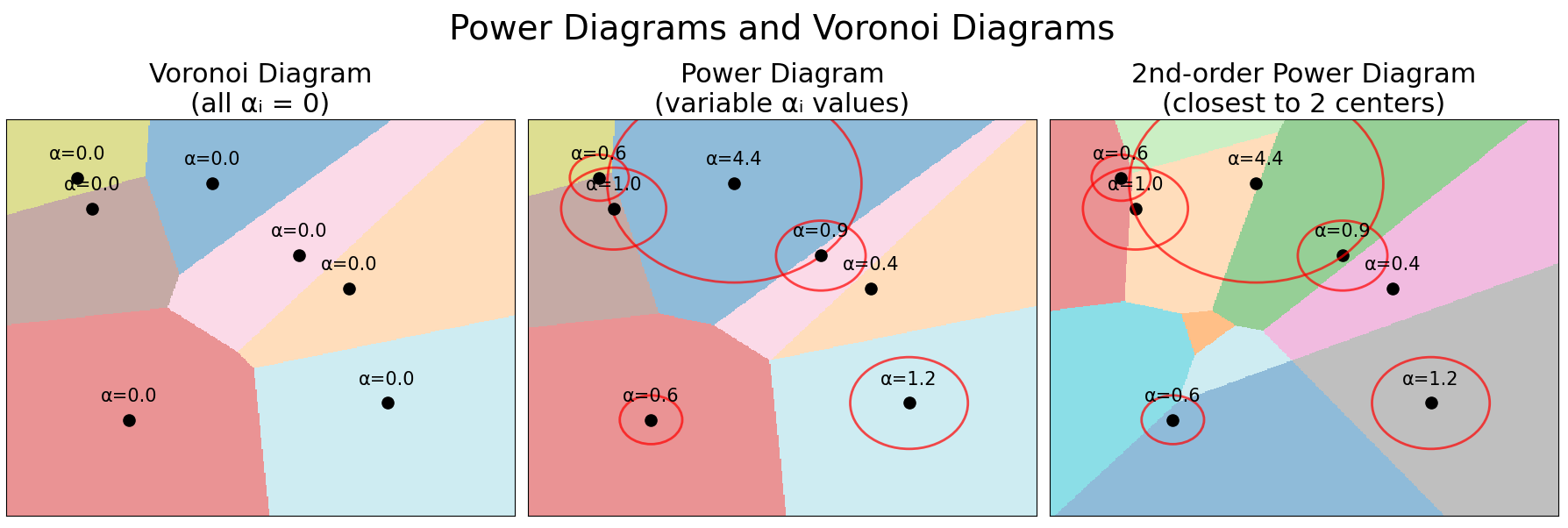}
        \caption{}
        \label{fig:power-diagram1}
    \end{subfigure}
    
    \vspace{1cm}  % Space between images
    
    \begin{subfigure}[b]{\linewidth}
        \centering
        \includegraphics[width=0.95\textwidth]{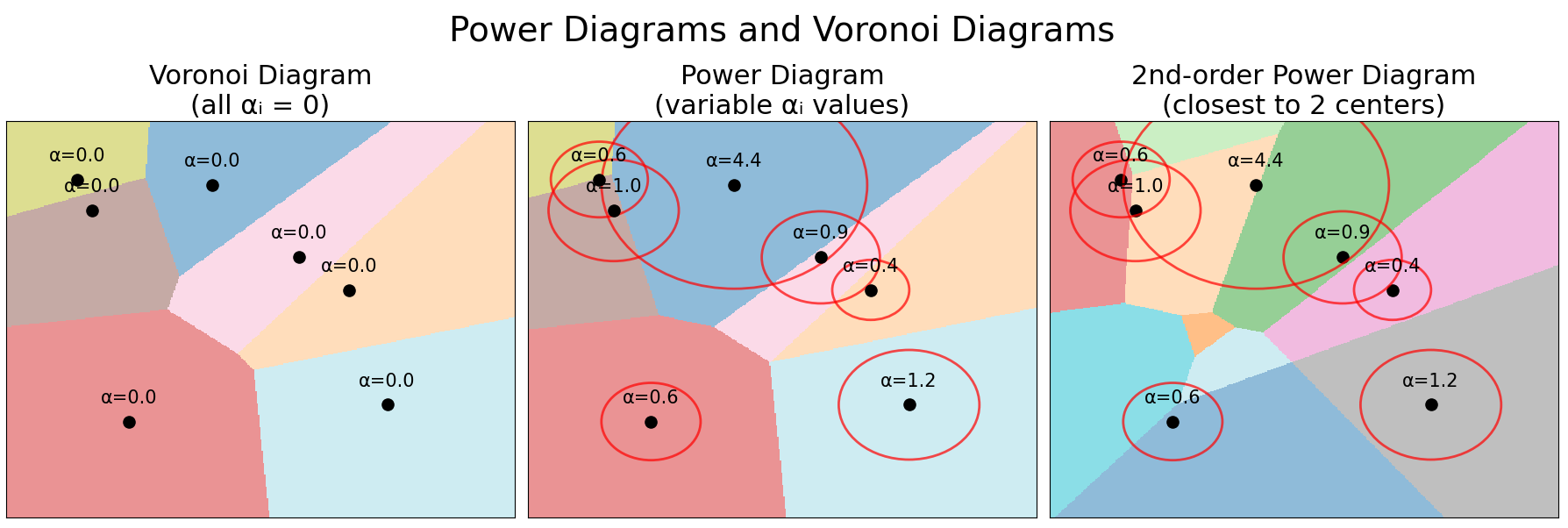}
        \caption{}
        \label{fig:power-diagram2}
    \end{subfigure}
    
    \caption{Progressive Generalization of Voronoi Diagrams. Comparison showing (left) standard Voronoi diagram, (center) first-order power diagram with linear distance weighting, and (right) second-order power diagram. This progression demonstrates how higher-order power diagrams can capture more sophisticated spatial relationships.}
    \label{fig:power-diagram-progression}
\end{figure}

\clearpage

\subsection{3D Voronoi diagrams to 2D power diagrams}

\begin{figure}[htbp]
    \centering
    \begin{subfigure}[b]{0.48\textwidth}
        \centering
        \includegraphics[width=\textwidth]{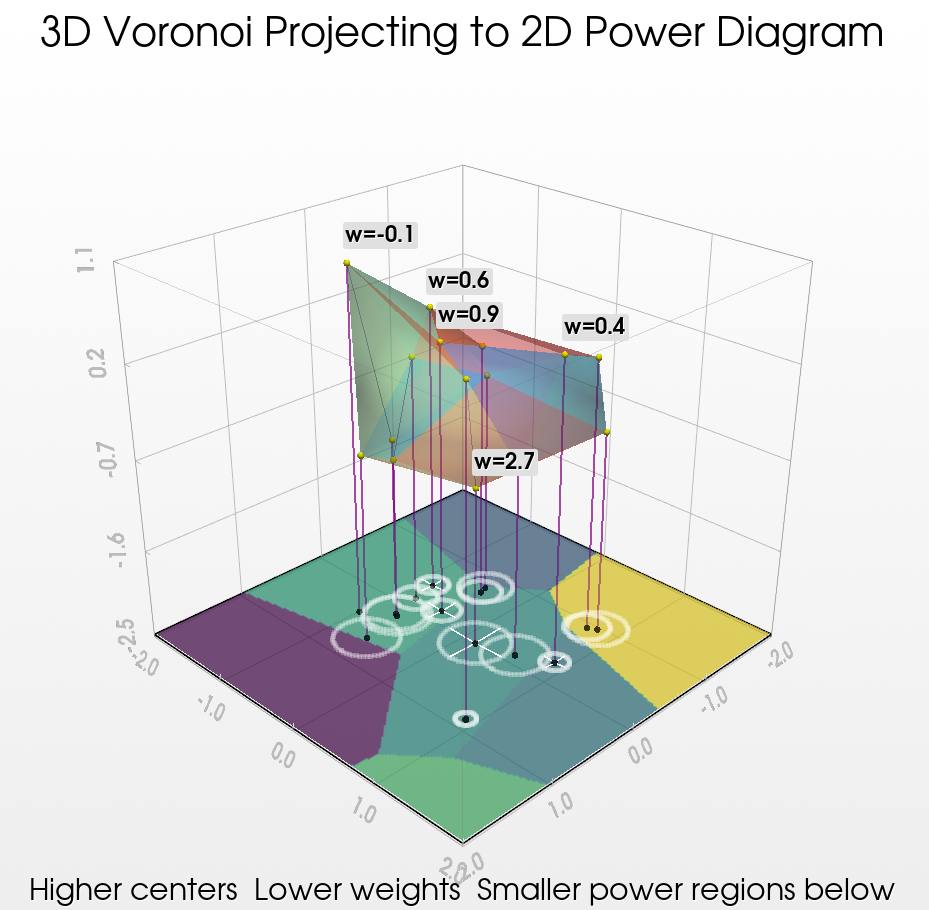}
        \caption{}
        \label{fig:voronoi1}
    \end{subfigure}
    \hfill
    \begin{subfigure}[b]{0.48\textwidth}
        \centering
        \includegraphics[width=\textwidth]{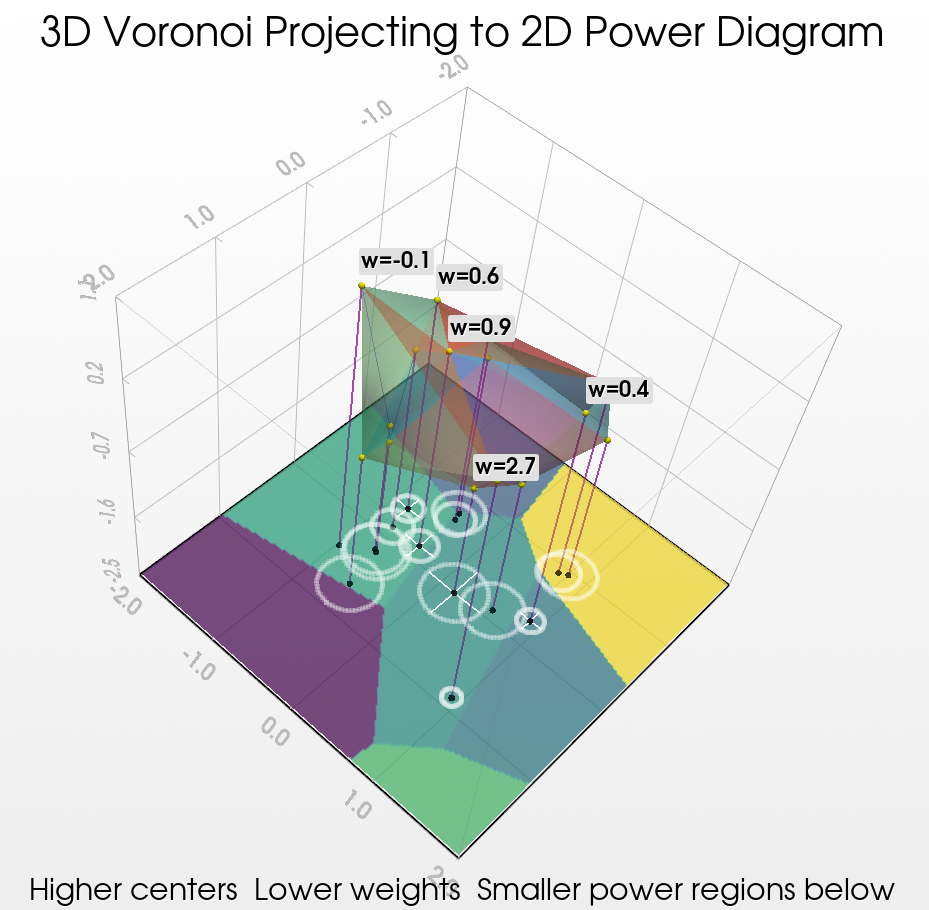}
        \caption{}
        \label{fig:voronoi2}
    \end{subfigure}
    \vskip\baselineskip
    \begin{subfigure}[b]{0.48\textwidth}
        \centering
        \includegraphics[width=\textwidth]{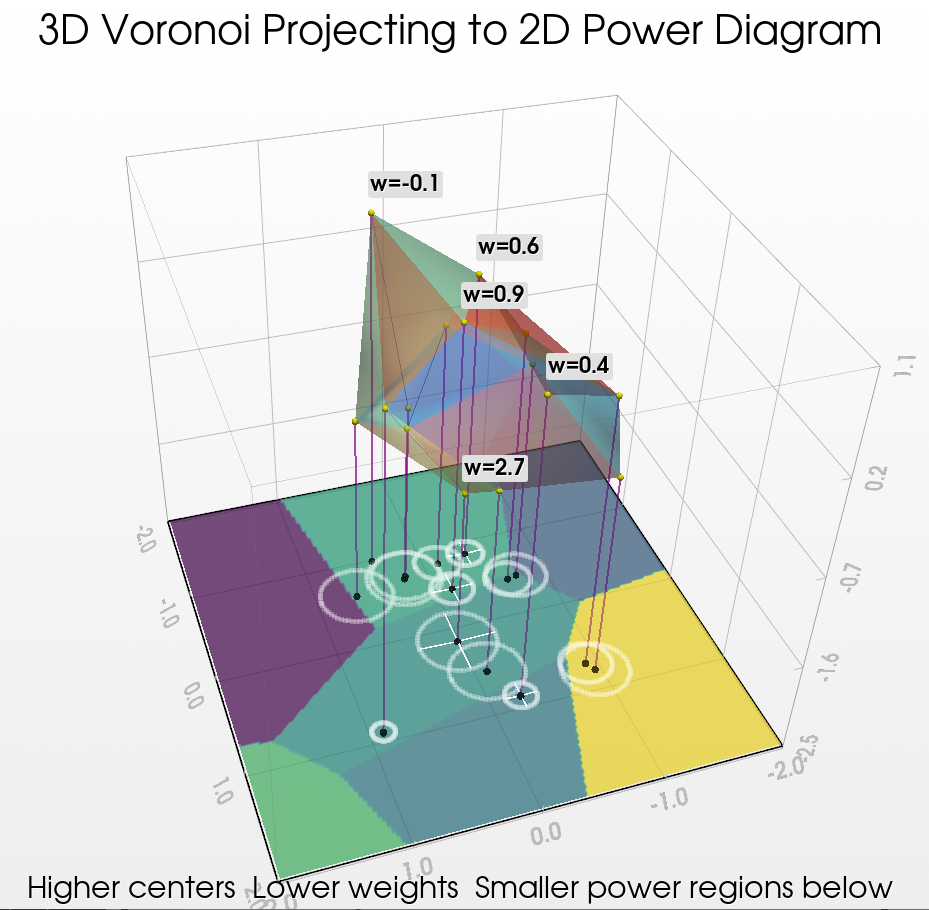}
        \caption{}
        \label{fig:voronoi3}
    \end{subfigure}
    \hfill
    \begin{subfigure}[b]{0.48\textwidth}
        \centering
        \includegraphics[width=\textwidth]{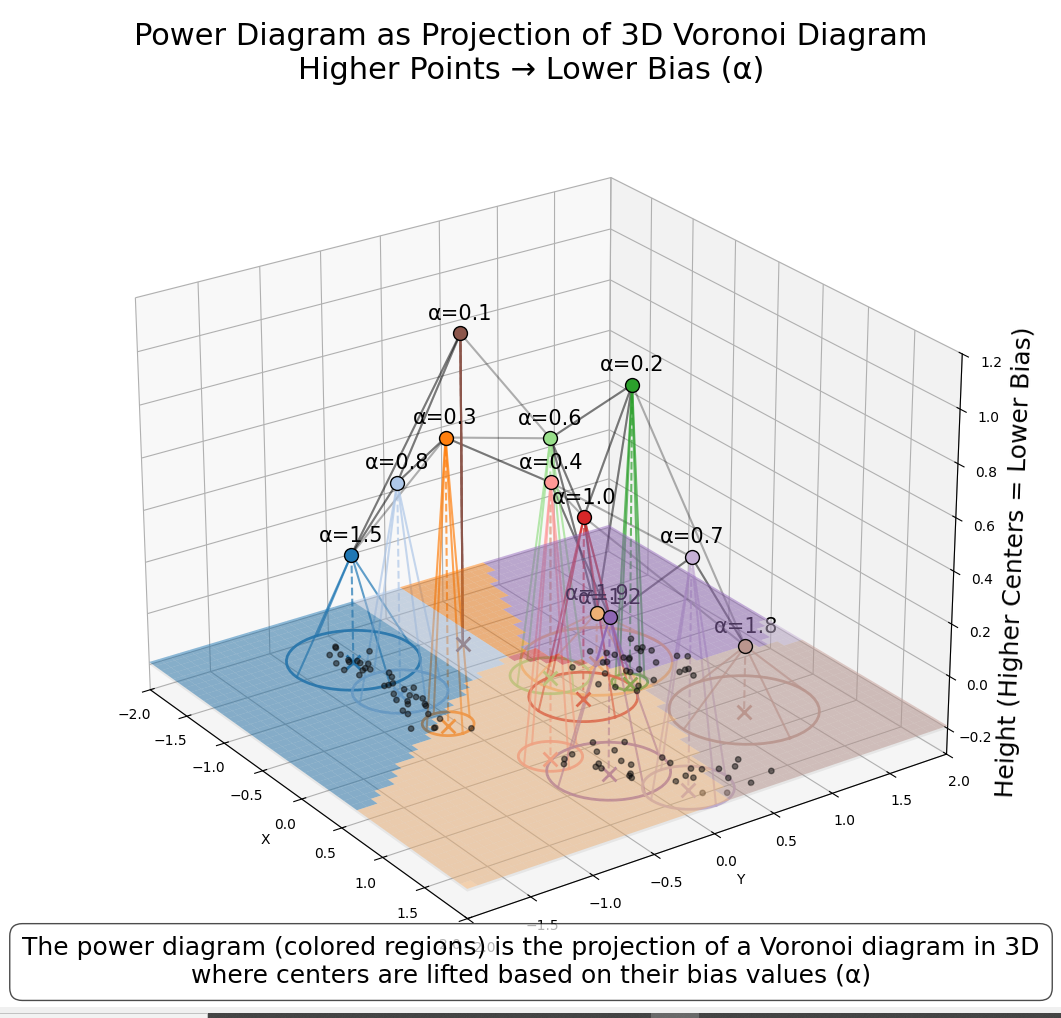}
        \caption{}
        \label{fig:voronoi4}
    \end{subfigure}
    \caption{Projection of 3D Voronoi diagrams onto 2D power diagrams. The resulting power cells demonstrate the preservation of topological structure through projection}
    \label{fig:voronoi_projections}
\end{figure}

There is a neat relationship between power diagrams and projections of Voronoi diagrams, observed in \cite{Ash1986}. Suppose that we consider a Vononoi diagram in $n+1$ dimensions with cells defined by centroids $\{(\mu_i,\zeta_i)\}_{i=1}^k \in \RR^{n+1}$, i.e. the  $i^\text{th}$ cell is defined to be
    \[
    C_i := \{(x,z) \in \RR^{n+1} : \|(x,z) - (\mu_i,\zeta_i)\|_2^2 <  \|(x,z) - (\mu_j,\zeta_j)\|_2^2 \text{ for all } j\neq i \}. 
    \]
Now suppose that we project this diagram into $n$-dimensional space, for example defining
\[
\hat C_i := \{ x \in \RR^n : (x,0) \in C_i \} = \{ x \in \RR^n :  \|x -\mu_i\|_2^2 + \zeta_i^2  <  \|x -\mu_j\|_2^2 + \zeta_j^2 \text{ for all } j\neq i \}.
\]
This is precisely a power diagram with centroids $\{\mu_i\}_{i=1}^k \in \RR^{n}$ (the projections of the Voronoi centroids) and weights $\alpha_i=-\zeta_i^2$. Conversely, for any power diagram defined by centroids $\{\mu_i\}_{i=1}^k \in \RR^n$ and weights $\{\alpha_i\}_{i=1}^k \in \RR$, we can subtract a constant from the $\alpha_i$ to get an equivalent power diagram with all non-positive weights, and therefore compute $\zeta_i$ such that the corresponding Voronoi diagram with centroids $\{(\mu_i,\zeta_i)\}_{i=1}^k \in \RR^{n+1}$ projects to that power diagram. We visualise this mathematical relationship between 3D Voronoi diagrams and 2D power diagrams in \cref{fig:voronoi_projections}. %through projection %This equivalence arises when 3D Voronoi cells are projected onto the plane, resulting in power diagrams (weighted Voronoi diagrams) in 2D. 

\clearpage
\section{Proofs in \cref{sec:splinegeometry}}
\label{app:splineproofs}

\begin{tcolorbox}[colback=IBMgold!20,colframe=IBMgold] 
\begin{theorem}\label{thm:openpoly}
Both $\Omega_S^{\JumpReLU}$ and $ \Omega_S^{\TopK}$ can be written in the form 
\[
\{ x\in \RR^n : Hx > c\},
\]
where in the in the former $H= (P_S - P_{S^c})W_\text{enc}$ and $c =  ( P_S - P_{S^c})(\tau\mathbf{1}-b_\text{enc})$, and in the latter $H= M_S W_{enc}$ where $M_{S}\in \RR^{K(d-K)\times d}$ has $(i,j)$-th row (with $i \in S$ and $j \notin S$) with a $1$ in column $i$, a $-1$ in column $j$, and $0$ otherwise, and $c = -M_S b_\text{enc}$. These regions are therefore convex, and are the interiors of the convex polyhedra $\{x\in \RR^n:Hx \geq c\}$ (unless $W_\text{enc}$ has a zero row, in the JumpReLU case, or has two identical rows, in the TopK case). 
\end{theorem}
\end{tcolorbox}
\begin{proof}[Proof of \cref{thm:openpoly}]
    The forms of $H$ and $c$ can be immediately derived by rearranging the inequalities in the definitions of $\Omega^{\JumpReLU}_S$ and $\Omega_S^{\TopK}$. Convexity immediately follows, since if $Hx_1>c$ and $Hx_2 >c$, then for all $t \in [0,1]$
    \[
    H(tx_1 + (1-t)x_2) = tHx_1 + (1-t)Hx_2 > tc + (1-t)c = c.
    \]
    Finally, suppose that $x\in \RR^n$ lies in the interior of $\{x :Hx\geq c\}$. That is, there exists $\varepsilon>0$ such thatr for all $\eta \in \RR^n$ with $\|\eta \|_2 < \varepsilon$, $H(x + \eta) \geq c$. We wish to show that $Hx > c$. 
    
    Suppose not, then for some $j$, $(Hx)_j = c_j$. Therefore for all $\eta \in \RR^n$ with $\|\eta \|_2 < \varepsilon$, $(H\eta)_j \geq 0$, and therefore for all $\eta \in \RR^n$, $(H\eta)_j =0$. This is possible if and only if the $j^\text{th}$ row of $H$ is all zeroes. In the JumpReLU case, the $j^\text{th}$ row of $H$ is $\pm$ the $j^\text{th}$ row of $W_\text{enc}$ (depending on if $j \in S$ or $j \notin S$) and hence is zero if and only if the the $j^\text{th}$ row of $W_\text{enc}$ is zero. In the TopK case, the $(i,j)^\text{th}$ row of $H$ is the difference between the $i^\text{th}$ and $j^\text{th}$ rows of $W_\text{enc}$, which is zero if and only if those rows are identical. 
\end{proof}
\begin{tcolorbox}[colback=IBMgold!20,colframe=IBMgold] 
\begin{theorem}\label{thm:Kthpowerdiagram}
    Let $\{\mu_i\}_{i=1}^k \in \RR^n$ and  $\{\alpha_i\}_{i=1}^k \in \RR$ define a $K^\text{th}$-order power diagram $\{C_S\}$ for $S$ the $K$-subsets of $\{1,...,k\}$. Then the power diagram given by 
    \begin{align}\label{eq:nubeta}
        \nu_S := \frac1K \sum_{i \in S} \mu_i && \text{ and } && \beta_S := \left\|  \frac1K \sum_{i \in S} \mu_i  \right\|_2^2 - \frac1K \sum_{i \in S} \|\mu_i\|_2^2 + \frac1K \sum_{i \in S} \alpha_i ,    \end{align}
        i.e.,
        \[
        R_S:= \{ x: \|x - \nu_S\|^2_2 - \beta_S < \|x - \nu_T\|^2_2 - \beta_T \text{ for all $T \neq S$, $T$ a $K$-subset of $\{1,...,k\}$}\}, 
        \]
        satisfies $R_S = C_S$ for all $S$. 
\end{theorem}
\end{tcolorbox}
\begin{proof}[Proof of \cref{thm:Kthpowerdiagram}]
    Define the power functions 
    \begin{align*}
        P_i(x):= -2\mu_i^Tx + \|\mu_i\|^2_2 - \alpha_i && \text{and} &&
          Q_S(x):= -2\nu_S^Tx + \|\nu_S\|^2_2 - \beta_S.
    \end{align*}
    Then by subtracting $\|x\|_2^2$ from both sides of the defining inequalities we get:
    \begin{align*}
       C_S &= \{ x : P_i(x) < P_j(x) \text{ for all $i \in S$, and $j \in S^c$}\},\\
         R_S &= \{ x : Q_S(x) < Q_T(x) \text{ for all $T\neq S$, $T$ a $K$-subset of $\{1,...,k\}$}\}.
    \end{align*}

It is straightforward to check that 
\[
Q_S(x) = \frac1K \sum_{i\in S} P_i(x),
\]
and hence 
\[
x \in R_S \text{ if and only if } \sum_{i \in S} P_i(x) < \sum_{i \in T} P_i(x) \text{ for all $T \neq S$ a $K$-subset}.
\]
Suppose that $x \in R_S$, and let $i \in S$ and $j \in S^c$. Let $T = (S \setminus\{i\}) \cup \{j\}$. This is a $K$-subset distinct from $S$, and so 
\[
P_i(x) + \sum_{k \in S\setminus\{i\}} P_k(x) = \sum_{i \in S} P_i(x) < \sum_{i \in T} P_i(x) = P_j(x) + \sum_{k \in S\setminus\{i\}} P_k(x)
\]
and hence $P_i(x) <P_j(x)$. Hence $x \in C_S$. 

Now suppose that $x \in C_S$ and let $T\neq S$ be a $K$-subset. Then 
\[
\sum_{i \in S} P_i(x) = \sum_{i \in S\cap T}  P_i (x) + \sum_{i \in S \setminus T} P_i(x) < \sum_{i \in S\cap T}  P_i (x) + \sum_{j \in T \setminus S} P_j(x) = \sum_{i \in T} P_i(x)
\]
where we have used that $|S\setminus T| = |T\setminus S|$ and for each $i \in S\setminus T$ and $j \in T\setminus S$, $P_i(x) <P_j(x)$. Hence $ x \in R_S$. 
\end{proof}
 \begin{proof}[Proof of \cref{nb:hexagon}]
        Identify $\RR^2$ with $\mathbb{C}$.    Then we would need $\{\mu_i\}_{i=1}^4 \in  \mathbb{C}$ such that 
    \begin{align*}
        \nu_{12} = 1 &= \frac12 \mu_1 + \frac12 \mu_2, &
        \nu_{13} = e^{\pi i/3} &= \frac12 \mu_1 + \frac12 \mu_3,\\
        \nu_{14} = e^{2\pi i/3} &= \frac12 \mu_1 + \frac12 \mu_4, &
        \nu_{23} = -1 &= \frac12 \mu_2 + \frac12 \mu_3,\\
        \nu_{24} = -e^{\pi i/3} &= \frac12 \mu_2 + \frac12 \mu_4, &
        \nu_{34} = - e^{2\pi i/3} &= \frac12 \mu_3 + \frac12 \mu_4.
    \end{align*}
    Hence $\mu_2 - \mu_4 = (\frac12 \mu_1 + \frac12 \mu_2) + (\frac12 \mu_2 +\frac12 \mu_3) -(\frac12 \mu_1 + \frac12\mu_4) -(\frac12 \mu_3 + \frac12\mu_4) = 1 + (-1) - e^{2\pi i/3} - (-e^{2\pi i/3}) = 0$, so $\mu_2 = \mu_4$ and hence $1 = \frac12 \mu_1 + \frac12 \mu_2 = \frac12 \mu_1 + \frac12\mu_4  = e^{2\pi i/3}$, a contradiction. 
    \end{proof}
\begin{proof}[Proof of \cref{thm:powerdiagram}]
    For each $S$,  $\Omega^{\TopK}_S$ is given by 
    \[
    \Omega^{\TopK}_S = \{ x \in \RR^n : e_i^TW_{enc}x + (b_{enc})_i > e_j^TW_{enc}x + (b_{enc})_j \text{ for all $i \in S$ and $j \in S^c $} \}.
    \]
    We can rewrite the definition of $C_S$ in a $K^\text{th}$-order power diagram as 
    \[
    C_S := \{x \in \RR^n : \mu_i^T x + \frac12\alpha_i - \frac12\|\mu_i\|_2^2 > \mu_j^T x + \frac12\alpha_j - \frac12\|\mu_j\|_2^2  \text{ for all $i \in S$ and $j \in S^c$} \}.
    \]
It follows that $\Omega^{\TopK}_S = C_S$ if for all $\ell$
\begin{align*}
    e^T_\ell W_{enc}= \mu_\ell^T && \text{and} && (b_{enc})_\ell = \frac12\alpha_\ell - \frac12 \|\mu_\ell \|_2^2, &&\text{i.e.} \\
    \mu_\ell= W^T_{enc}e_\ell && \text{and} && \alpha_\ell = 2(b_{enc})_\ell +  \|W^T_{enc}e_\ell\|_2^2.
\end{align*}
Hence, any $(W_{enc},b_{enc})$ gives rise to a $K^\text{th}$-order power diagram, and any $K^\text{th}$-order power diagram gives rise to an $(W_{enc},b_{enc})$. Finally, \cref{eq:TopKpower2} follows from \cref{eq:nubeta} and \cref{eq:omegatopower}. 
\end{proof}
\begin{proof}[Proof of \cref{thm:generalPA}]
    We seek to minimise 
\[
\cL = \sum_{i=1}^k \lambda N_i K_i +\sum_{x^r \in R_i} \|x^r - U_iV_i^Tx^r - c_i \|_2^2.
\]
For the same reason as for the $\nu_i$ in $k$-means, the $c_i$ will be minimising when 
\[
c_i = \frac{1}{N_i} \sum_{x^r \in R_i} (I-U_iV_i^T)x^r = (I-U_iV_i^T) \bar{x}_i,
\]
 and so $\cL$ simplifies to 
\[
\cL = \sum_{i=1}^k \lambda N_i K_i +\sum_{x^r \in R_i} \|(I- U_iV_i^T)(x^r -\bar{x}_i) \|_2^2.
\]
We claim that this is minimised when $U_i = V_i$ and the columns of $U_i$ are the top $K_i$ eigenvectors of 
\[
X_i := \frac{1}{N_i}\sum_{x^r \in R_i} (x^r - \bar{x}_i) (x^r - \bar{x}_i)^T,
\]
and in this case $\cL$ reduces to 
\begin{equation*} \label{eq:PCAL}
\cL = \sum_{i=1}^k \lambda N_i K_i +\left(\sum_{x^r \in R_i} \|(x^r -\bar{x}_i) \|_2^2 \right) - N_i \sum_{\ell=1}^{K_i} \lambda_{\ell}(X_i),
\end{equation*}
where $ \lambda_{\ell}(X_i)$ are the eigenvalues of $X_i$ in descending order. 

This follows because    \begin{align*}
    \sum_{x^r} \|(I- UV^T)(x^r -\bar{x}) \|_2^2 &= \sum_{x^r} (x^r -\bar{x})^T(I- VU^T)(I- UV^T)(x^r -\bar{x})\\
    &= \sum_{x^r} \operatorname{tr}\left((x^r -\bar{x})^T(I- VU^T)(I- UV^T)(x^r -\bar{x})\right)\\
     &= \sum_{x^r} \operatorname{tr}\left((x^r -\bar{x})(x^r -\bar{x})^T(I- VU^T)(I- UV^T)\right)\\
     &=N \operatorname{tr}\left(X (I- VU^T)(I- UV^T)\right)
     \\&=N \operatorname{tr}\left((X - XVU^T)(I- UV^T)\right)
      \\&= N\operatorname{tr}\left(X - XVU^T - XUV^T + XVV^T \right)
        \\&= N\operatorname{tr}(X)  + N\operatorname{tr}\left(-U^TXV- V^TXU + V^TXV\right)
             \\&= N\operatorname{tr}(X)  + N\operatorname{tr}\left((V-U)^TX(V-U)\right) - N\operatorname{tr}\left(U^TXU\right)
     % \\&= \operatorname{tr}\left(\sum_j \lambda_j \xi_j \xi_j^T (I- vu^T)(I- uv^T)\right)
     %      \\&= \sum_j \lambda_j \|\xi_j - uv^T \xi_j\|^2
    \end{align*}
    is minimised when $U=V$ and $\operatorname{tr}\left(U^TXU\right)$ is maximised (with the constraint that $U^TU=I$). This occurs when $U$ has columns the top $K$ leading eigenvectors of $X$. At this choice:
    \[
         \sum_{x^r} \|(I- UV^T)(x^r -\bar{x}) \|^2 =N\operatorname{tr}(X) - N\sum_{\ell=1}^k\lambda_\ell  =  \sum_{x^r} \operatorname{tr}((x^r -\bar{x})^T(x^r -\bar{x})) - N\sum_{\ell=1}^k\lambda_\ell = \sum_{x^r} \|x^r -\bar{x} \|_2^2  -N\sum_{\ell=1}^k\lambda_\ell.
    \]
\end{proof}

\section{Proofs in \cref{sec:optdecoding}} \label{app:optdecoding}

\begin{tcolorbox}[colback=IBMgold!20,colframe=IBMgold,enhanced]

 \begin{theorem}\label{thm:optdecoding2}

% Suppose that we are given training data $\{x^r\}_{r=1}^N \in \RR^n$. Let $R_S:= \Omega_S$ in the ReLU case and $R_S:=\omega_S$ in the TopK case (with appropriate domains of $S$), and suppose that $x^r \in \cup_S R_S$ for all $r$.  Let 
% \[
% z^r := \begin{cases}
%     P_S(W_\text{enc}x^r + b_\text{enc}), & x^r \in R_S. 
% \end{cases}
% \]
The $(W_\text{dec}^{t+1},b_\text{dec}^{t+1})$ solving \cref{eq:PAMdec} for $G := \alpha\|W_\text{dec}\|^2_F + \beta\|b_\text{dec}\|_2^2$ are given by 
\begin{align*}
    b^{t+1}_\text{dec} &= \frac{\nu^t_\text{dec}}{N+\beta+ \nu^t_\text{dec}}b_\text{dec}^t  + \frac{N}{N+\beta+\nu^t_\text{dec}}(\bar x - W_\text{dec} \bar z_{t+1}),\\
    W^{t+1}_\text{dec} & = \begin{pmatrix}
        \Phi_t & \sqrt{\mu^t_\text{dec}}W^t_\text{dec} & 0_{n \times d}\end{pmatrix}\begin{pmatrix}
        \Psi_t & \sqrt{\mu^t_\text{dec}} I_d & \sqrt{\alpha} I_d
    \end{pmatrix}^\dagger, 
   % \\&= (\Phi_t \Psi_t^T + \mu^t_\text{dec} W^t_\text{dec}) \left(  \Psi_t \Psi_t^T +  (\alpha + \mu^t_\text{dec}) I_d \right)^{-1},
\end{align*}
where $z^r_{t+1} := \rho(W^{t+1}_\text{enc} x^r + b^{t+1}_\text{enc})$, $\bar{z}_{t+1}$ is the mean of the $ z^r_{t+1}$, and $\Psi_t \in \RR^{d \times (N+2)}$ and $\Phi_t \in \RR^{n \times (N+2)}$ are matrices with columns (for $r = 1$ to $N$)
\begin{align*}
 \hspace{-0.75em}\psi_t^r &:= z^r_{t+1} -  \frac{N \bar{z}_{t+1}}{N+\beta+\nu^t_\text{dec}},  &\hspace{-0.75em}&\psi_t^{N+1} := \frac{N \sqrt{\nu_\text{dec}^t} }{N+\beta + \nu_\text{dec}^t} \bar{z}_{t+1}, &\hspace{-0.75em}& \psi_t^{N+2} :=  \frac{N\sqrt{\beta}}{N+\beta+\nu_\text{dec}^t} \bar{z}_{t+1},\\
\hspace{-0.75em}\phi_t^r &:= x^r - \frac{N\bar x+ \nu^t_\text{dec}b_\text{dec}^t  }{N+\beta + \nu^t_\text{dec}},&\hspace{-0.75em}& \phi_t^{N+1}:= \sqrt{\nu_\text{dec}^t} \frac{N \bar x - (N+\beta) b^t_\text{dec}}{N+\beta+\nu_\text{dec}^t} , &\hspace{-0.75em}&\phi_t^{N+2} := \sqrt{\beta}  \frac{N\bar x+ \nu^t_\text{dec}b_\text{dec}^t  }{N+\beta + \nu^t_\text{dec}}.
\end{align*}
\end{theorem}
\end{tcolorbox}
\begin{proof}[Proof of \cref{thm:optdecoding2}]
    By completing the square, the optimal $b_\text{dec}$ is given by 
\[
b_\text{dec} = \frac1{N+\beta + \nu^t_\text{dec}} \left( \nu^t_\text{dec}b_\text{dec}^t  + \sum_{r=1}^t x^r - W_\text{dec} z_{t+1}^r \right) = \frac{\nu^t_\text{dec}}{N+\beta+ \nu^t_\text{dec}}b_\text{dec}^t  + \frac{N}{N+\beta+\nu^t_\text{dec}}(\bar x - W_\text{dec} \bar z_{t+1}),
\]
where 
\begin{align*}
z^r_{t+1} := \rho(W^{t+1}_\text{enc} x^r + b^{t+1}_\text{enc}) && \text{and} && \bar{z}_{t+1}:= \frac1N \sum_{r=1}^N z^r_{t+1}.
\end{align*}

This reduces $\cL$ to:
\begin{align*}
\sum_{r=1}^N \left\| W_\text{dec} \left(z^r_{t+1} - \frac{N}{N+\beta + \nu^t_\text{dec}} \bar{z}_{t+1}\right)+ \frac{\nu^t_\text{dec}}{N+\beta + \nu^t_\text{dec}}b_\text{dec}^t  + \frac{N}{N+\beta + \nu^t_\text{dec}}\bar x  - x^r\right\|_2^2  \\ + \mu_\text{dec}^t \|W_\text{dec} - W_\text{dec}^t\|_F^2  + \alpha\|W_\text{dec}\|^2_F
+ \nu^t_\text{dec} \left \| \frac{N}{N+\beta+\nu^t_\text{dec}}(\bar x - W_\text{dec} \bar z_{t+1})- \frac{N+\beta}{N+\beta+ \nu^t_\text{dec}} b_\text{dec}^t  \right\|_2^2 \\ + \beta \left \| \frac{\nu^t_\text{dec}}{N+\beta+ \nu^t_\text{dec}}b_\text{dec}^t  + \frac{N}{N+\beta+\nu^t_\text{dec}}(\bar x - W_\text{dec} \bar z_{t+1})\right\|_2^2 .
\end{align*}

Hence 
\begin{align*}
    \cL &= \sum_{r=1}^{N+2} \|W_\text{dec} \psi_t^r -\phi_t^r\|^2_2 +    \mu_\text{dec}^t \|W_\text{dec} - W_\text{dec}^t\|_F^2 + \alpha\|W_\text{dec}\|^2_F \\
    &= \|W_\text{dec} \Psi_t - \Phi_t\|_F^2 + \mu_\text{dec}^t \|W_\text{dec} - W_\text{dec}^t\|_F^2 + \alpha\|W_\text{dec}\|^2_F \\
    &= \left\|W_\text{dec} \begin{pmatrix}
        \Psi_t & \sqrt{\mu^t_\text{dec}} I_d & \sqrt{\alpha} I_d
    \end{pmatrix} -  \begin{pmatrix}
        \Phi_t & \sqrt{\mu^t_\text{dec}}W^t_\text{dec} & 0_{n \times d} \end{pmatrix}\right \|_F^2
\end{align*}
and so 
\begin{align*}
    W_\text{dec} & = \begin{pmatrix}
        \Phi_t & \sqrt{\mu^t_\text{dec}}W^t_\text{dec} & 0_{n \times d}\end{pmatrix}\begin{pmatrix}
        \Psi_t & \sqrt{\mu^t_\text{dec}} I_d & \sqrt{\alpha} I_d
    \end{pmatrix}^\dagger 
    \\&= (\Phi_t \Psi_t^T + \mu^t_\text{dec} W^t_\text{dec}) \left(  \Psi_t \Psi_t^T +  (\alpha + \mu^t_\text{dec}) I_d \right)^{-1} . 
\end{align*}
\end{proof}
\subsection{Convergence proof}
\label{app:convproof}
We will need to make the following assumptions. 
\begin{tcolorbox}[colback=IBMultramarine!20,colframe=IBMultramarine,enhanced]
\begin{assumption}\label{ass:conv}
$F$, $G$, $\rho$, and $\{\mu_\text{enc}^t,\nu_\text{enc}^t,\mu_\text{dec}^t,\nu_\text{dec}^t\}_{t=0}^\infty$ are such that:
\begin{enumerate}
    \item $F$ and $G$ are continuous and bounded below.
    \item For all $x \in \RR^n $, $\| W_\text{dec} \rho(W_\text{enc}x + b_\text{enc}) + b_\text{dec} - x\|_2^2$ is $C^1$ in $W_\text{enc}$, $b_\text{enc}$, $W_\text{dec}$, and $b_\text{dec}$, with a gradient that is locally Lipschitz. 
    \item $\{\mu_\text{enc}^t,\nu_\text{enc}^t,\mu_\text{dec}^t,\nu_\text{dec}^t\}_{t=0}^\infty \in (a,b)$ for some $0<a<b<\infty$. 
\end{enumerate}
These together entail that \cite[Assumptions $(\mathcal{H})$ and $(\mathcal{H}_1)$]{attouch2010proximal} are satisfied. 
\begin{enumerate}[resume]
    \item  $F$, $G$, and $\rho$ are piecewise (real) analytic functions with finitely many pieces.
\end{enumerate}
This entails that $\cL$ is a continuous and piecewise analytic function, with finitely many pieces.
\end{assumption}
\end{tcolorbox}

\begin{tcolorbox}[colback=gray!10,colframe=gray,enhanced]
\begin{nb}\label{nb:phomodifications}
    In the ReLU case, we must modify $\rho$ for \cref{ass:conv}(2) to hold. For example taking $\rho$ to be any smooth and analytic approximation of ReLU, such as the Swish activation function, will suffice. 
    In the TopK case, $\rho= \TopK$ is not continuous. This can however be patched by using the following analytic approximation of TopK: for any $T>0$, let
    \[
    \rho_T(v) := \sum_{S \subseteq \{1,...,d\}, |S|=K} \left(\sum_{S'}\exp\left( \frac{1}{T}\sum_{j\in S'}v_{j } \right)\right)^{-1} \exp\left( \frac{1}{T}\sum_{j\in S}v_{j} \right) P_Sv.
    \]
    It will then follow that if $\rho=\rho_T$ then \cref{ass:conv} holds. Furthermore, as $T \to 0$, $\rho_T(v) \to \TopK(v)$, so long as $v$ has a unique set of largest $K$ entries.
\end{nb}
\end{tcolorbox}
\begin{proof}[Proof of \cref{nb:phomodifications}]
    It is straightforward to check that if $\rho$ is smooth and (real) analytic, then the $\rho$-dependent conditions of \cref{ass:conv} will be satisfied. 

    As for the convergence of $\rho_T$ to $\TopK$, let $v$ have a unique set of $K$ largest entries, and denote this set $S^*$. Then multiplying the numerator and denominator by $\exp\left(- \frac{1}{T}\sum_{j\in S^*}v_{j} \right)$ we get that
     \[
    \rho_T(v) = \sum_{S \subseteq \{1,...,d\}, |S|=K} \frac{\exp\left( \frac{1}{T}\left[\sum_{j\in S}v_{j}-\sum_{j\in S^*}v_{j}\right] \right)}{1+\sum_{S'\neq S^*}\exp\left( \frac{1}{T}\left[\sum_{j\in S'}v_{j}-\sum_{j\in S^*}v_{j}\right] \right)}  P_Sv.
    \]
    As $T \to 0$, if $S \neq S^*$ then $\exp\left( \frac{1}{T}\left[\sum_{j\in S}v_{j}-\sum_{j\in S^*}v_{j}\right] \right)\to 0$, since $\sum_{j\in S}v_{j}-\sum_{j\in S^*}v_{j}$ is stricly negative. Hence, as $T\to 0$
    \[
    \rho_T(v) \to P_{S^*}v = \TopK(v).
    \]
\end{proof}

The theory in \cite{attouch2010proximal} relies crucially on the Kurdyka--Łojasiewicz property, which we now define. 

\begin{tcolorbox}[colback=IBMultramarine!20,colframe=IBMultramarine,enhanced]

\begin{definition}[Kurdyka--Łojasiewicz property]\label{def:KL}
	A proper lower semi-continuous function $g:\mathbb{R}^n \to (-\infty,\infty]$ has the \emph{Kurdyka--Łojasiewicz property} at $\hat x \in \operatorname{dom} \partial g$\footnote{We denote by $\operatorname{dom} g$ the set of $x$ such that $g(x)< \infty$, and by $\operatorname{dom} \partial g$ the set of $x\in \operatorname{dom} g$ such that the (limiting) subdifferential of $g$ at $x$, $\partial g(x)$ \cite[Definition 2.1]{attouch2010proximal}, is non-empty. } if there exist $\eta \in (0,\infty]$, a neighbourhood $U$ of $\hat x$, and a continuous concave function $\varphi:[0,\eta) \to [0,\infty)$, such that 
	\begin{itemize}
		\item $\varphi$ is $C^1$ with $\varphi(0)=0$ and $\varphi' >0$ on $(0,\eta)$, and
		\item for all $x \in U$ such that $g(\hat x) < g(x) < g(\hat x)+ \eta$, the  \emph{Kurdyka--Łojasiewicz inequality} holds:
		\[
		\varphi'(g(x) - g(\hat x))\operatorname{dist}(\mathbf{0},\partial g(x)) \geq 1. 
		\]
			\end{itemize}
		If $\varphi(s) := cs^{1-\theta}$ is a valid concave function for the above with $c > 0$ and $\theta \in [0,1)$, then we will say that $g$ has the  \emph{Kurdyka--Łojasiewicz property with exponent $\theta$} at $\hat x$. 

        Note that if $g$ is differentiable on $U$ and $\varphi(s):= cs^{1-\theta}$, this inequality becomes 
        \[
        c(1-\theta) \|\nabla g(x)\|_2 \geq (g(x)-g(\hat x))^\theta.
        \]
	\end{definition}
\end{tcolorbox}

\begin{tcolorbox}[colback=IBMgold!20,colframe=IBMgold,enhanced]
 \begin{theorem}\label{thm:LisKL}
If \cref{ass:conv} holds, then for all $W_\text{enc},b_\text{enc},W_\text{dec},b_\text{dec}$, $\cL$ has the Kurdyka--Łojasiewicz property, with some exponent $\theta\in [0,1)$, at $W_\text{enc},b_\text{enc},W_\text{dec},b_\text{dec}$. 
 \end{theorem}
\end{tcolorbox}
\begin{proof}
    Since $\cL$ is continuous and piecewise analytic with finitely many pieces, it follows that it is semi-analytic (see \cite{lojasiewicz1964}). Then if $(W_\text{enc},b_\text{enc},W_\text{dec},b_\text{dec})$ is not a critical point of $\cL$, the result follows by \cite[Lemma 2.1]{KLcalculus}, and if $(W_\text{enc},b_\text{enc},W_\text{dec},b_\text{dec})$ is a critical point, the result follows by \cite[Theorem 3.1]{Bolte2007}.
\end{proof}
We therefore prove a more detailed version of \cref{thm:PAMconvlite}.
\begin{tcolorbox}[colback=IBMgold!20,colframe=IBMgold,enhanced]
 \begin{theorem}\label{thm:PAMconv}
If \cref{ass:conv} holds, then for all $W^0_\text{enc},b^0_\text{enc},W^0_\text{dec},b^0_\text{dec}$ the sequence of SAE parameters $\{ \Theta_t :=(W^t_\text{enc},b^t_\text{enc},W^t_\text{dec},b^t_\text{dec})\}_{t=0}^\infty$ defined by \cref{eq:PAM} obeys:
\begin{enumerate}[i.]
    \item %$\cL(W^{t+1}_\text{enc},b^{t+1}_\text{enc},W^{t+1}_\text{dec},b^{t+1}_\text{dec}) \leq \cL(W^t_\text{enc},b^t_\text{enc},W^t_\text{dec},b^t_\text{dec})$
$\cL(\Theta_{t+1}) \leq \cL(\Theta_t)$ 
    with equality if and only if $\Theta_{t+1}=\Theta_t$. %$(W^{t+1}_\text{enc},b^{t+1}_\text{enc},W^{t+1}_\text{dec},b^{t+1}_\text{dec}) = (W^t_\text{enc},b^t_\text{enc},W^t_\text{dec},b^t_\text{dec})$.
    \item %$(W^{t+1}_\text{enc}-W^t_\text{enc},b^{t+1}_\text{enc}-b^t_\text{enc},W^{t+1}_\text{dec}-W^t_\text{dec},b^{t+1}_\text{dec}-b^t_\text{dec}) \to 0$ 
    $\Theta_{t+1}-\Theta_t \to 0$ as $t \to \infty$, and the number of $t$ such that $\Theta_{t+1}-\Theta_t$ has norm greater than some threshold $\varepsilon>0$ is proportional to at most $\varepsilon^{-2}$.  %square-summable differences. 
    % \[
    % \sum_{t=0}^\infty \|W^{t+1}_\text{enc}-W^t_\text{enc}\|_F^2 +\|b^{t+1}_\text{enc}-b^t_\text{enc}\|_2^2 +\|W^{t+1}_\text{dec}-W^t_\text{dec}\|_F^2 +\|b^{t+1}_\text{dec}-b^t_\text{dec}\|_2^2 < \infty,
    % \]
    \item Every limit point of $\{ %W^t_\text{enc},b^t_\text{enc},W^t_\text{dec},b^t_\text{dec}
    \Theta_t
    \}_{t=0}^\infty$ is a critical point of $\cL$. 
\end{enumerate}
And furthermore if the sequence $\{ %W^t_\text{enc},b^t_\text{enc},W^t_\text{dec},b^t_\text{dec}
\Theta_t
\}_{t=0}^\infty$ is bounded:
\begin{enumerate}[i.,resume]
\item The number of %$(W^{t+1}_\text{enc}-W^t_\text{enc},b^{t+1}_\text{enc}-b^t_\text{enc},W^{t+1}_\text{dec}-W^t_\text{dec},b^{t+1}_\text{dec}-b^t_\text{dec})$ 
$\Theta_{t+1}-\Theta_t$ with norm greater than $\varepsilon$ is proportional to at most $\varepsilon^{-1}$.
%\[
%     \sum_{t=0}^\infty \|W^{t+1}_\text{enc}-W^t_\text{enc}\|_F +\|b^{t+1}_\text{enc}-b^t_\text{enc}\|_2 +\|W^{t+1}_\text{dec}-W^t_\text{dec}\|_F +\|b^{t+1}_\text{dec}-b^t_\text{dec}\|_2 < \infty.
%     \]
    \item %(W^t_\text{enc},b^t_\text{enc},W^t_\text{dec},b^t_\text{dec})$ 
    $\Theta_t$ converges to a critical point of $\cL$ as $t \to \infty$. 
\end{enumerate}
Finally, if %$(W^t_\text{enc},b^t_\text{enc},W^t_\text{dec},b^t_\text{dec}) \to (W^\infty_\text{enc},b^\infty_\text{enc},W^\infty_\text{dec},b^\infty_\text{dec})$ 
$\Theta_t \to \Theta_\infty$
and $\theta \in [0,1)$ is the Kurdyka--Łojasiewicz exponent of $\cL$ at %$(W^\infty_\text{enc},b^\infty_\text{enc},W^\infty_\text{dec},b^\infty_\text{dec})$
$\Theta_\infty$, then:
\begin{enumerate}[i.,resume]
\item If $\theta = 0$, $\{ %W^t_\text{enc},b^t_\text{enc},W^t_\text{dec},b^t_\text{dec}
\Theta_t\}_{t=0}^\infty$ converges after finitely many steps. 
\item If $\theta \in (0,1/2]$, there exist $c > 0$ and $\zeta \in [0,1)$ such that 
$
\| %(W^t_\text{enc},b^t_\text{enc},W^t_\text{dec},b^t_\text{dec}) - (W^\infty_\text{enc},b^\infty_\text{enc},W^\infty_\text{dec},b^\infty_\text{dec})
\Theta_t - \Theta_\infty\| \leq c\zeta^t$.
\item If $\theta \in (1/2,1)$, there exists $c > 0$ such that 
$
\| %(W^t_\text{enc},b^t_\text{enc},W^t_\text{dec},b^t_\text{dec}) - (W^\infty_\text{enc},b^\infty_\text{enc},W^\infty_\text{dec},b^\infty_\text{dec})
\Theta_t - \Theta_\infty\|\leq ct ^{-\frac{1-\theta}{2\theta -1}}$.
\end{enumerate}
 \end{theorem}
\end{tcolorbox}
\begin{proof}
    (i) and (ii) follow from \cite[Lemma 3.1]{attouch2010proximal}. (iii) follows from \cite[Proposition 3.1]{attouch2010proximal}. (iv) and (v) follow from \cref{thm:LisKL} and \cite[Theorem 3.2]{attouch2010proximal}.
    (vi) to (viii) follow from \cref{thm:LisKL} and \cite[Theorem 3.4]{attouch2010proximal}.
\end{proof}
\begin{tcolorbox}[colback=gray!10,colframe=gray,enhanced]
\begin{nb}
    If $F$ and $G$ include weight decay terms this will not impede \cref{ass:conv}, and furthermore will by (i) ensure that  $\{ W^t_\text{enc},b^t_\text{enc},W^t_\text{dec},b^t_\text{dec}\}_{t=0}^\infty$ is always bounded, and hence weight decay ensures convergence of trajectories of \cref{eq:PAM} to critical points of $\cL$. 
\end{nb}
\end{tcolorbox}

\section{Experimental settings}
\label{app:setups}

All  computations were performed on: WS Obsidian 750D AirFlow / AMD Ryzen 9 3900X 12x3.8 Ghz / 2x32GB DDR4 3600 / X570 WS / DIS. NOCTUA / 1000W Platinum / 2TB NVME Ent. / RTX 3090 24GB.

All code is available at: \url{https://github.com/splInterp2025/splInterp}. 

\subsection{SAEs as a bridge between $k$-means and PCA}\label{app:SAEbridgesetup}

\begin{itemize}
    \item \textbf{Data sampling:} 100 points in 2D, sampled as three clusters:
    \begin{itemize}
        \item 40\% on a noisy horizontal line ($x$ from $-1.5$ to $0$, $y \approx -0.8$)
        \item 30\% in a dense square ($x \in [-0.4, 0.4]$, $y \in [0.4, 1.2]$)
        \item 30\% along a noisy diagonal ($x \approx 0.8 + t$, $y \approx t - 0.5$, $t \in [0,1]$)
    \end{itemize}
    
    \item \textbf{Number of data points:} 100

    \item \textbf{SAE architecture:} Linear sparse autoencoder with 80 dictionary elements ($d = 80$), 2D input/output, using a Top1 or Top3 sparse coding.

    \item \textbf{Training method and hyperparameters:}
    \begin{itemize}
        \item Adam optimiser (learning rate $8 \times 10^{-3}$)
        \item 5000 steps, full-batch (all data at once)
        \item Dictionary initialised near data points
    \end{itemize}

    \item \textbf{Runtime:} 6 minutes
\end{itemize}

\subsection{PAM-SGD vs. SGD on MNIST}
\label{app:MNISTsetup}
\begin{itemize}
    \item \textbf{Number of data points and training/test split:} Uses the standard MNIST dataset:
    \begin{itemize}
        \item 60{,}000 training images
        \item 10{,}000 test images
        \item Images are $28 \times 28$ grayscale digits
    \end{itemize}

    \item \textbf{SAE architecture(s):} Two models:
    \begin{itemize}
        \item \textbf{SGD Autoencoder:} Linear encoder/decoder, tied weights, sparsity via TopK ($K = 15$) or ReLU (without L1 regularisation), 256 latent dimensions
        \item \textbf{PAM-SGD Autoencoder:} Linear encoder, decoder weights solved analytically (not tied), same latent size and sparsity.
    \end{itemize}

    \item \textbf{Training (hyper)parameters:}
    \begin{itemize}
        \item Optimiser: Adam (learning rate $0.003$)
        \item Batch size: 128 (SGD), 1024 (PAM-SGD encoder update)
        \item Number of epochs: 50 (default, or fewer for small ablation subsets)
        \item $K$-sparsity: $K = 15$ (for TopK)
        \item L1 regularisation: not used
        \item Input dimension: 784 ($28 \times 28$)
        \item Ablation studies vary training set size, $K$-sparsity, number of SGD steps per batch, activation type, weight decay parameters, and cost-to-move parameters. 
    \end{itemize}

    \item \textbf{Running time:} 1.5 to 2 hours
\end{itemize}

\subsection{PAM-SGD vs. SGD on Gemma}
\label{app:Gemmasetup}
\begin{itemize}
    \item \textbf{Details on Gemma version and license:} Uses activations from \href{https://huggingface.co/google/gemma-2-2b}{Gemma-2-2B} (\url{https://huggingface.co/google/gemma-2-2b}) (Google, 2.2B parameters). License: see \href{https://ai.google.dev/gemma/terms}{Gemma Terms of Use} (\url{https://ai.google.dev/gemma/terms}) (accessed 8th May 2025).
    
    \item \textbf{Number of data points and training/test split:} Up to 10{,}000 LLM activation vectors extracted from real text (default: 90\% train, 10\% test split).
    
    \item \textbf{SAE architecture(s):} Two models:
    \begin{itemize}
        \item \textbf{SGD Autoencoder:} Linear encoder/decoder (tied weights), 4096 latent dimensions, sparsity via TopK ($K = 320$) or ReLU, with L1 regularisation and “cost-to-move” penalties.
        \item \textbf{PAM-SGD Autoencoder:} Linear encoder, decoder weights solved analytically (not tied), same latent size and sparsity, with additional regularisation.
    \end{itemize}

    \item \textbf{Training (hyper)parameters:}
    \begin{itemize}
        \item Optimiser: Adam (learning rate $0.001$)
        \item Batch size: 256 (SGD), 2048 (PAM-SGD Encoder update)
        \item Epochs: 100
        \item $K$-sparsity: $K = 320$ (for TopK)
        \item L1 regularisation: $0.01$ (TopK), $0.00001$ (ReLU)
        \item “Cost-to-move” and weight decay regularisation for encoder/decoder
        \item Ablation studies vary training set size, $K$-sparsity, number of SGD steps per batch, activation type, weight decay parameters, and cost-to-move parameters. 
    \end{itemize}

    \item \textbf{Runtimes:} 3 to 7 minutes
\end{itemize}

\section{Additional figures and ablation studies}
\label{app:figures}
%%%%%%%%%%%%%%%%%%%%%%%%%%%%%%%%%%%%%%%%%%%%%%%%%%%%%%%%%%%%

\subsection{SAEs as a bridge between $k$-means and PCA}

\begin{figure}[h]
    \centering
        \includegraphics[width=\textwidth]{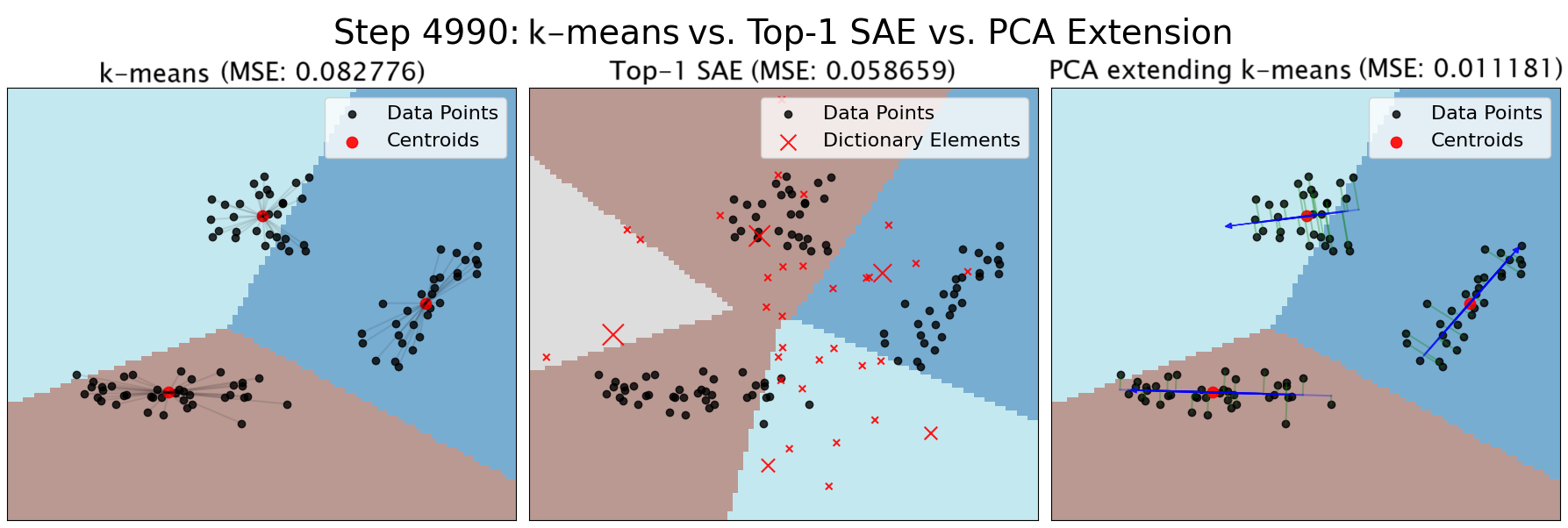}
    
    \caption{\textbf{Visualising the SAE bridge between $k$-means clustering and PCA.} Top-1 SAE. %(b,c) Soft top-1 SAE 
    %(Bottom) %(d) 
    %Top-3 SAE.
    }
    \label{fig:sae_spectrum2}
\end{figure}
\newpage

\subsection{MNIST experiments}\label{app:MNISTfigs}
\subsubsection{TopK experiments}
\begin{figure}[htbp]
    \centering
    \includegraphics[width=\textwidth]{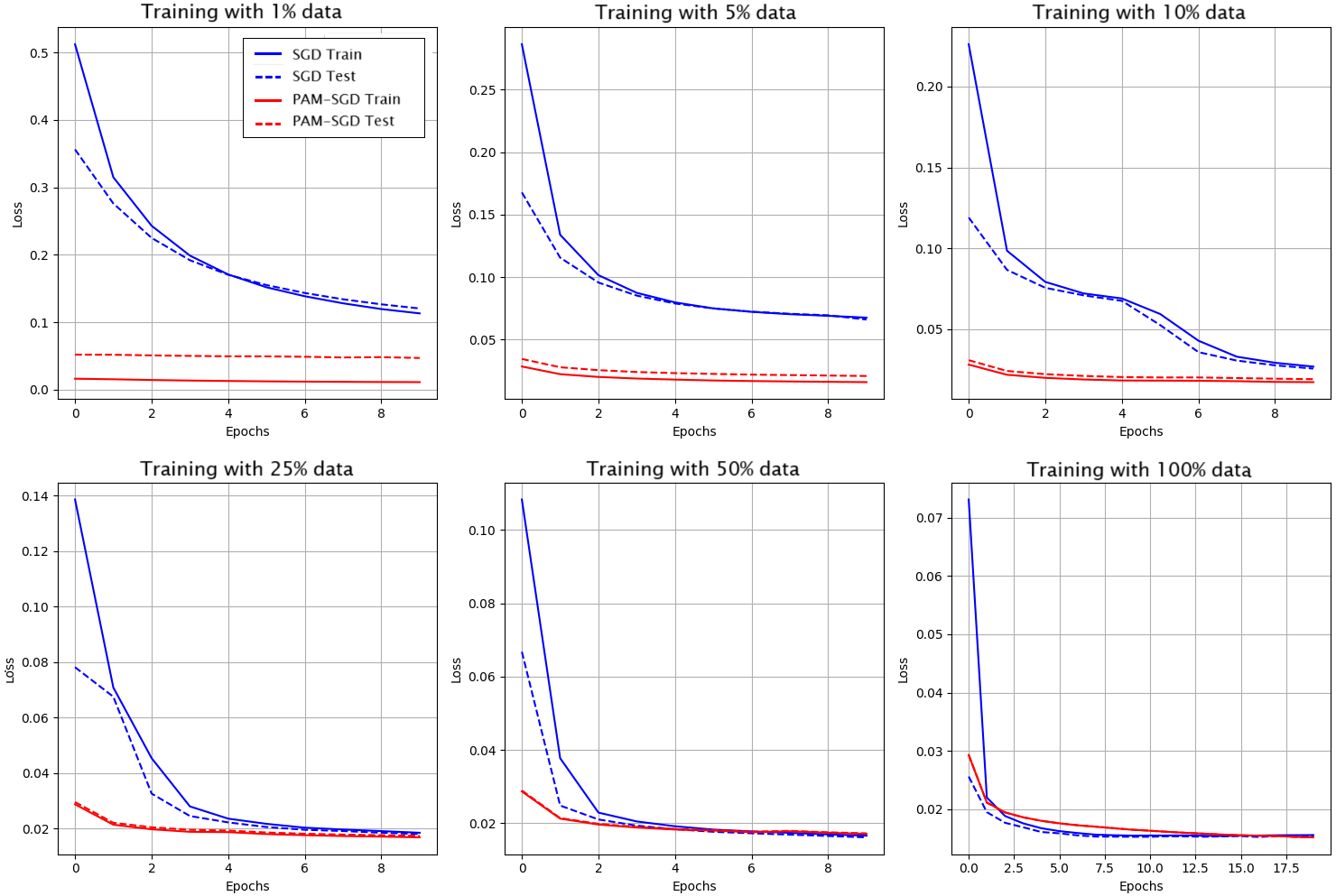}
    \caption{\textbf{Training and test loss curves at different data sizes for MNIST, with TopK ($K=15$) activation.} The chart highlights PAM-SGD's superior sample efficiency.}
    \label{fig:mnistTopK}
\end{figure}

\paragraph{PAM-SGD similarly outperforms SGD with TopK. (\cref{fig:mnistTopK})} We tested PAM-SGD using TopK activation for $K=15$. We again saw PAM-SGD outperform SGD, especially at low training data levels. 

\subsubsection{Reconstruction accuracy and interpretability}
\begin{figure}[htbp]
    \centering
    \includegraphics[width=1\linewidth]{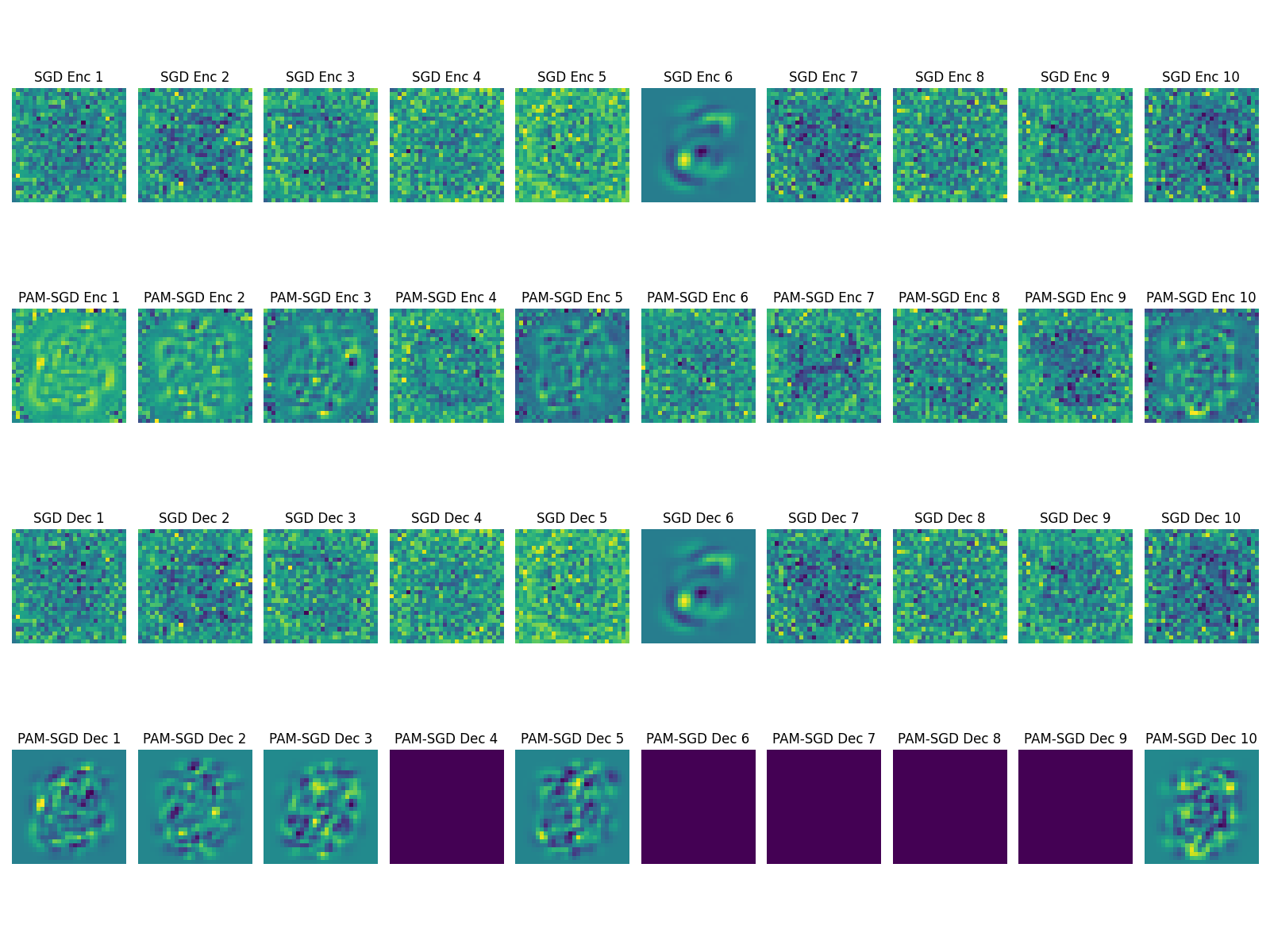}
    \caption{\textbf{Learned Dictionary Elements using ReLU.} Visualization of encoder and decoder weights as filters: SGD encoder (top row), PAM-SGD encoder (second row), SGD decoder (third row), and PAM-SGD decoder (bottom row). }
    \label{fig:mnist_filters}
\end{figure}
\begin{figure}[htbp]
    \centering
    \includegraphics[width=1\linewidth]{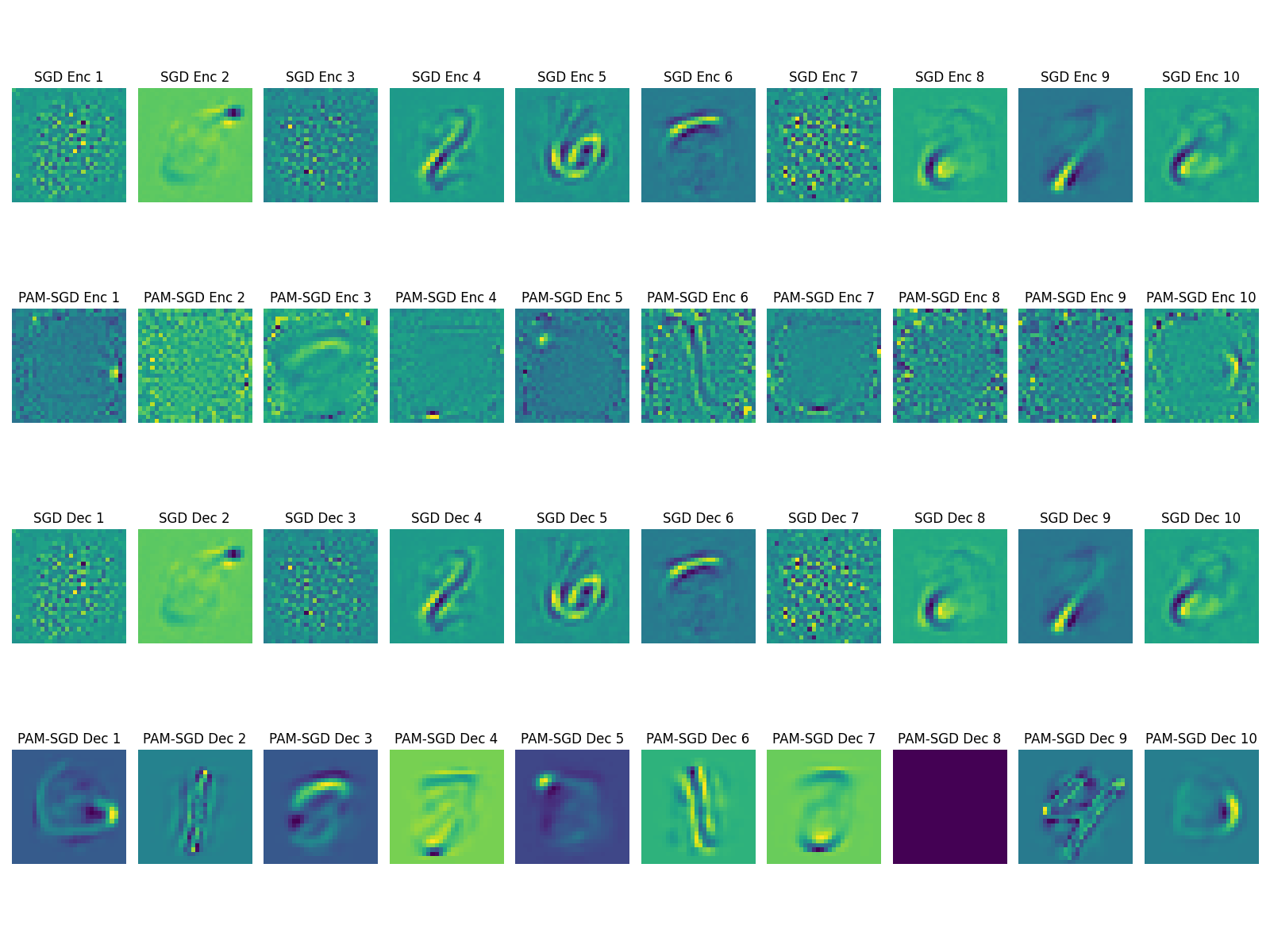}
    \caption{\textbf{Learned Dictionary Elements using TopK $K=15$.} Visualization of encoder and decoder weights as filters: SGD encoder (top row), PAM-SGD encoder (second row), SGD decoder (third row), and PAM-SGD decoder (bottom row). With TopK, PAM-SGD produces more interpretable features representative of digit components }
    \label{fig:mnist_filters_TopK}
\end{figure}
\begin{figure}[htbp]
    \centering
    \includegraphics[width=1\linewidth]{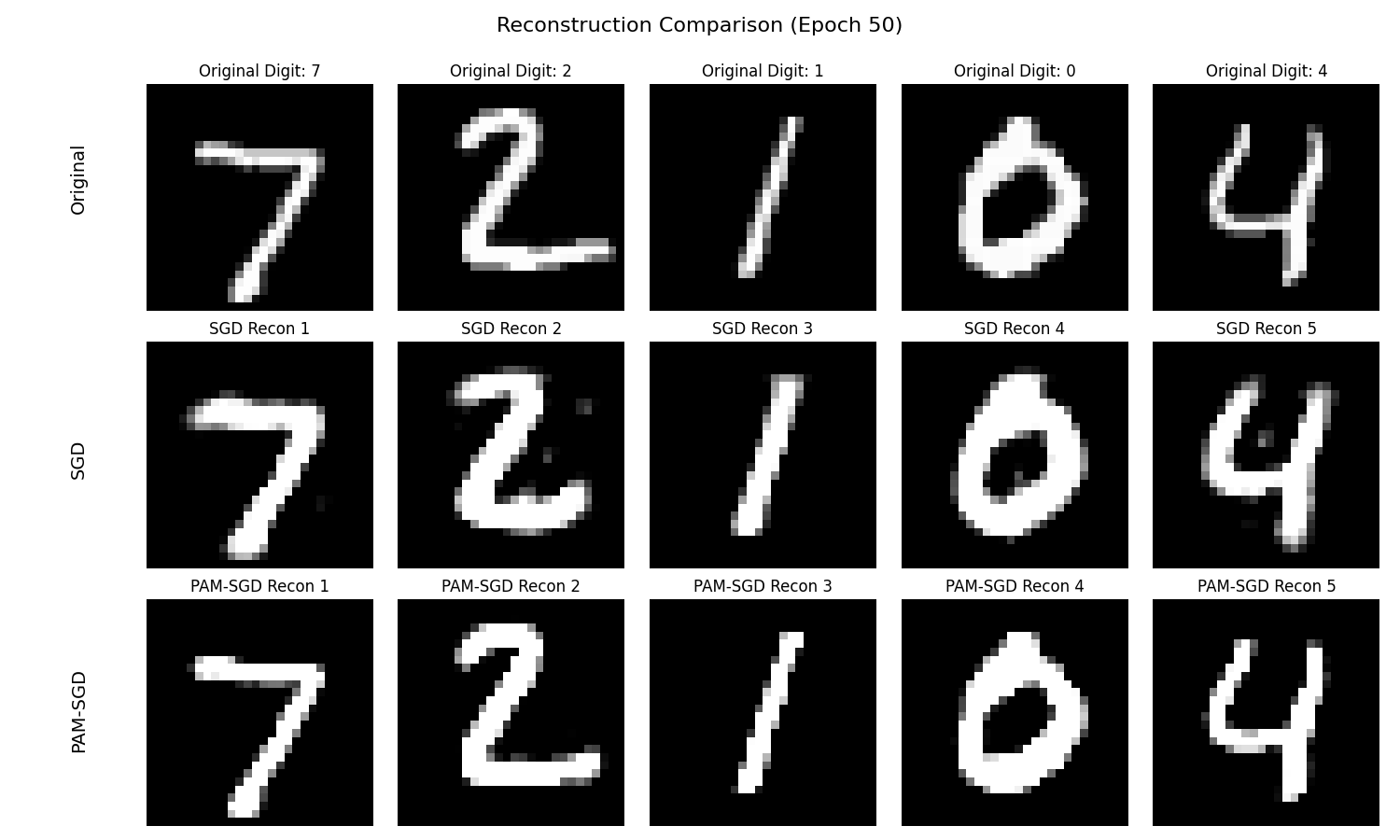}
    \caption{\textbf{Reconstruction Quality Comparison using ReLU.} Original MNIST digits (top row) with their reconstructions using SGD optimization (middle row) and PAM-SGD optimization (bottom row). PAM-SGD produces cleaner, more accurate reconstructions. }
    \label{fig:mnist_reconstruction}
    \end{figure}
\begin{figure}[htbp]
    \centering
    \includegraphics[width=1\linewidth]{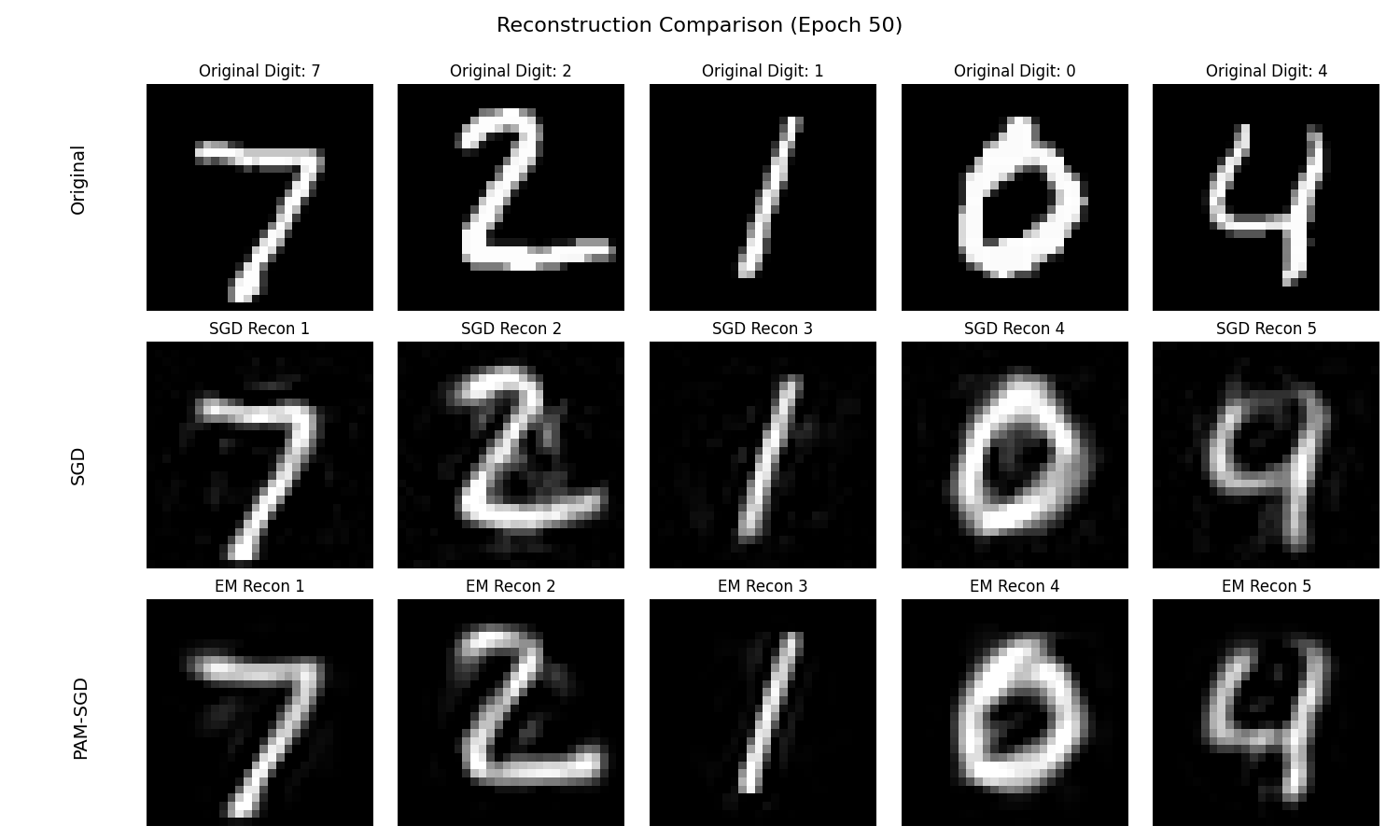}
    \caption{\textbf{Reconstruction Quality Comparison using TopK $K=15$.} Original MNIST digits (top row) with their reconstructions using SGD optimization (middle row) and PAM-SGD optimization (bottom row). PAM-SGD produces cleaner, more accurate reconstructions. }
    \label{fig:mnist_reconstruction_TopK}
\end{figure}
\begin{figure}[htbp]
    \centering
    \includegraphics[width=1\linewidth]{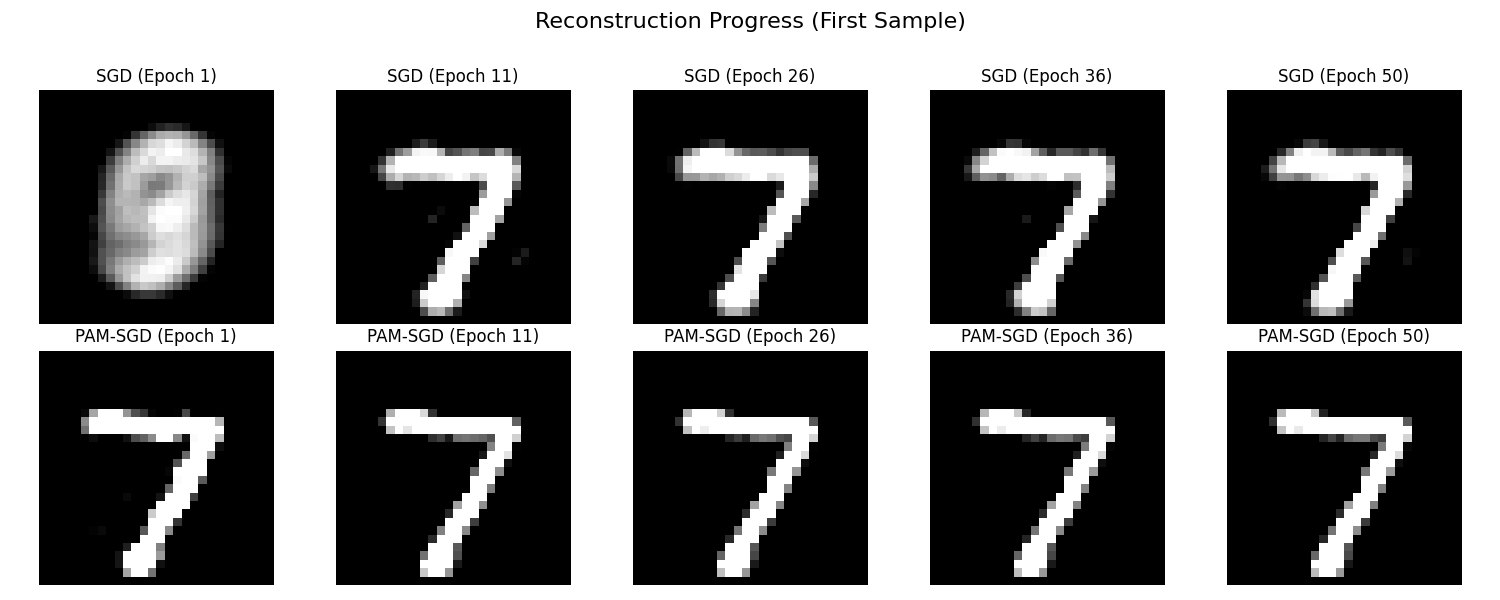}
    \caption{\textbf{Evolution of Reconstruction Quality Over Training using ReLU.} Progression of a single digit's reconstruction across epochs, comparing SGD (top row) and PAM-SGD (bottom row) approaches, showing how representation quality improves with training. Here PAM-SGD converges almost immediately. }
    \label{fig:mnist_evolution}
\end{figure}
\begin{figure}[htbp]
    \centering
    \includegraphics[width=1\linewidth]{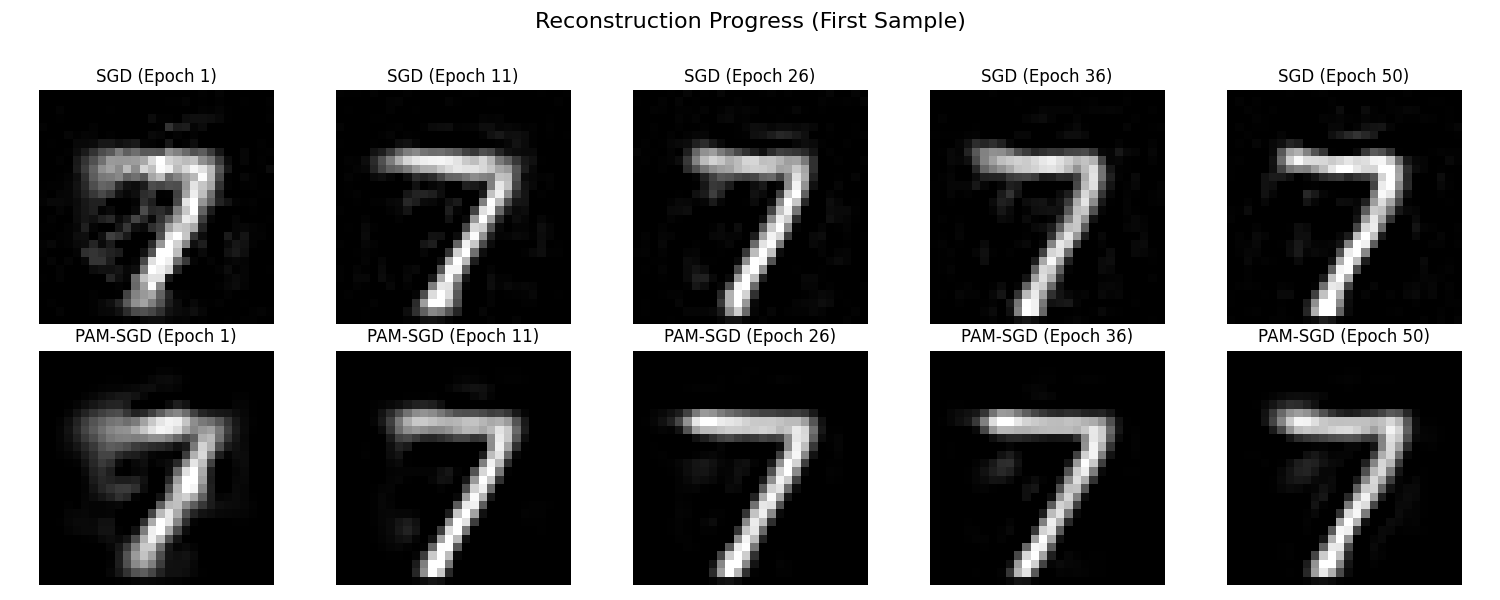}
    \caption{\textbf{Evolution of Reconstruction Quality Over Training using TopK $K=15$.} Progression of a single digit's reconstruction across epochs, comparing SGD (top row) and PAM-SGD (bottom row) approaches, showing how representation quality improves with training }
    \label{fig:mnist_evolution_TopK}
\end{figure}
\clearpage
\subsubsection{Ablation study varying SGD updates per batch in PAM-SGD}
\begin{figure}[htbp]
    \centering
    \includegraphics[width=1\linewidth]{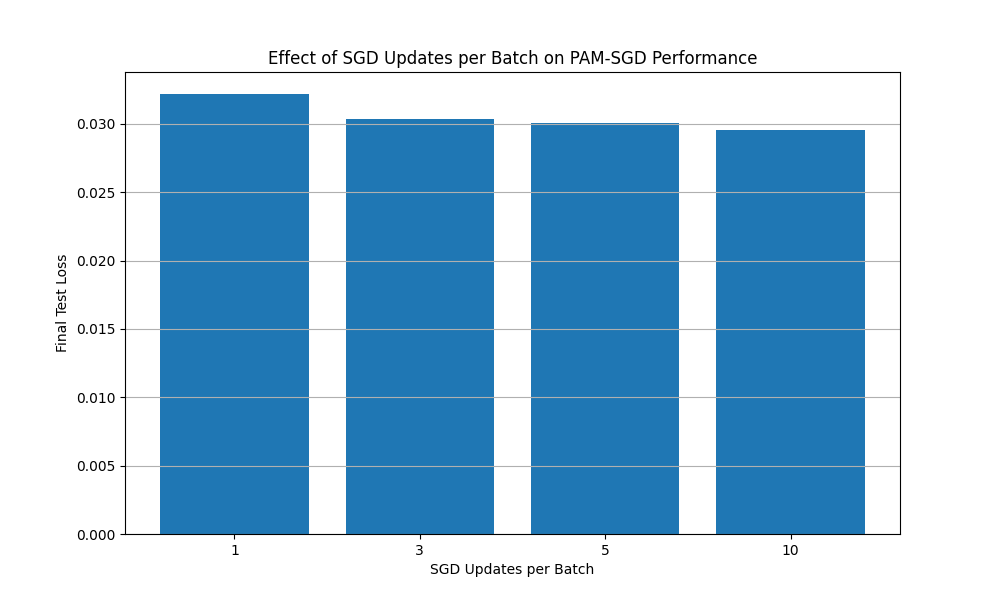}
    \caption{\textbf{Effect of Multiple SGD Updates Per Batch on PAM-SGD Performance with ReLU.} Test loss decreases with more updates per batch.}
    \label{fig:mnist_ablation_SGD_updates}
\end{figure}

\begin{figure}[htbp]
    \centering
    \includegraphics[width=1\linewidth]{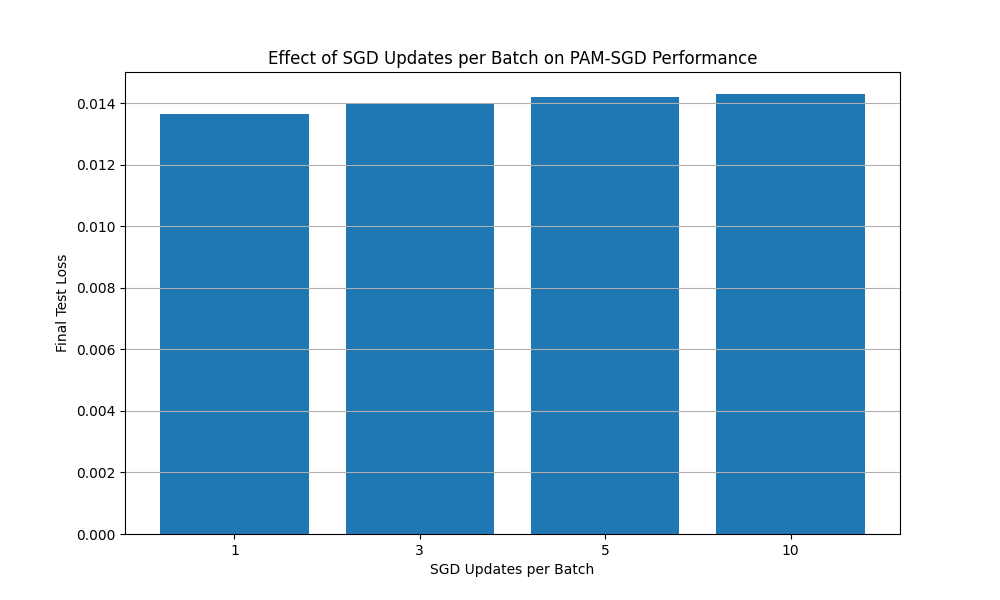}
    \caption{\textbf{Effect of Multiple SGD Updates Per Batch on PAM-SGD Performance with TopK $K=15$.} Test loss increases with more updates per batch, suggesting simpler optimization (single updates) maintains better balance.}
    \label{fig:mnist_ablation_SGD_updates_TopK}
\end{figure}

\paragraph{Stability Across SGD Updates. (\cref{fig:mnist_ablation_SGD_updates,fig:mnist_ablation_SGD_updates_TopK})}
Unlike in LLM experiments, PAM-SGD on MNIST is robust to the number of SGD updates per batch. Varying this hyperparameter from 1 to 10 has slightly improves final performance in the ReLU case and has little impact in the TopK case, suggesting that the optimization landscape is smoother and less sensitive in this setting.


\clearpage

\subsubsection{TopK and ReLU 
activation patterns}
\begin{figure}[h]
    \centering
    \includegraphics[width=1\linewidth]{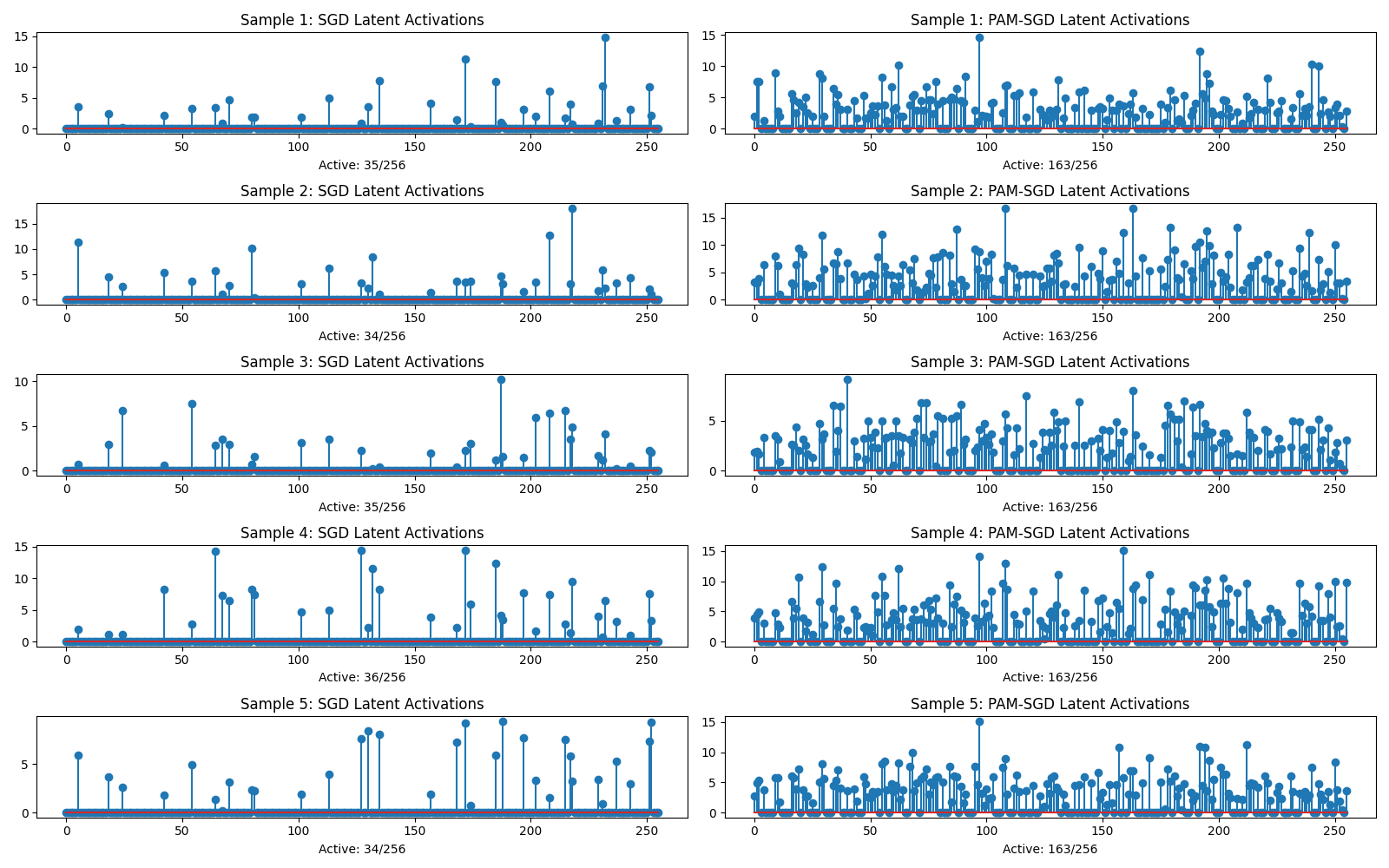}
    \caption{\textbf{Sparse activation patterns for ReLU activations.} Plots showing which latent neurons activate for 5 different input digits, comparing SGD (left) and PAM-SGD (right) models. PAM-SGD activations are roughly five times denser. }
    \label{fig:mnist_sparsity}
\end{figure}
\begin{figure}[h]
    \centering
    \includegraphics[width=1\linewidth]{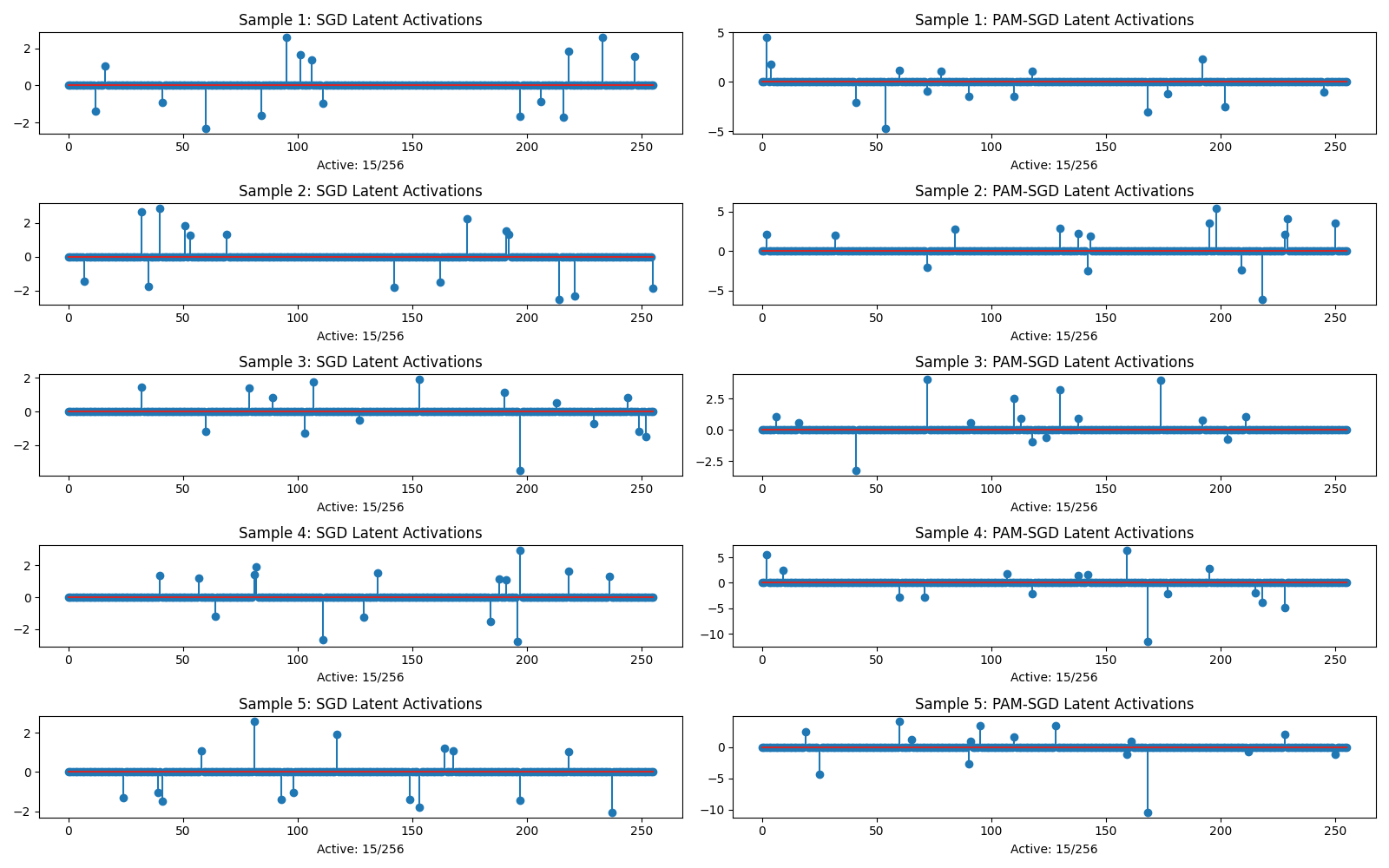}
    \caption{\textbf{Sparse activation patterns for TopK $K=15$ activations.} Plots showing which latent neurons activate for 5 different input digits, comparing SGD (left) and PAM-SGD (right) models. Each sample activates exactly $K=15$ neurons from the 256-dimensional latent space. }
    \label{fig:mnist_sparsity_TopK}
\end{figure}
\clearpage

\subsubsection{Ablation study adding weight decay}

\begin{figure}[h]
    \centering
    \includegraphics[width=\linewidth]{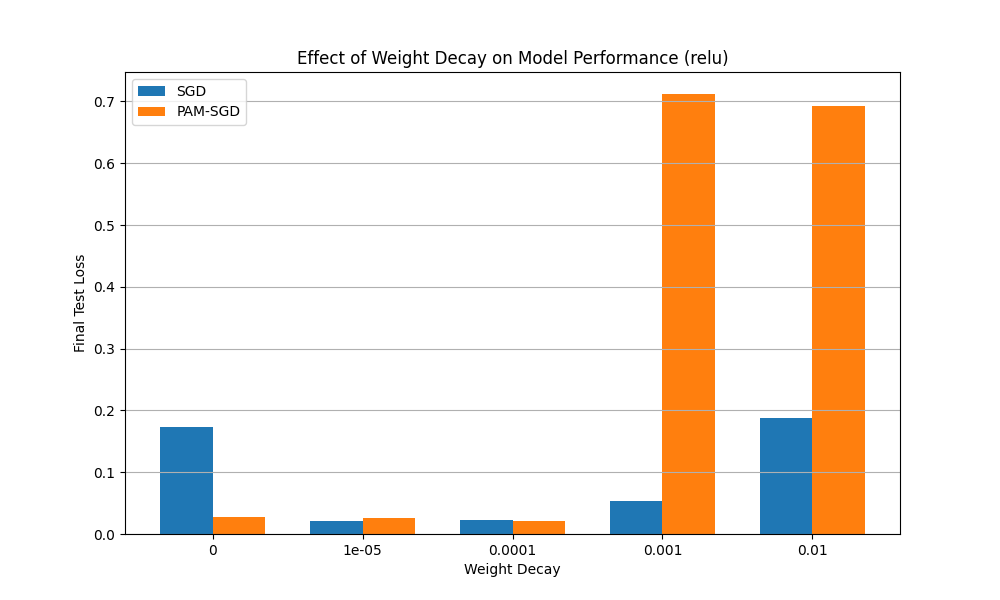}
    \caption{\textbf{Impact of weight decay on final test loss in the ReLU case.}}
    \label{fig:MNIST_weight_decay_ReLU}
\end{figure}

\begin{figure}[h]
    \centering
    \includegraphics[width=\linewidth]{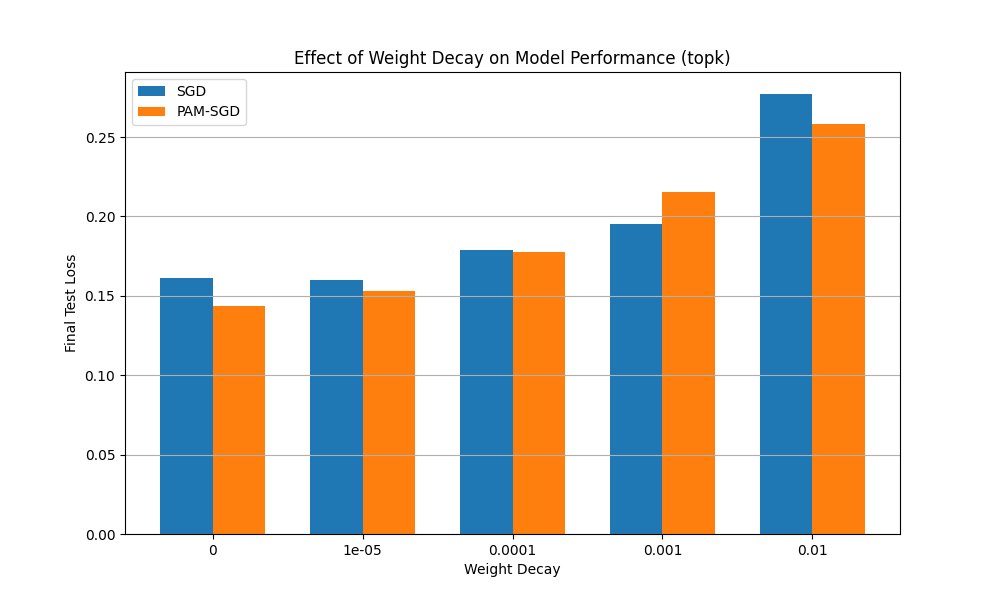}
    \caption{\textbf{Impact of weight decay on final test loss in the TopK $K=15$ case.}}
\label{fig:MNIST_weight_decay_topk}
\end{figure}

\paragraph{Small amounts of weight decay make SGD compete with PAM-SGD in the ReLU setting. (\cref{fig:MNIST_weight_decay_ReLU,fig:MNIST_weight_decay_topk})}
We experimented with adding weight decay in the 100\% data setting. In the ReLU setting, we found that small amounts aided SGD performance to be competitive with PAM-SGD and had little effect on PAM-SGD. Increasing weight decay further however degraded both performances, especially PAM-SGD's. In the TopK setting, weight decay just steadily degraded both performances.
\clearpage

\subsubsection{Ablation study varying $\mu$ and $\nu$}
\begin{figure}[h]
    \centering
\includegraphics[width=\linewidth]{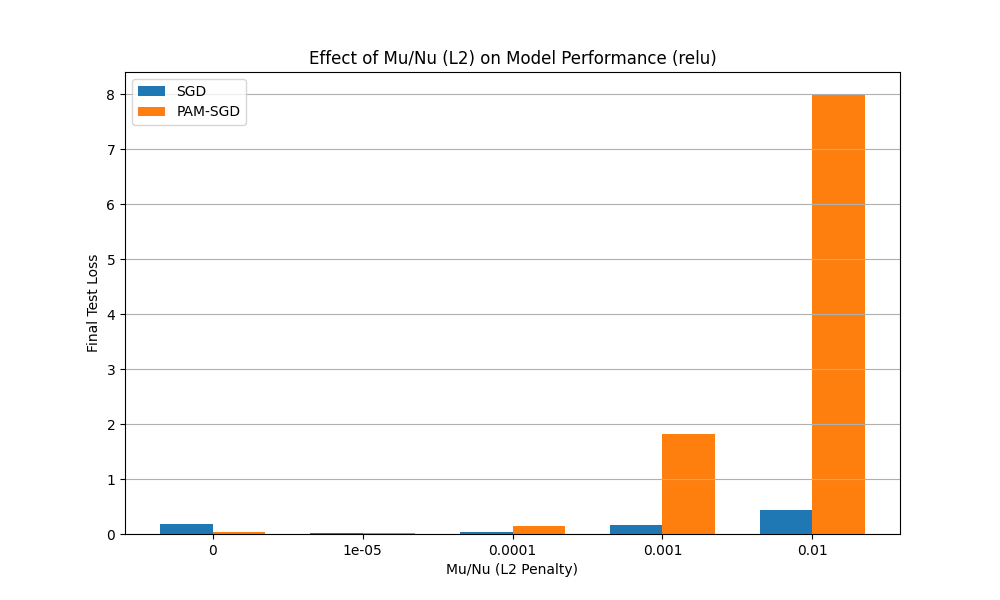}
    \caption{\textbf{Effect of modifying the cost-to-move parameters on final test loss in the ReLU case.}}
    \label{fig:MNISTmunu_relu}
\end{figure}

\begin{figure}[h]
    \centering
    \includegraphics[width=\linewidth]{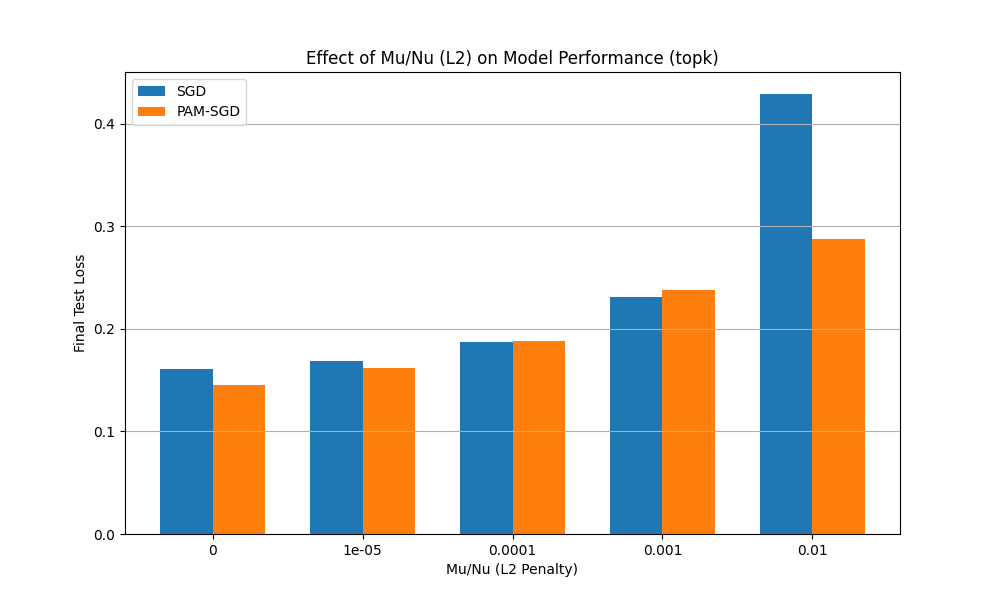}
    \caption{\textbf{Effect of modifying the cost-to-move parameters on final test loss in the TopK $K=15$ case.}}
    \label{fig:MNISTmunu_topk}
\end{figure}

\paragraph{Sensitivity to quadratic costs to move $\mu$ and $\nu$. (\cref{fig:MNISTmunu_relu,fig:MNISTmunu_topk}) }
We studied the effect of varying the values of the parameters $\mu_\text{enc},\mu_\text{dec},\nu_\text{enc},$ and $\nu_\text{dec}$ from \cref{eq:PAM}, in the 100\% data setting. For ReLU activation, very small values for these parameters improve the test loss almost to zero for both SGD (where similar parameters can easily be introduced) and PAM-SGD. Further increases however degrade performance for both, rapidly in the case of PAM-SGD. For TopK increasing these parameters steadily degrades performance in both cases, though this may simply be due to these parameters slowing convergence and therefore worsening performance at the 50 epoch cut-off. 

\clearpage
\subsection{LLM Activation Experiments (Gemma--2-2B)}
\label{app:Gemmafigs}
\subsubsection{Additional ReLU test runs}
\label{app:additionalReLU}
\begin{figure}[h]
    \centering
    \includegraphics[width=0.8\linewidth]{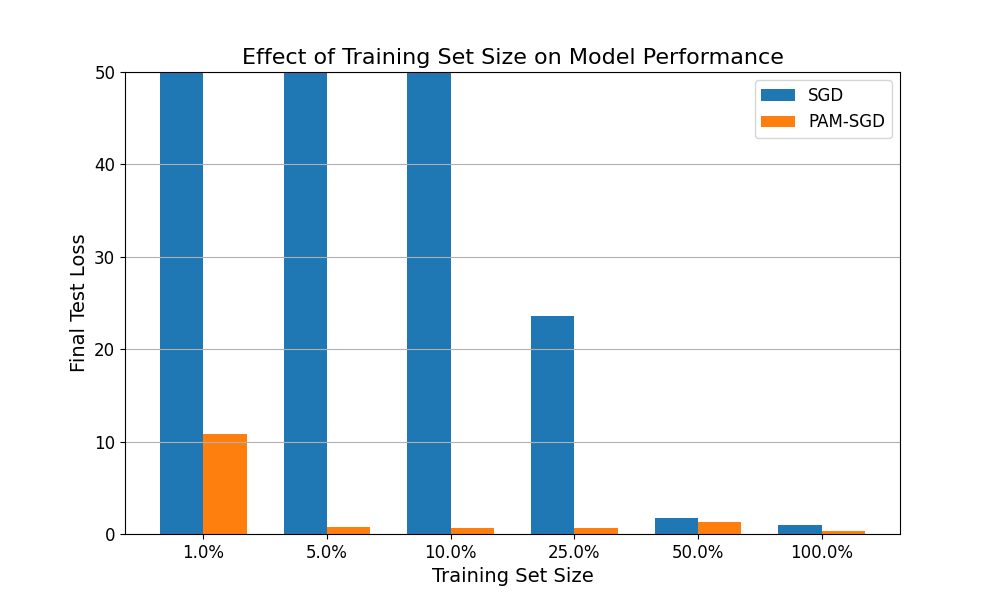}
    \includegraphics[width=0.8\linewidth]{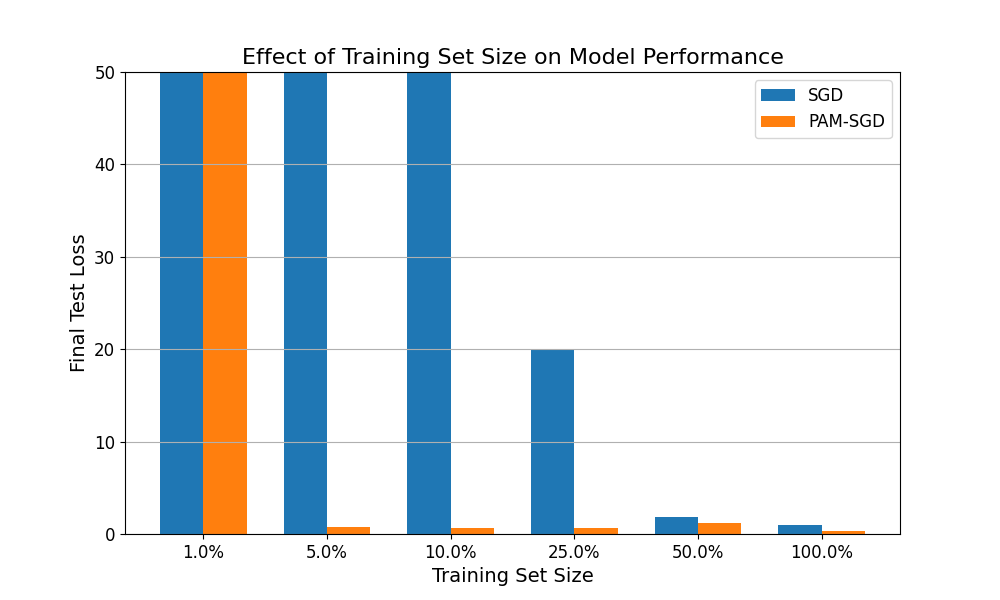}
    \caption{\textbf{Final test loss across training set sizes for Gemma-2-2B, with ReLU activation, for two additional training runs.} PAM-SGD retains an advantage at low data, and remains superior throughout, in both runs. %The top figure is the same run as \cref{fig:gemma_training_size}.
    }
    \label{fig:ReLU_reruns}
\end{figure}

\paragraph{PAM-SGD consistently outperforms SGD. (\cref{fig:ReLU_reruns})} Owing to the stochasticity of the training algorithms, different results are obtained in every re-run of the training. However, the pattern of PAM-SGD outperforming SGD remained consistent. 

\clearpage
\subsubsection{TopK activation experiments}
\begin{figure}[h]
    \centering
    \includegraphics[width=\linewidth]{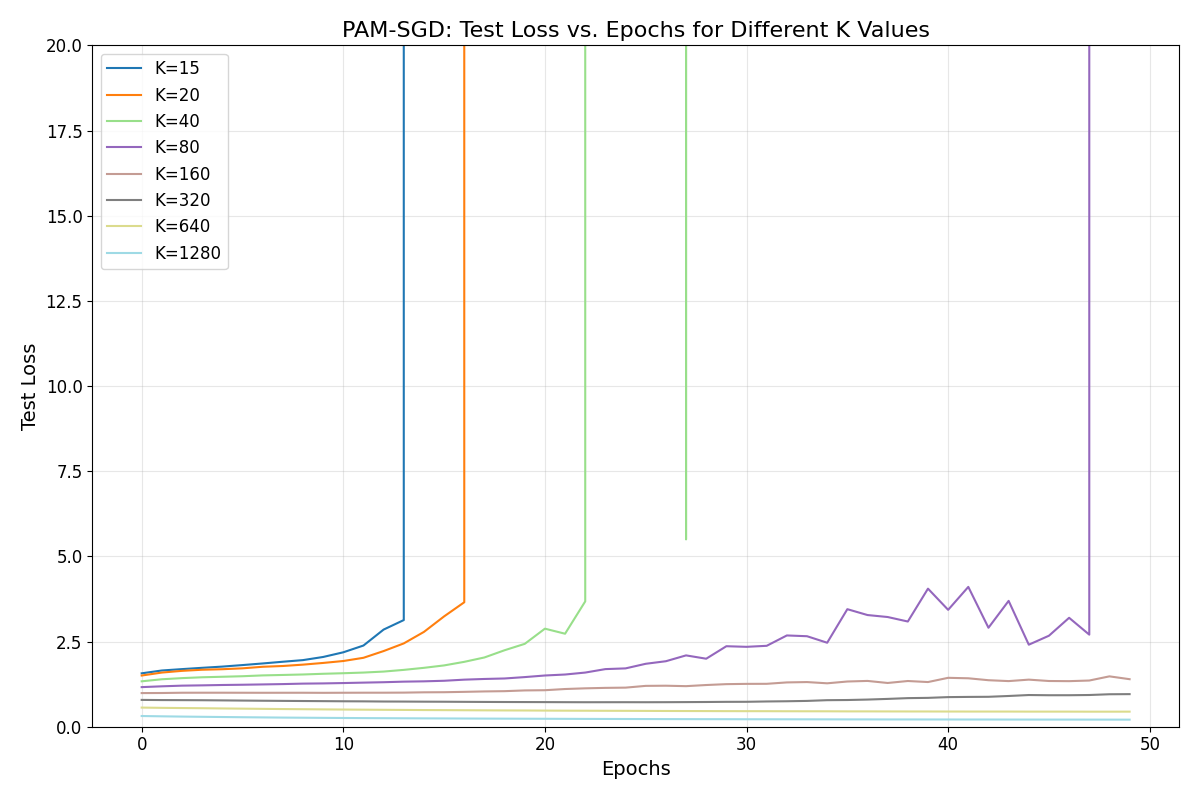}
    \caption{\textbf{PAM-SGD test loss stable only for high values of $K$.}}
    \label{fig:gemma_K_values}
\end{figure}

\begin{figure}[h]
    \centering
    \includegraphics[width=0.9\linewidth]{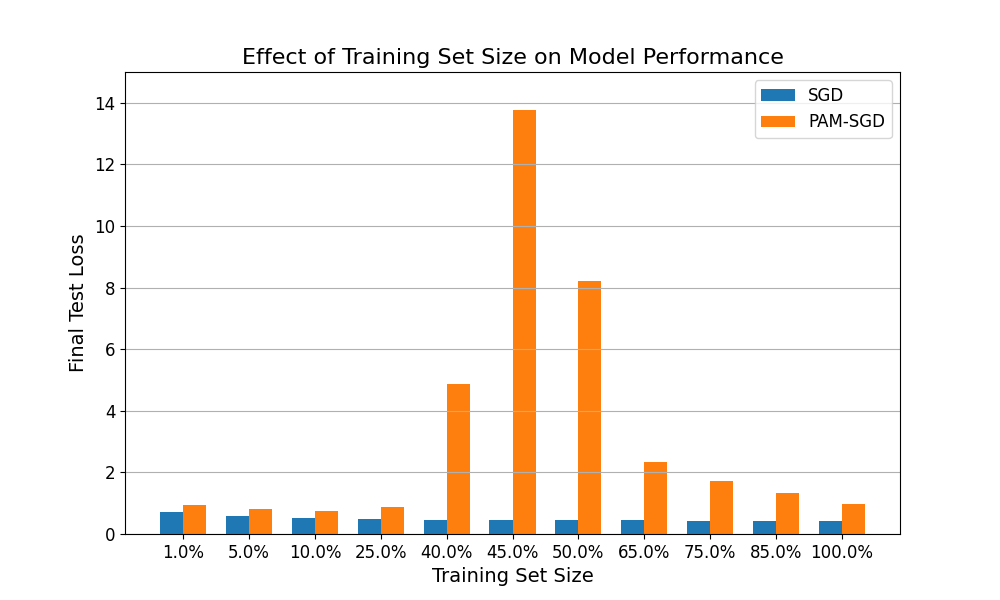}
    \caption{\textbf{Final test loss across training set sizes for Gemma-2-2b with TopK activation, for $K=320$.} PAM-SGD here consistently underperforms SGD, with a major peak at 45\% data.}
    \label{fig:gemma_training_size_Top320}
\end{figure}

\begin{figure}[h]
    \centering
    \includegraphics[width=0.9\linewidth]{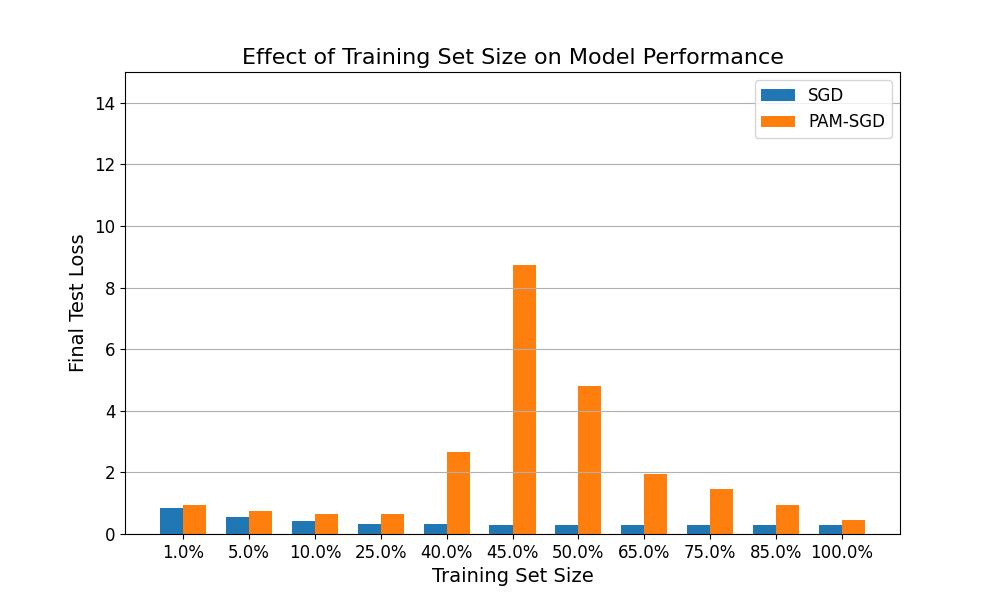}
    \caption{\textbf{Final test loss across training set sizes for Gemma-2-2b with TopK activation, for $K=640$.} PAM-SGD here consistently underperforms SGD, with a major peak at 45\% data.}
    \label{fig:gemma_training_size_Top640}
\end{figure}

\begin{figure}[h]
    \centering
    \includegraphics[width=\linewidth]{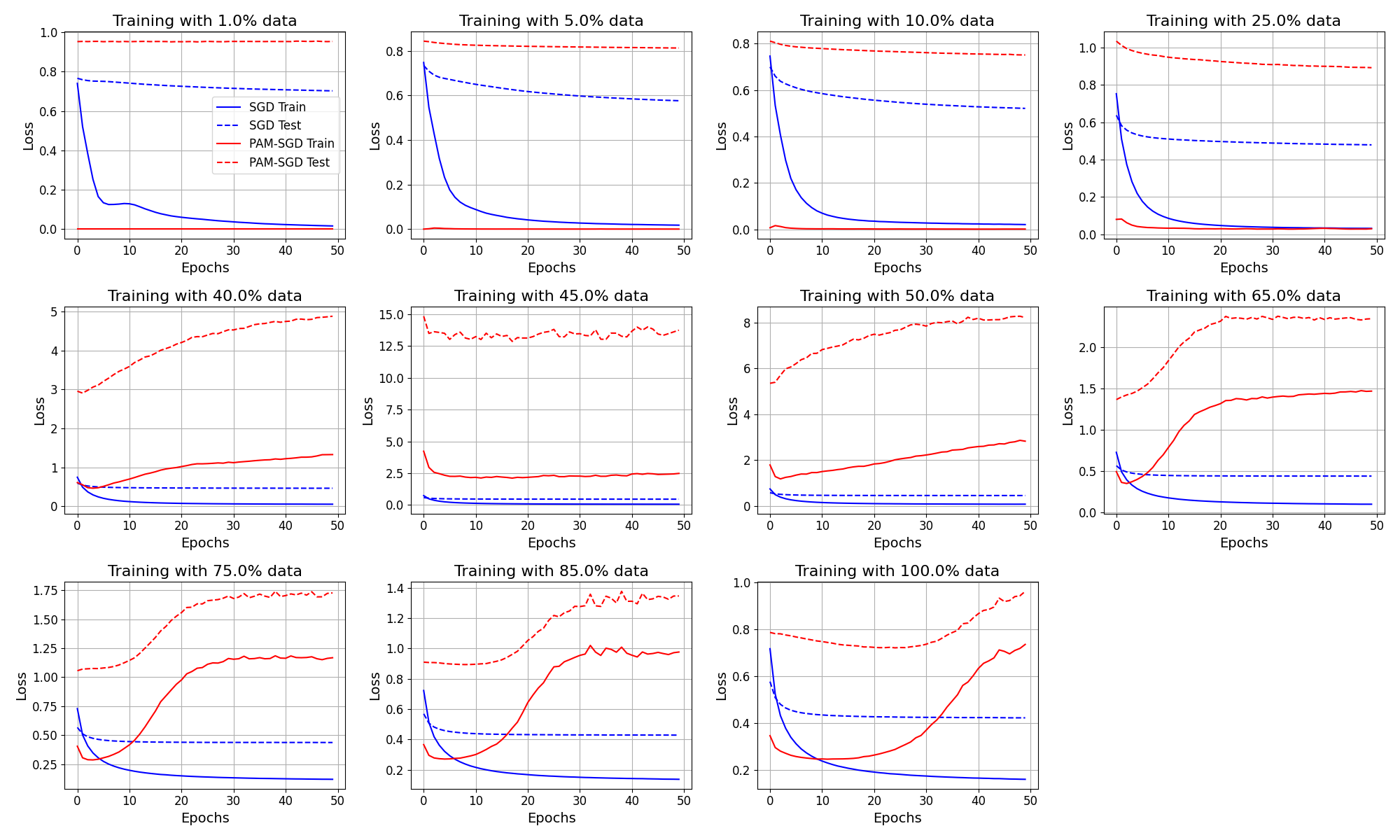}
    \caption{\textbf{Training and test loss curves across training set sizes for Gemma-2-2b with TopK activation, for $K=320$.} PAM-SGD has well-behaved training and test loss only for low data. }
    \label{fig:gemma_loss_curves_Top320}
\end{figure}

\begin{figure}[h]
    \centering
    \includegraphics[width=\linewidth]{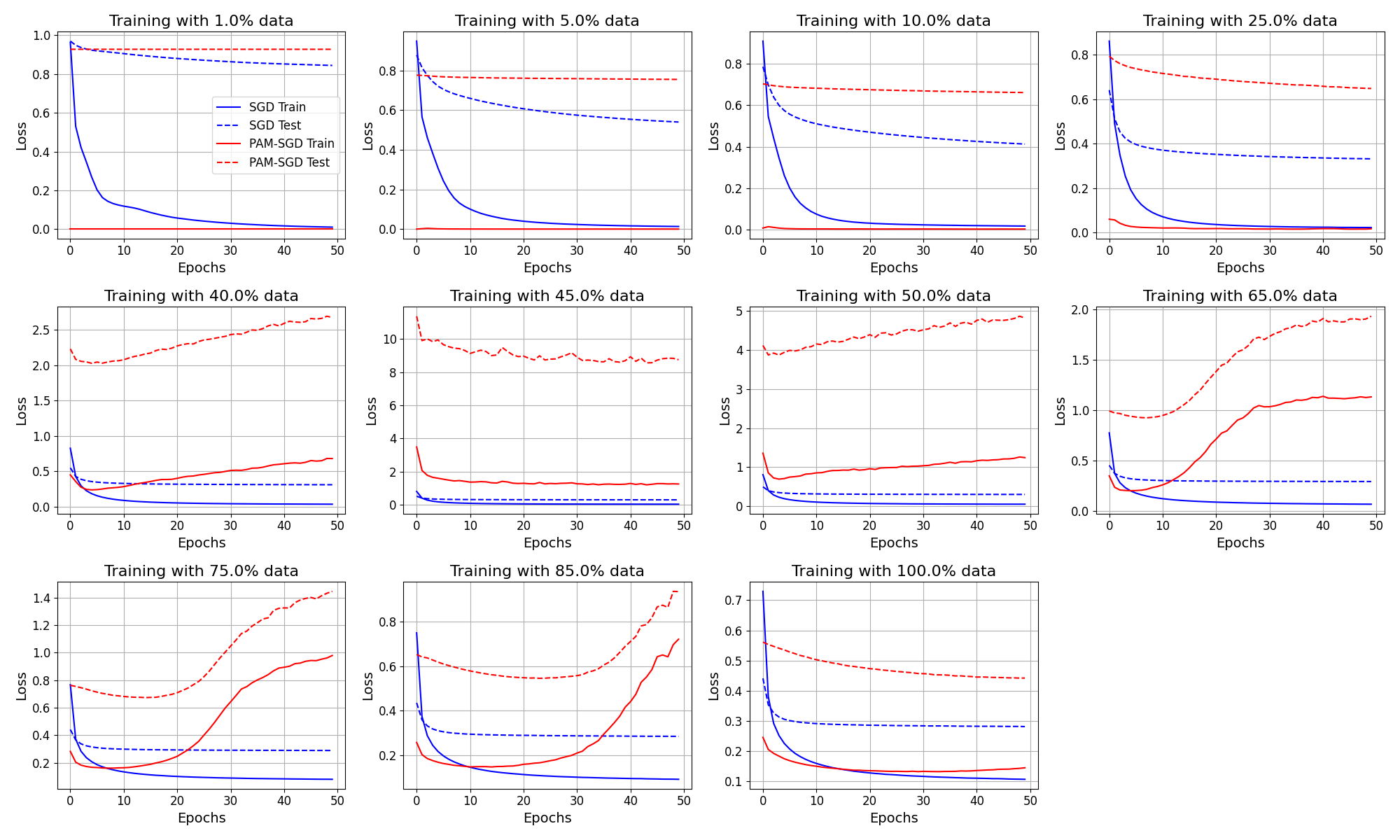}
    \caption{\textbf{Training and test loss curves across training set sizes for Gemma-2-2b with TopK activation, for $K=640$.} PAM-SGD has well-behaved training and test loss only for low data and 100\% data. }
    \label{fig:gemma_loss_curves_Top640}
\end{figure}

\paragraph{Stability only at high sparsity and underperforms SGD. (\cref{fig:gemma_K_values,fig:gemma_training_size_Top320,fig:gemma_loss_curves_Top640,fig:gemma_loss_curves_Top320,fig:gemma_training_size_Top640})}
PAM-SGD was highly unstable for low values of $K$, with the test loss diverging rapidly. Only for larger values was the test loss stable, but fairly stagnant agross epochs even for very large $K$ (over 30\% of the hidden dimension) leading us to choose $K=320$ as our default TopK sparsity. We speculate that this is because the LLM reconstruction is sufficiently complicated as to make being able to capture it with small $K$ unrealistic.

We furthermore compared SGD and PAM-SGD at various training data sizes for $K=320$ and $K=640$. We found that PAM-SGD consistently underperformed SGD in both cases, with the difference the smallest at low data sizes and again at high data sizes, with a surprising big rise in test loss for medium data sizes (with a maximum around 45\%). This peak was consistent across multiple runs, so we suspect it is some fundamental issue perhaps caused by numerical instability. Inspecting the loss curves in the two cases shows that PAM-SGD only has well-behaved training and test loss in the low data regime, or at 100\% data in the $K=640$ case.


\clearpage

\subsubsection{TopK and ReLU activation patterns}

\begin{figure}[htbp]
    \centering
    \includegraphics[width=0.8\linewidth]{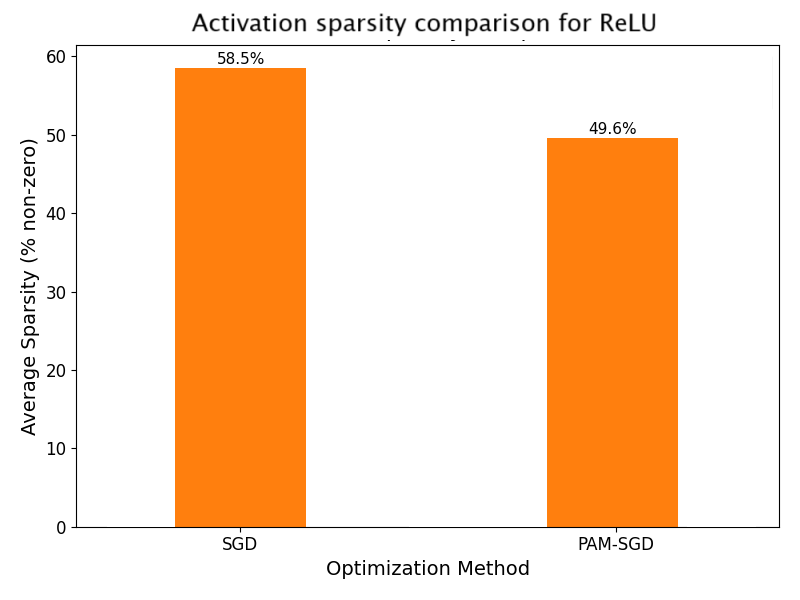}
    \caption{\textbf{Activation sparsity comparison.} ReLU yields much denser activations than TopK, and PAM-SGD activations about 15\% sparser than SGD activations.}
    \label{fig:gemma_sparsity}
\end{figure}

\begin{figure}
    \centering
    \includegraphics[width=\linewidth]{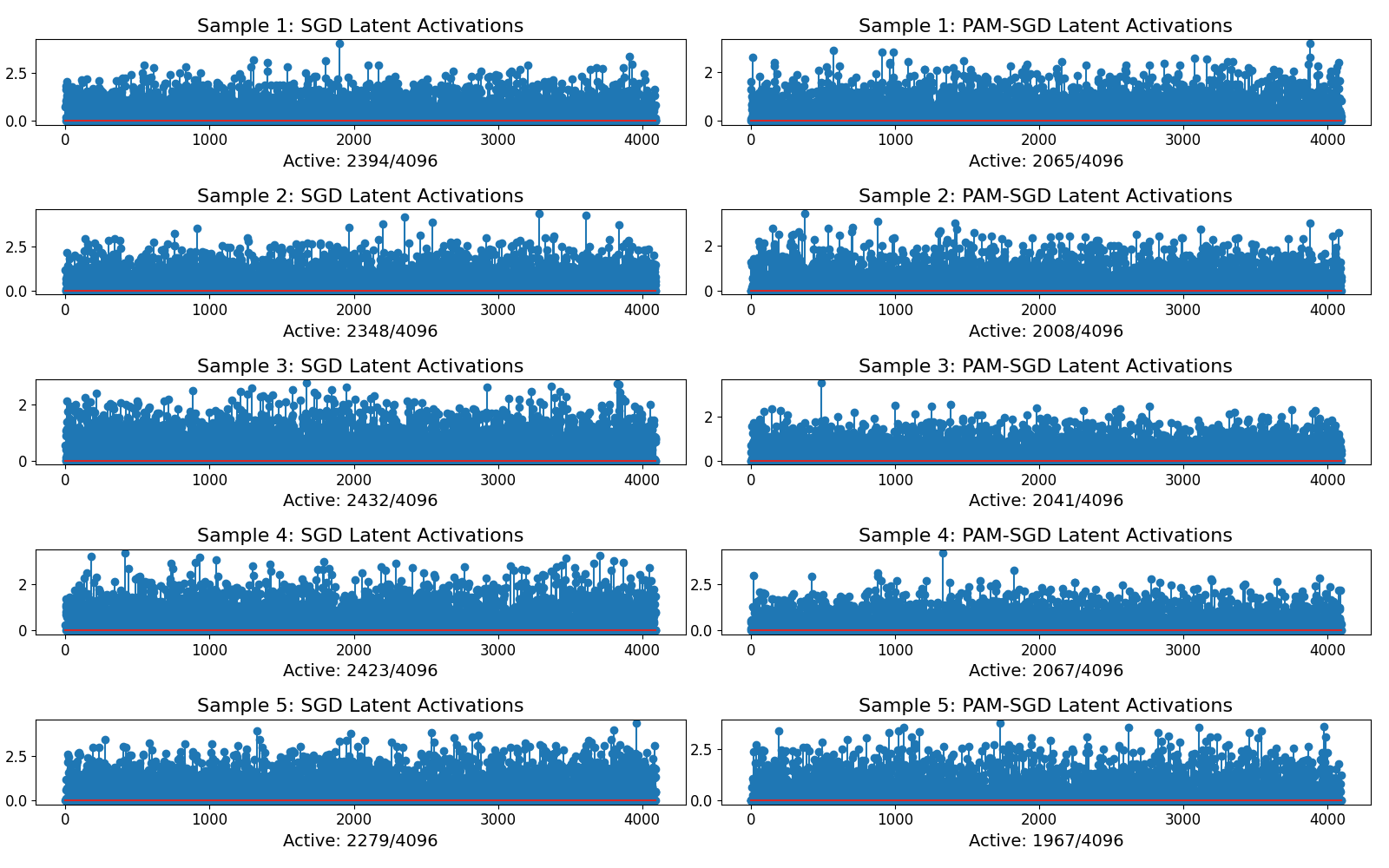}
    \caption{\textbf{Sparse activation patterns.} Plots showing which latent neurons were active in the ReLU case, comparing SGD (left) and PAM-SGD (right). }
    \label{fig:activation_RELU_Gemma_100}
\end{figure}

\begin{figure}
    \centering
    \includegraphics[width=\linewidth]{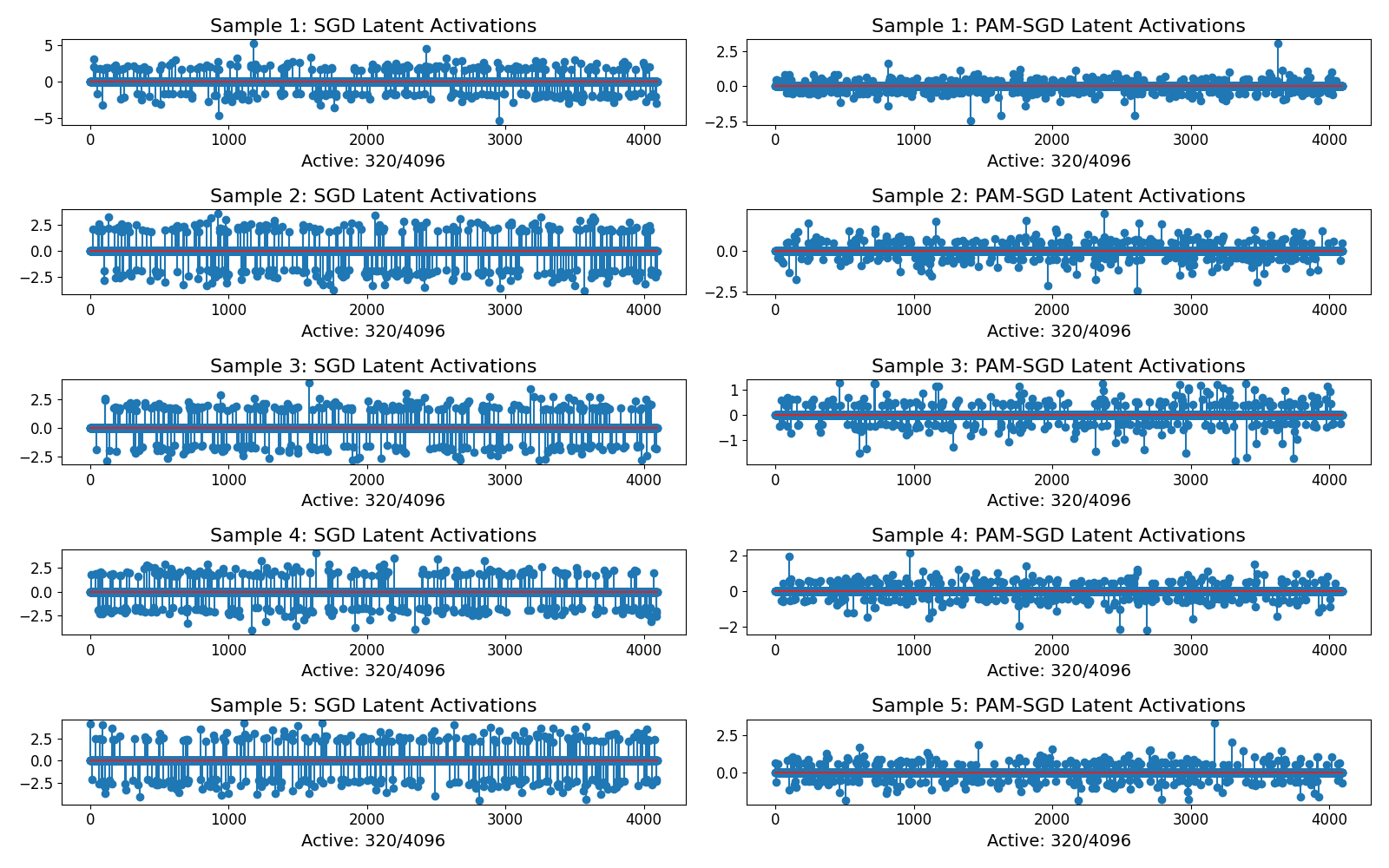}
    \caption{\textbf{Sparse activation patterns.} Plots showing which latent neurons were active in the TopK ($K=320$) case, comparing SGD (left) and PAM-SGD (right). }
    \label{fig:activation_TopK_Gemma_100}
\end{figure}

\paragraph{Sparsity Comparison. (\cref{fig:gemma_sparsity,fig:activation_RELU_Gemma_100,fig:activation_TopK_Gemma_100})}
TopK by design produces a constant sparsity $K=320$ for both SGD and PAM-SGD. ReLU produces much denser activations, around 58.5\% (approx. 2400) for SGD and 49.6\% (approx. 2000) for PAM-SGD. This increased sparsity from PAM-SGD is an important advantage of the method.

\clearpage
\clearpage

\subsubsection{Ablation study varying SGD updates per batch in PAM-SGD}
\begin{figure}[htbp]
    \centering
    \includegraphics[width=1\linewidth]{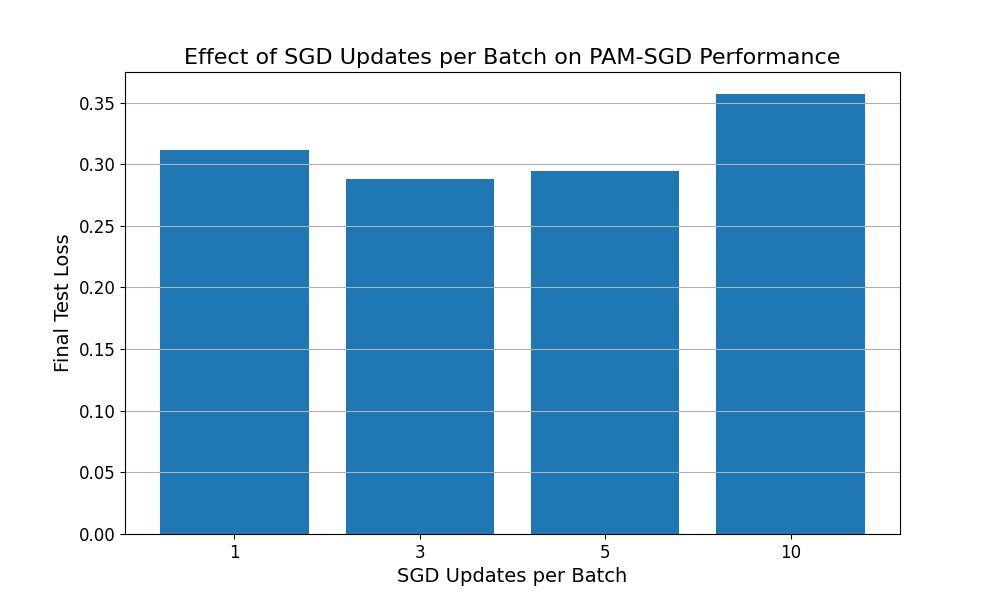}
    \caption{\textbf{Effect of SGD Updates per Batch on PAM-SGD Test Loss with ReLU activation.} Performance improves up to 3 updates but degrades beyond that, suggesting an optimal trade-off.}
    \label{fig:gemma_sgd_updates}
\end{figure}

\begin{figure}[htbp]
    \centering
    \includegraphics[width=1\linewidth]{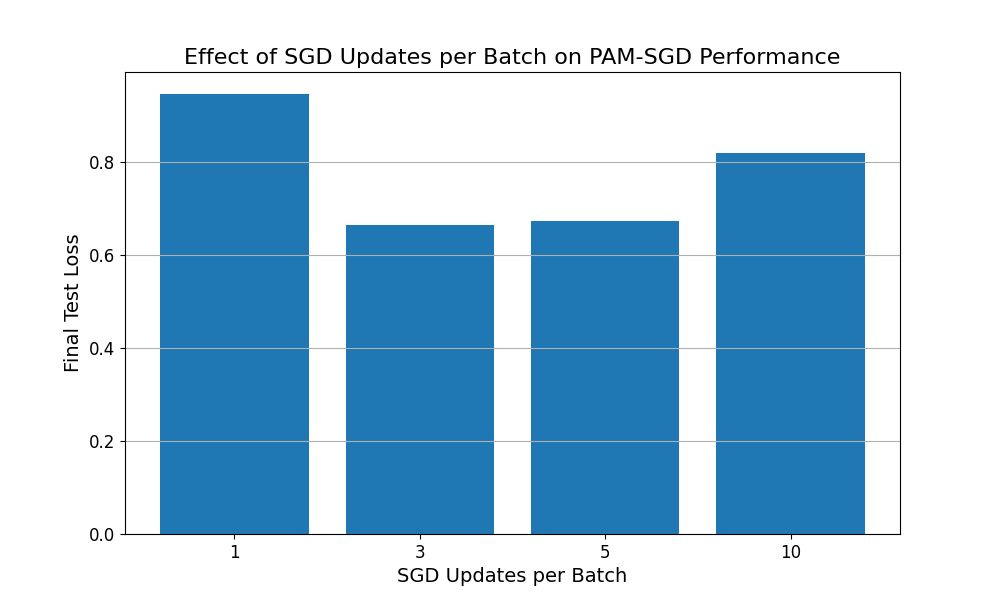}
    \caption{\textbf{Effect of SGD Updates per Batch on PAM-SGD Test Loss with TopK activation.} Performance again improves up to 3 updates but degrades beyond that.}
    \label{fig:gemma_sgd_updates_TopK}
\end{figure}

\paragraph{Number of SGD steps per batch matters for PAM-SGD. (\cref{fig:gemma_sgd_updates,fig:gemma_sgd_updates_TopK})}
For both the ReLU and TopK activations (with $K=320$), performance improves slightly when increasing SGD updates per batch from 1 to 3, but degrades beyond that. Too few updates prevent convergence of the inner optimization loop. Too many updates may lead to overfitting within the inner loop or instability due to misaligned gradients. PAM-SGD benefits from a moderate number of decoder updates per batch. An optimal value provides enough adaptation without overfitting, highlighting the importance of tuning this hyperparameter for practical deployments.

\clearpage
\subsubsection{Ablation study adding weight decay}

\begin{figure}[h]
    \centering
    \includegraphics[width=0.9\linewidth]{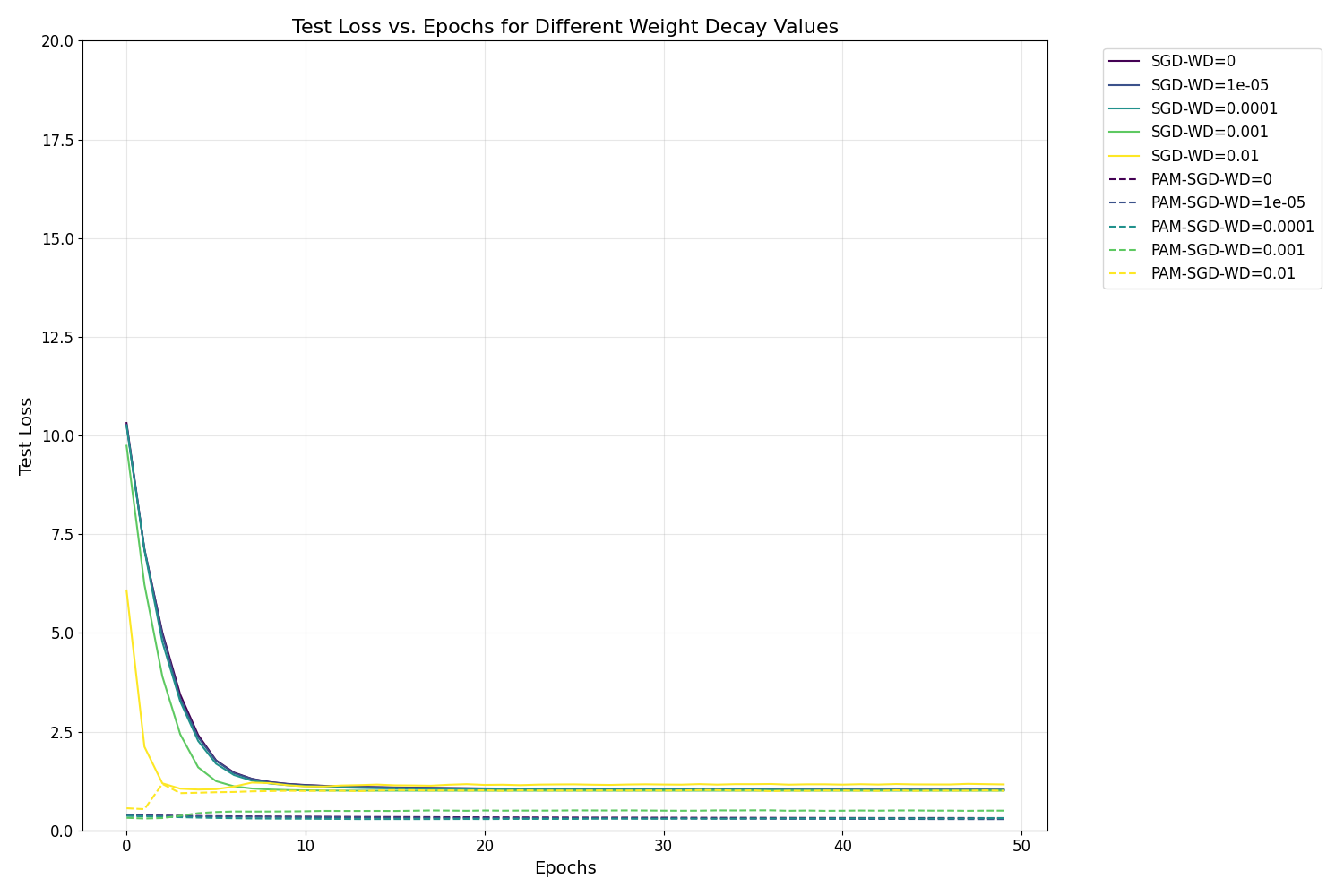}
    \caption{\textbf{Training and test loss curves with various weight decay parameters, with ReLU activation.} }
    \label{fig:weight_decay_Gemma_RELU}
\end{figure}

\begin{figure}[h]
    \centering
    \includegraphics[width=0.9\linewidth]{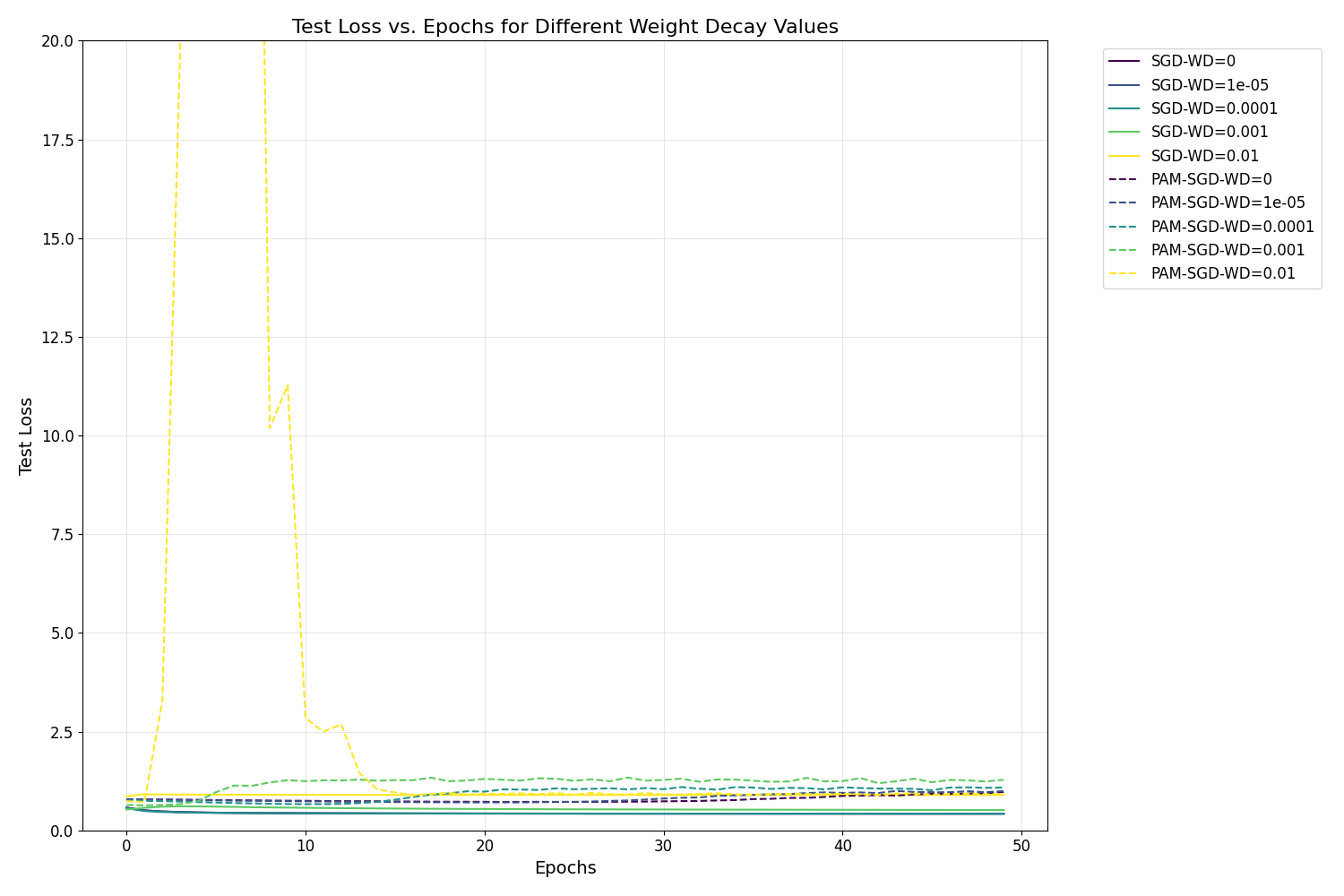}
    \caption{\textbf{Training and test loss curves with various weight decay parameters, with TopK $K=320$ activation.}}
    \label{fig:weight_decay_Gemma_TopK}
\end{figure}

\paragraph{Weight decay had a minor effect on performance. (\cref{fig:weight_decay_Gemma_RELU,fig:weight_decay_Gemma_TopK})}
Weight decay had a very minor effect on the SAE performance using either ReLU or TopK activations, with final test loss relatively constant, and PAM-SGD slightly outperforming SGD in the ReLU case and underperforming SGD in the TopK case. However, in the TopK case large values of weight decay are intially divergent before converging, whilst in the ReLU case this behaviour is less pronounced. 

\clearpage
\subsubsection{Ablation study varying the quadratic costs to move $\mu$ and $\nu$}

\begin{figure}[h]
    \centering
    \includegraphics[width=0.9\linewidth]{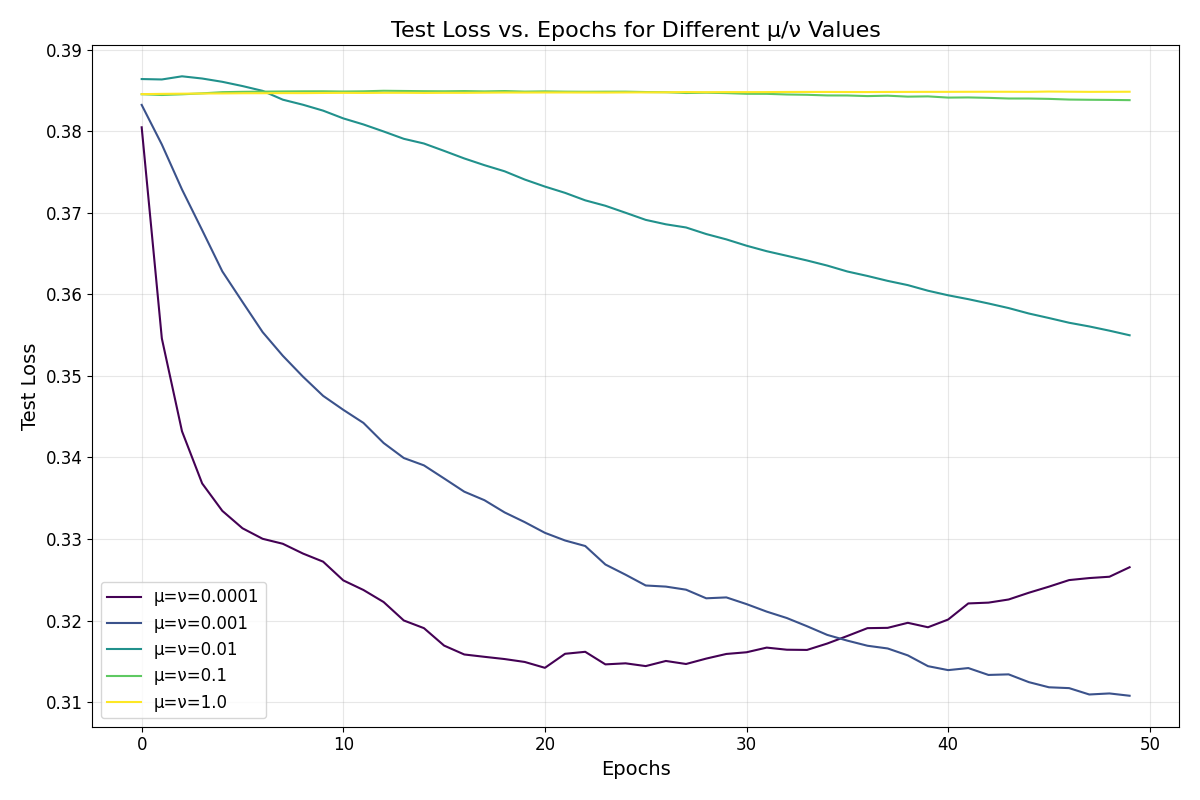}
    \caption{\textbf{Loss curves for ReLU activation at different ``cost to move'' parameters.}}
    \label{fig:mu_nu_Gemma_RELU}
\end{figure}

\paragraph{Sensitivity to quadratic costs to move $\mu$ and $\nu$. (\cref{fig:mu_nu_Gemma_RELU}) }
We studied the effect of varying the values of the parameters $\mu_\text{enc},\mu_\text{dec},\nu_\text{enc},$ and $\nu_\text{dec}$ from \cref{eq:PAM}. For ReLU activation, we found that very small values of these parameters caused the test loss to begin diverging, perhaps due to numerical instability. Slightly larger values improved performance, but increases beyond that slowed the learning process to no clear gain. In the TopK case, this divergence occurred at larger values than for ReLU, but went away once the parameters were sufficiently large.

\begin{figure}[h]
    \centering
    \includegraphics[width=0.9\linewidth]{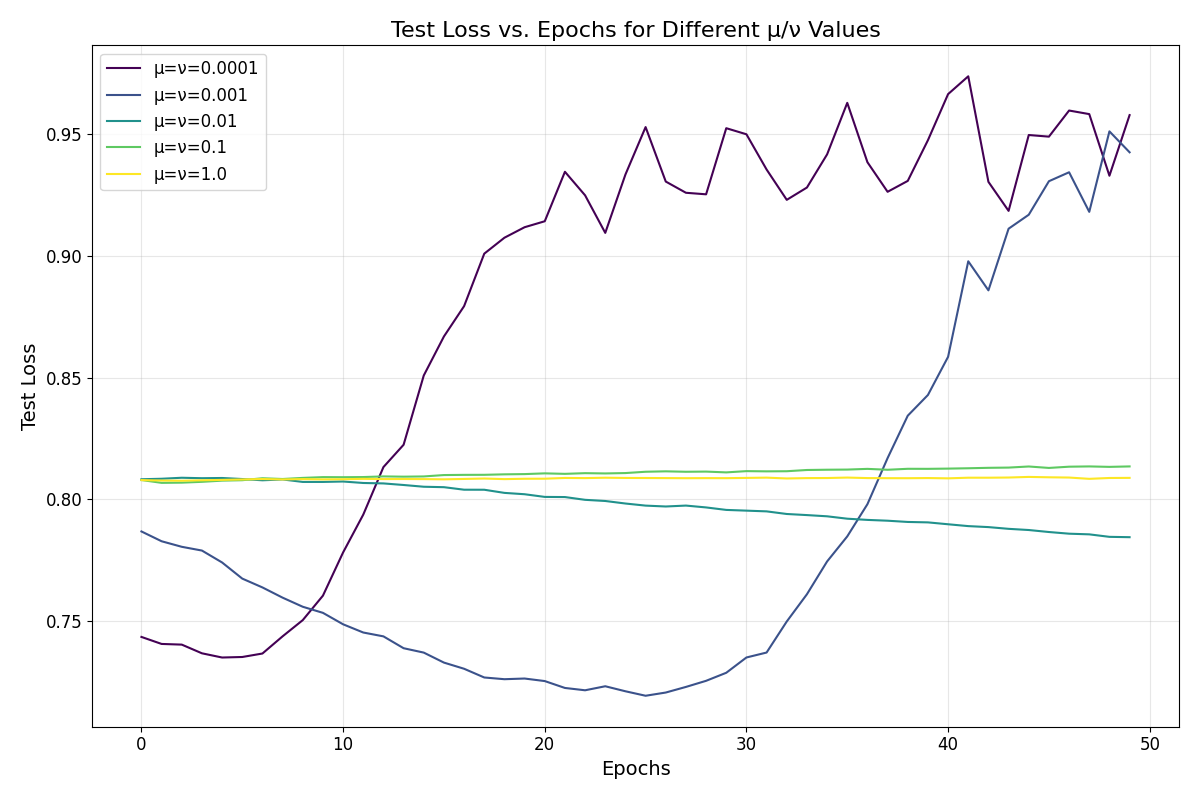}
    \caption{\textbf{Loss curves for TopK activation ($K=320$) at different ``cost to move'' parameters.}}
    \label{fig:mu_nu_Gemma_TopK}
\end{figure}

\end{document}